\documentclass[twoside]{article}

\usepackage[accepted]{aistats2024}
\usepackage[sort&compress,round]{natbib}

\usepackage{graphicx}
\usepackage[dvipsnames]{xcolor}
\usepackage{setspace}
\usepackage{nicefrac}
\usepackage{pgfplots}
\usepackage{float}
\usepackage[inline]{enumitem}
\usepackage{pifont}
\usepackage{tabularx,booktabs,threeparttable,makecell,colortbl}

\usepackage{calc}

\usepackage{amssymb}

\usepackage{amsmath,amsthm}
\usepackage{mathtools}
\usepackage{iftex}

\ifxetex
  \usepackage{fontspec}
  \usepackage{microtype}
  \usepackage{unicode-math}
  \newcommand{\mathmat}[1]{\mathbfit{#1}}
  \newcommand{\mathvec}[1]{\mathbfit{#1}}
  \newcommand{\mathvecgreek}[1]{\mathbfit{#1}}
  \newcommand{\mathmatgreek}[1]{\mathbfit{#1}}
  
  \newcommand{\boldupright}[1]{\symbfup{#1}}

  \newcommand{\mathrv}[1]{\mathsfit{#1}}
  \newcommand{\mathrvvec}[1]{\mathbfsfit{#1}}
  
  \newcommand{\mathrvvecgreek}[1]{\mathbfsfit{#1}}
\else
  \usepackage[notext,lcgreekalpha]{stix2}
  \usepackage{textcomp}

  \newcommand{\mathmat}[1]{\mathbfit{#1}}
  \newcommand{\mathvec}[1]{\mathbfit{#1}}
  \newcommand{\mathvecgreek}[1]{\mathbfit{#1}}
  \newcommand{\mathmatgreek}[1]{\mathbfit{#1}}
  
  \newcommand{\boldupright}[1]{\mathbf{#1}}

  \newcommand{\mathrv}[1]{\mathsfit{#1}}
  \newcommand{\mathrvvec}[1]{\mathbfsfit{#1}}
  
  \newcommand{\mathrvvecgreek}[1]{\mathbfsfit{#1}}
\fi

\usepackage{booktabs,threeparttable,arydshln}
\usepackage{multirow}
\usepackage{nicematrix}

\usepackage[algo2e, ruled]{algorithm2e}

\usepackage{pgfplots}
\pgfkeys{/pgfplots/tuftelike/.style={
  semithick,
  tick style={major tick length=4pt,semithick,black},
  separate axis lines,
  axis x line*=bottom,
  axis x line shift=2ex,
  xlabel shift=-2pt,
  axis y line*=left,
  axis y line shift=2ex,
  ylabel shift=-2pt}}
  
\usepackage{proof-at-the-end}

\usepackage{mdframed}
\mdfdefinestyle {mdleftbarcoloredstyle} {%
    leftline          = true,%
    rightline         = false,%
    topline           = false,%
    bottomline        = false,%
    linewidth         = 2.5pt,%
    backgroundcolor   = gray!10,%
    linecolor         = gray,%
    innertopmargin    = 0.7ex,%
    innerbottommargin = 0.7ex,%
    innerleftmargin   = 1.ex,%
    skipabove         = \parskip,%
    skipbelow         = \parskip,%
    ntheorem          = false,%
}

\mdfdefinestyle {coloredstyle} {%
    leftline          = false,%
    rightline         = false,%
    topline           = false,%
    bottomline        = false,%
    backgroundcolor   = gray!10,%
    skipabove         = \parskip,%
    skipbelow         = \parskip,%
    innertopmargin    = 1.0ex,%
    innerbottommargin = 0.7ex,%
    innerleftmargin   = 1.ex,%
}

\declaretheoremstyle[
    spaceabove    = \parsep,
    spacebelow    = \parsep,
    bodyfont      = \normalfont\itshape,
]{theoremsty}

\declaretheorem[name=Proposition,style=theoremsty, mdframed={style = coloredstyle}]{proposition}
\declaretheorem[name=Corollary,  style=theoremsty, mdframed={style = coloredstyle}]{corollary}

\declaretheorem[name=Theorem,    style=theoremsty, mdframed={style = coloredstyle}, numbered=no]{theorem*}
\declaretheorem[name=Proposition,style=theoremsty, mdframed={style = coloredstyle}, numbered=no]{proposition*}
\declaretheorem[name=Corollary,  style=theoremsty, mdframed={style = coloredstyle}, numbered=no]{corollary*}
\declaretheorem[name=Lemma,      style=theoremsty, mdframed={style = coloredstyle}, numbered=no]{lemma*}

\declaretheorem[name=Open Problem,style=theoremsty, mdframed={style = coloredstyle}, numbered=no]{openproblem*}
\declaretheorem[name=Conjecture,  style=theoremsty, mdframed={style = coloredstyle}, numbered=no]{conjecture*}

\declaretheoremstyle[
    spaceabove=\parsep,
    spacebelow=\parsep,
    bodyfont=\normalfont,
]{normalsty}
\declaretheorem[name=Remark,      style=normalsty]{remark}
\declaretheorem[name=Definition,  style=normalsty, mdframed={style = 
 coloredstyle}]{definition}
\declaretheorem[name=Assumption,  style=normalsty, mdframed={style = 
 coloredstyle} ]{assumption}

\declaretheorem[name=Remark,      style=normalsty, mdframed={style = 
 coloredstyle}, numbered=no]{remark*}
\declaretheorem[name=Definition,  style=normalsty, mdframed={style = coloredstyle}, numbered=no]{definition*}
\declaretheorem[name=Assumption,  style=normalsty, mdframed={style = coloredstyle}, numbered=no]{assumption*}

\makeatletter
\providecommand\IfFormatAtLeastTF{\@ifl@t@r\fmtversion}
\IfFormatAtLeastTF{2020/10/02}{%
  \usepackage[bookmarksopen=true,bookmarks=true,colorlinks=true,breaklinks,backref=page]{hyperref}
  \usepackage[capitalise, nameinlink]{cleveref}
  \renewcommand*{\backref}[1]{}
  \renewcommand*{\backrefalt}[4]{({\footnotesize%
  \ifcase #1 Not cited.%
    \or page~#2%
    \else pages #2%
  \fi%
  })}

}{%
  \usepackage{hyperref}
  \usepackage[capitalise, nameinlink]{cleveref}%
}
\makeatother

\makeatletter
\newcommand{\crefnames}[3]{%
  \@for\next:=#1\do{%
    \expandafter\crefname\expandafter{\next}{#2}{#3}%
  }%
}
\makeatother
\crefnames{part,chapter,section}{\S}{\S\S}

\definecolor{linkcolor}{HTML}{6929C4}
\definecolor{citecolor}{HTML}{0043CE}
\hypersetup{colorlinks={true},linkcolor=linkcolor,citecolor=citecolor}

\crefname{assumption}{Assumption}{Assumption}
\crefname{condition}{Condition}{Condition}

\crefname{framedtheorem}{Theorem}{Theorems}
\crefname{framedproposition}{Proposition}{Propositions}
\crefname{framedlemma}{Lemma}{Lemmas}

\makeatletter
\def\adl@drawiv#1#2#3{%
        \hskip.5\tabcolsep
        \xleaders#3{#2.5\@tempdimb #1{1}#2.5\@tempdimb}%
                #2\z@ plus1fil minus1fil\relax
        \hskip.5\tabcolsep}
\newcommand{\cdashlinelr}[1]{%
  \noalign{\vskip\aboverulesep
           \global\let\@dashdrawstore\adl@draw
           \global\let\adl@draw\adl@drawiv}
  \cdashline{#1}
  \noalign{\global\let\adl@draw\@dashdrawstore
           \vskip\belowrulesep}}
\makeatother

\pgfkeys{/prAtEnd/global custom defaults/.style={
    end,
    restate,
    text proof={Proof},
     text link={\textit{Proof.} See~\cref{proof:prAtEnd\pratendcountercurrent}.}
  }
}

\DeclareMathOperator*{\minimize}{minimize}

\DeclareMathOperator*{\argmin}{arg\,min} 

\newcommand*\xbar[1]{%
  \hbox{%
    \vbox{%
      \hrule height 0.6pt 
      \kern0.33ex
      \hbox{%
        \kern-0.1em
        \ensuremath{#1}%
        \kern-0.1em
      }%
    }%
  }%
}

\newcommand{\Vsub}[2]{\mathbb{V}_{#1}\left[ #2 \right]}

\newcommand{\DHF}[2]{\mathrm{D}_{\mathrm{F}}(#1,#2)}
\newcommand{\DKL}[2]{\mathrm{D}_{\mathrm{KL}}(#1,#2)}

\newcommand{\norm}[1]{{\left\lVert #1 \right\rVert}}
\newcommand{\abs}[1]{{\left| #1 \right|}}

\newcommand{\vg}{\mathvec{g}}

\newcommand{\vm}{\mathvec{m}}

\newcommand{\vu}{\mathvec{u}}

\newcommand{\vx}{\mathvec{x}}
\newcommand{\vy}{\mathvec{y}}
\newcommand{\vz}{\mathvec{z}}

\newcommand{\vmu}{\mathvecgreek{\mu}}
\newcommand{\vnu}{\mathvecgreek{\nu}}

\newcommand{\vlambda}{\mathvecgreek{\lambda}}

\newcommand{\rvu}{\mathrv{u}}

\newcommand{\rvx}{\mathrv{x}}

\newcommand{\rvvg}{\mathrvvec{g}}

\newcommand{\rvvu}{\mathrvvec{u}}

\newcommand{\rvvx}{\mathrvvec{x}}

\newcommand{\rvvz}{\mathrvvec{z}}

\newcommand{\rvmU}{\mathrvvecgreek{U}}

\newcommand{\mA}{\mathmat{A}}
\newcommand{\mB}{\mathmat{B}}
\newcommand{\mC}{\mathmat{C}}
\newcommand{\mD}{\mathmat{D}}

\newcommand{\mL}{\mathmat{L}}

\newcommand{\mR}{\mathmat{R}}
\newcommand{\mS}{\mathmat{S}}

\newcommand{\mSigma}{\mathmatgreek{\Sigma}}

\newcommand{\inner}[2]{\left\langle #1, #2 \right\rangle}

\usepackage{etoc}
\etocframedstyle[1]{\textbf{\textsc{Table of Contents}}}
\etocsettocdepth{3}

\begin{document}

%
\runningtitle{Linear Convergence of Black-Box Variational Inference}

%

\twocolumn[

\aistatstitle{Linear Convergence of Black-Box Variational Inference: \\ Should We Stick the Landing?}

\aistatsauthor{ Kyurae Kim \And Yi-An Ma \And Jacob R. Gardner }

\aistatsaddress{ University of Pennsylvania \And University of California San Diego \And  University of Pennsylvania }
]

\begin{abstract}
  We prove that black-box variational inference (BBVI) with control variates, particularly the sticking-the-landing (STL) estimator, converges at a geometric (traditionally called ``linear'') rate under perfect variational family specification.
  In particular, we prove a quadratic bound on the gradient variance of the STL estimator, one which encompasses misspecified variational families.
  Combined with previous works on the quadratic variance condition, this directly implies convergence of BBVI with the use of projected stochastic gradient descent.
  For the projection operator, we consider a domain with triangular scale matrices, which the projection onto is computable in \(\Theta(d)\) time, where \(d\) is the dimensionality of the target posterior.
  We also improve existing analysis on the regular closed-form entropy gradient estimators, which enables comparison against the STL estimator, providing explicit non-asymptotic complexity guarantees for both.
\end{abstract}

\section{INTRODUCTION}
Despite the massive success of black-box variational inference (BBVI; \citealp{ranganath_black_2014,titsias_doubly_2014,kucukelbir_automatic_2017}), our understanding of its computational properties has only recently started to make progress \citep{domke_provable_2019,hoffman_blackbox_2020,domke_provable_2020,domke_provable_2023,kim_convergence_2023,kim_practical_2023}. 
Notably, \citet{domke_provable_2023,kim_convergence_2023} have independently established the convergence of ``full'' BBVI.
This is a significant advance from the previous results where simplified versions of BBVI were analyzed \citep{hoffman_blackbox_2020,bhatia_statistical_2022} and results that \textit{a-priori} assumed regularity of the ELBO~\citep{alquier_concentration_2020,cherief-abdellatif_generalization_2019,regier_fast_2017,khan_faster_2016,khan_kullbackleibler_2015,fujisawa_multilevel_2021,buchholz_quasimonte_2018,liu_quasimonte_2021}.
We now have rigorous convergence guarantees that, for certain well-behaved posteriors, BBVI achieves a convergence rate of \(\mathcal{O}\left(1/T\right)\), corresponding to a computational complexity of \(\mathcal{O}\left(1/\epsilon\right)\)~\citep{domke_provable_2023,kim_convergence_2023}.
A remaining theoretical question is whether BBVI can achieve better rates, in particular geometric convergence rates, which is traditionally called ``linear'' convergence in the optimization literature (see the textbook by \citealt[\S 1.2.3]{nesterov_introductory_2004}), corresponding to a complexity of \(\mathcal{O}\left(\log\left( 1/\epsilon \right) \right)\).

For stochastic gradient descent (SGD; \citealp{robbins_stochastic_1951,bottou_online_1999,nemirovski_robust_2009}), it is known that improving the \(\mathcal{O}\left(1/T\right)\) convergence rate is challenging~\citep{rakhlin_making_2012,harvey_tight_2019}.
This is because, once in the stationary regime, it is necessary to either decrease the stepsize or average the iterates, where the latter reduces SGD to Markov chain Monte Carlo~\citep{dieuleveut_bridging_2020}.
Not surprisingly, both cases result in a significant slowdown compared to deterministic gradient descent.
Overall, SGD is known to achieve \(\mathcal{O}\left(1/\sqrt{T}\right)\) for general convex functions and \(\mathcal{O}\left(1/T\right)\) for strongly convex functions (\citealp{nemirovski_robust_2009,shalev-shwartz_pegasos_2011,moulines_nonasymptotic_2011}; for more modern analysis techniques, see~\citealp{garrigos_handbook_2023}).

Meanwhile, under a condition known as ``interpolation,'' which assumes that the gradient variance becomes zero at the optimum, SGD is known to achieve a linear convergence rate~\citep{schmidt_fast_2013}.
This can automatically hold for certain problems, such as empirical risk minimization (ERM) with overparameterized models, explaining the fast empirical convergence of modern machine learning models~\citep{vaswani_fast_2019, ma_power_2018}.
Also, control variate methods such as ``variance-reduced'' gradients \citep{schmidt_minimizing_2017,johnson_accelerating_2013,gower_variancereduced_2020} algorithmically achieve the same effect and have been successful both in theory and practice.
Unfortunately, variance-reduced gradient methods are strictly restricted to the finite-sum setting, which BBVI is not part of (See \S 2.4 by \citealp{kim_convergence_2023}).
Thus, it is yet unclear how BBVI could benefit from the advances in variance-reduced gradients.

Fortunately, other types of control variates have been actively pursued in BBVI \citep{geffner_using_2018,miller_reducing_2017, ranganath_black_2014, paisley_variational_2012, geffner_approximation_2020,wang_joint_2024}. In particular, the \textit{sticking-the-landing} (STL; \citealp{roeder_sticking_2017}) estimator satisfies the interpolation condition (e.g., achieves zero gradient variance at the optimum) when the variational family $\mathcal{Q}$ contains the true posterior $\pi$~\footnote{Although the term interpolation does not literally make sense outside of the ERM context, we will stick to this term to stay in line with the SGD literature.}. It is thus natural to ask whether existing control variate approaches such as STL are sufficient to achieve linear convergence under realizable conditions.
In fact, this possibility has been mentioned by~\citet[\S 5]{hoffman_blackbox_2020}.

In this work, we confirm these conjectures by establishing a linear convergence rate of BBVI with STL when the variational family contains the true posterior--\textit{i.e.}, is perfectly specified.
For a \(d\)-dimensional strongly log-concave posterior with a condition number of \(\kappa\) and a location-scale variational family with a full rank scale, BBVI with the STL estimator finds variational parameters \(\epsilon\)-close to the global optimum at a rate of \(\mathcal{O}\left(d \kappa^2 \log\left(1/\epsilon\right) \right)\). Even beyond the perfectly specified setting, our theoretical results characterize the behavior of the STL estimator in the misspecified setting, which is closer to practice.
This provides some intuition as to why the comparisons between the STL and the ``standard'' closed-form entropy (CFE;~\citealp{titsias_doubly_2014,kucukelbir_automatic_2017}) estimators appear mixed in practice~\citep{geffner_rule_2020,agrawal_advances_2020}.

While our results are built on top of the theoretical framework of \citet{domke_provable_2023}, a similar convergence result on the STL estimator appeared in a later, recent preprint version \citep{domke_provable_2023a} concurrently with this work.
The details of the result differ, and we provide additional contributions specific to the STL estimator.
We discuss the differences in more detail in \cref{section:related} along with other related works.

\vspace{-1ex}
\paragraph{Contributions} Our contributions are summarized in the following list.
An overview of the theorems is provided in \cref{table:theorems} of \cref{section:overviewtheorems}.
We also provide an overview of previous rigorous complexity analyses on BBVI in \cref{table:relatedworks} of \cref{section:related}.
\begin{enumerate}[label=\ding{228}]
  \vspace{-1ex}
    \setlength\itemsep{1ex}
    \item \textbf{We prove that BBVI with the STL estimators can converge at a linear rate.} \\
      When the variational family is perfectly specified such that the posterior is contained in the variational family, \cref{thm:projsgd_bbvistl_complexity} establishes this through \cref{thm:stl_upperbound}.
      This is the first result for ``full'' BBVI without algorithmic simplification. 
      
    \item \textbf{Our analysis encompasses variational family misspecification.}
      When the variational family is misspecified, the Fisher-Hyv\"arinen divergence between the variational posterior and the true posterior captures the behavior of STL.
      
    \item \textbf{We establish a matching lower bound on the gradient variance.}
      Our upper bound in \cref{thm:stl_upperbound} and the concurrent result by \citet{domke_provable_2023a} are proven to be tight by a constant factor through \cref{thm:stl_lowerbound_unimprovability}.
    
    \item \textbf{We improve previously obtained gradient variance bounds for the CFE estimator.} \\
      In \cref{thm:cfe_upperbound}, we tighten the constants of the bounds previously obtained by \citet{domke_provable_2023}.
      This makes the theoretical results for the CFE and STL estimators comparable.
      
    \item \textbf{We provide a parameterization with a projection operator with \(\Theta(d)\) complexity.} \\
      In \cref{section:triangular_scale}, we propose a triangular scale parameterization with a corresponding projection operator that can be computed in \(\Theta(d)\) time.
      This improves over the matrix square-root parameterization used by \citet{domke_provable_2023}, which involved a \(\mathcal{O}(d^3)\) projection operator based on the singular value decomposition.
      
    \item \textbf{We prove precise quantitative complexity guarantees for SGD with QV gradient estimators.}
      We obtain quantatitive non-asymptotic complexity guarantees from the ``anytime convergence'' results of \citet{domke_provable_2023}.
\end{enumerate}

\section{PRELIMINARIES}\label{section:preliminaries}

{
\paragraph{Notation}
Random variables are denoted in sans-serif (\textit{e.g.}, \(\rvx\), \(\rvvx\)), vectors are in bold (\textit{e.g.}, \(\vx\), \(\rvvx\)), and matrices are in bold capitals (\textit{e.g.} \(\mA\)).
For a vector \(\vx \in \mathbb{R}^d\), we denote the inner product as \(\vx^{\top}\vx\) and \(\inner{\vx}{\vx}\), the \(\ell_2\) norm as \(\norm{\vx}_2 = \sqrt{\vx^{\top}\vx}\).
For a matrix {\footnotesize\(\mA\), \(\norm{\mA}_{\mathrm{F}} = \sqrt{\mathrm{tr}\left(\mA^{\top} \mA\right)}\)} denotes the Frobenius norm, and for some matrix \(\mB\), \(\mA \succeq \mB\) is the Loewner order implying that \(\mA - \mB\) is a positive semi-definite matrix.
\(\mathbb{S}^d\), \(\mathbb{S}^d_{++}\), \(\mathbb{L}^d\), \(\mathbb{L}^d_{++}\) are the set of symmetric, positive definite, triangular, and triangular matrices with strictly positive eigenvalues (Cholesky factors).
\(\sigma_{\mathrm{min}}\left(\mA\right)\) is the smallest eigenvalue of \(\mA\).
}

\subsection{Variational Inference}

Variational inference (VI,~\citealp{zhang_advances_2019, blei_variational_2017, jordan_introduction_1999}) aims to minimize the exclusive (or reverse) Kullback-Leibler (KL; \citealp{kullback_information_1951}) divergence as:
{\begingroup
\setlength{\belowdisplayskip}{1.5ex} \setlength{\belowdisplayshortskip}{1.5ex}
\setlength{\abovedisplayskip}{1.5ex} \setlength{\abovedisplayshortskip}{1.5ex}
\begin{align*}
    \minimize_{\vlambda \in \Lambda}\; \mathrm{D}_{\mathrm{KL}}\left(q_{\vlambda}, \pi\right)
    \triangleq
    \mathbb{E}_{\rvvz \sim q_{\vlambda}} -\log \pi \left(\rvvz\right) -\mathbb{H}\left(q_{\vlambda}\right),
\end{align*}
\endgroup}
\begin{center}
  \vspace{-2ex}
  {\begingroup
  \begin{tabular}{lll}
    where 
    & \(\mathrm{D}_{\mathrm{KL}}\) & is the KL divergence, \\
    & \(\mathbb{H}\)     & is the differential entropy, \\
    & \(\pi\) & is the (target) posterior distribution,  \\
    & \(q_{\vlambda}\) & is the variational approximation. \\
  \end{tabular}
  \vspace{-2ex}
  \endgroup}
\end{center}

For Bayesian posterior inference, the KL divergence is, unfortunately, intractable.
Instead, one equivalently minimizes the negative \textit{evidence lower bound} (ELBO;~\citealp{jordan_introduction_1999}) \(F\) such that:
{%
\setlength{\belowdisplayskip}{1.5ex} \setlength{\belowdisplayshortskip}{1.5ex}
\setlength{\abovedisplayskip}{1.5ex} \setlength{\abovedisplayshortskip}{1.5ex}
\[
  \minimize_{\vlambda \in \Lambda}\; F\left(\vlambda\right)
  \triangleq
  \mathbb{E}_{\rvvz \sim q_{\vlambda}} -\log \ell \left(\rvvz\right) - \mathbb{H}\left(q_{\vlambda}\right),
\]
}%
where \(\ell\left(\vz\right) \propto \pi\left(\vz\right)\) is the unnormalized posterior proportional up to a multiplicative constant.
In typical use cases of VI, we only have access to \(\ell\) but not \(\pi\), and the normalizing constant is intractable.

\vspace{-1ex}
\paragraph{Black-Box Variational Inference}
Black-box variational inference (BBVI; \citealp{ranganath_black_2014,titsias_doubly_2014}) minimizes \(F\) by leveraging stochastic gradient descent (SGD; \citealp{robbins_stochastic_1951,bottou_online_1999,nemirovski_robust_2009}).
By obtaining a stochastic estimate \(\rvvg\left(\vlambda\right)\) which is \textit{unbiased}  as \(\mathbb{E} \rvvg\left(\vlambda\right) = \nabla F\left(\vlambda\right)\), BBVI repeats:
{
\setlength{\belowdisplayskip}{1ex} \setlength{\belowdisplayshortskip}{1ex}
\setlength{\abovedisplayskip}{1ex} \setlength{\abovedisplayshortskip}{1ex}
\[
  \vlambda_{t+1} = \mathrm{proj}\left( \vlambda_t - \gamma_t \rvvg \right), 
\]
}%
where \(\gamma_t\) is called the stepsize.
The use of the projection operator \(\mathrm{proj}\left(\cdot\right)\) forms a subset of the broad SGD framework called \textit{projected} SGD.
The convergence of BBVI with projected SGD has recently been established by \citet{domke_provable_2023}.

In addition to the KL divergence, our analysis invokes the Fisher-Hyv\"arinen divergence~\citep{otto_generalization_2000,hyvarinen_estimation_2005}:
\begin{definition}[\textbf{Fisher-Hyv\"arinen Divergence}]
The \(p\)th order Fisher-Hyv\"arinen divergence between two distributions \(\pi\) and \(q\) is given as
{%
\setlength{\belowdisplayskip}{1.ex} \setlength{\belowdisplayshortskip}{1.ex}
\setlength{\abovedisplayskip}{1.5ex} \setlength{\abovedisplayshortskip}{1.5ex}
\[
  \mathrm{D}_{\mathrm{F}^p}\left(q, \pi\right)
  \triangleq 
  \mathbb{E}_{\rvvz \sim q} 
  \norm{\nabla \log \pi\left(\rvvz\right) - \nabla \log q\left(\rvvz\right) }_2^p.
\]
}%
\end{definition}
Here, we use the \(p\)th order generalization~\citep{huggins_practical_2018} of the original Fisher-Hyv\"arinen divergence. 
We denote the standard 2nd order Fisher-Hyv\"arinen divergence as \(\mathrm{D}_{\mathrm{F}}\left(q, \pi\right) \triangleq \mathrm{D}_{\mathrm{F}^2}\left(q, \pi\right)\).
This divergence was first defined by~\citet{otto_generalization_2000} (attributed by \citealp{zegers_fisher_2015}) as the ``relative Fisher information'' in the context of optimal transport.
It was later introduced to the machine learning community by \citet{hyvarinen_estimation_2005} for score matching.

\subsection{Variational Family}
Throughout this paper, we restrict our interest to the location-scale variational family, which has been successfully used by \citet{kim_practical_2023,kim_convergence_2023,domke_provable_2019,domke_provable_2020,domke_provable_2023, fujisawa_multilevel_2021, titsias_doubly_2014} for analyzing the properties of BBVI.
It encompasses many practical families such as the Gaussian, Student-t, and elliptical distributions.
In particular, the location-scale family is part of the broader reparameterized family:
\begin{definition}[\textbf{Reparameterized Family}]\label{def:family}
  Let \(\varphi\) be some \(d\)-variate distribution.
  Then, \(q_{\vlambda}\) that can be equivalently represented as
{%
\setlength{\belowdisplayskip}{-.5ex} \setlength{\belowdisplayshortskip}{-.5ex}
\setlength{\abovedisplayskip}{-.5ex} \setlength{\abovedisplayshortskip}{-.5ex}
  \begin{alignat*}{2}
    \rvvz \sim q_{\vlambda}  \quad\Leftrightarrow\quad &\rvvz \stackrel{d}{=} \mathcal{T}_{\vlambda}\left(\rvvu\right); \quad \rvvu \sim  \varphi,
  \end{alignat*}
  }%
  where \(\stackrel{d}{=}\) is equivalence in distribution, is said to be part of a reparameterized family generated by the base distribution \(\varphi\) and the reparameterization function \(\mathcal{T}_{\vlambda}\).
\end{definition}
Naturally, this means we focus on the \textit{reparameterization gradient estimator}~\citep{titsias_doubly_2014,kingma_autoencoding_2014,rezende_stochastic_2014}, often observed to achieve lower variance than alternatives~\citep{xu_variance_2019}.  
(See \citealt{mohamed_monte_2020} for a comprehensive overview.)
From this, we obtain the location-scale family through the following reparameterization function:
\begin{definition}[\textbf{Location-Scale Reparameterization Function}]\label{def:reparam}
  A mapping \(\mathcal{T}_{\vlambda} : \mathbb{R}^p \times \mathbb{R}^d \rightarrow \mathbb{R}^d\) defined as
{
\setlength{\belowdisplayskip}{.5ex} \setlength{\belowdisplayshortskip}{.5ex}
\setlength{\abovedisplayskip}{1.ex} \setlength{\abovedisplayshortskip}{1.ex}
  \begin{align*}
    &\mathcal{T}_{\vlambda}\left(\vu\right) \triangleq \mC \vu + \vm
  \end{align*}
}%
  with \(\vlambda \in \mathbb{R}^p\) containing the parameters for forming the location \(\vm \in \mathbb{R}^d\) and scale \(\mC \in \mathbb{R}^{d \times d}\) is called the location-scale reparameterization function.
\end{definition}
For the scale matrix \(\mC\), various parameterizations are used in practice, as shown by \citet[Table 1]{kim_practical_2023}.
We discuss our parameterization of choice in \cref{section:scale_parameterization}.

The choice for the base distribution \(\varphi\) completes the specifics of the variational family.
For example, choosing \(\varphi\) to be a univariate Gaussian result in the Gaussian variational family.
We impose the following general assumptions on the base distribution:
\begin{assumption}[\textbf{Base Distribution}]\label{assumption:symmetric_standard}
  \(\varphi\) is a \(d\)-dimensional distribution such that \(\rvvu \sim \varphi\) and \(\rvvu = \left(\rvu_1, \ldots, \rvu_d \right)\) with indepedently and identically distributed components.
  Furthermore, \(\varphi\) is
  \begin{enumerate*}[label=\textbf{(\roman*)}]
      \item symmetric and standardized such that \(\mathbb{E}\rvu_i = 0\), \(\mathbb{E}\rvu_i^2 = 1\), \(\mathbb{E}\rvu_i^3 = 0\), and 
      \item has finite kurtosis \(\mathbb{E}\rvu_i^4 = \kappa < \infty\).
  \end{enumerate*}
\end{assumption}

Overall, the assumptions on the variational family are collected as follows:
\vspace{0.5ex}
\begin{assumption}\label{assumption:variation_family}
  The variational family is the location-scale family formed by \cref{def:family,def:reparam} with the base distribution \(\varphi\) satisfying \cref{assumption:symmetric_standard}.
\end{assumption}

\vspace{-1ex}
\subsection{Scale Parameterization}\label{section:scale_parameterization}
\vspace{-1ex}
For the scale parameterization \(\vlambda \mapsto \mC\), in principle, any choice that results in a positive-definite covariance matrix is valid.
However, recently, \citet{kim_convergence_2023} have shown that a seemingly innocent choice of parameterization can have a massive impact on computational performance.
For example, nonlinear parameterizations can easily break the strong convexity of the ELBO~\citep{kim_convergence_2023}, which could have been otherwise obtained~\citep{domke_provable_2020}.
Therefore, the scale parameterization is subject to the constraints:
\vspace{-1ex}
\begin{enumerate}[label=\textbf{(\roman*)}]
    \setlength\itemsep{0ex}
  \item \textbf{Positive Definiteness}: \(\mC\mC^{\top} \succ 0\).  \\
    This is needed to ensure that \(\mC\mC^{\top}\) forms a valid covariance in \(\mathbb{S}_{++}^d\).
    \label{item:positive_definite}
  \item \textbf{Linearity}: \(\norm{\vlambda - \vlambda'}_2^2 = \norm{\vm - \vm'}_2^2 + \norm{ \mC - \mC'}_{\mathrm{F}}^2 \).  \\
    As shown by \citet{kim_convergence_2023}, this constraint is sufficient to form a \(\mu\)-strongly convex ELBO from a \(\mu\)-strongly log-concave posterior.
    \label{item:linearity}
    
  \item \textbf{Convexity}: The mapping \(\vlambda \mapsto \mC\mC^{\top}\) is convex on \(\Lambda_{S}\).\\
    This is needed to ensure that the ELBO is convex whenever the target posterior is log-concave \citep{domke_provable_2023,kim_convergence_2023}.
    \label{item:convexity}
    
  \item \textbf{Smooth Log-Determinant}: \(\vlambda \mapsto \log \operatorname{det} \mC\) is Lipschitz smooth on \(\Lambda_{S}\). \label{item:smooth_entropy}
  
  This condition is only required by projected SGD so that the ELBO is Lipschitz smooth on \(\Lambda_S\).
  
\end{enumerate}
\vspace{-1ex}
\citet{domke_provable_2020,domke_provable_2023} guaranteed \labelcref{item:smooth_entropy} by setting the domain of \(\vlambda\) to be
{%
\setlength{\belowdisplayskip}{1.5ex} \setlength{\belowdisplayshortskip}{1.5ex}
\setlength{\abovedisplayskip}{1.5ex} \setlength{\abovedisplayshortskip}{1.5ex}
\begin{align*}
  \big\{\, \left(\vm, \mC\right) \in \mathbb{R}^{d} \times \mathbb{S}_{++}^d \mid \mC \mC^{\top} \succeq 
   S^{-1} \boldupright{I} \,\big\},
\end{align*}
}%
with \(S = L\), where \(L\) is the log-smoothness constant of the posterior.
That is, \(\mC\) is chosen to be a proper matrix square root of the covariance \(\mSigma = \mC \mC^{\top}\) such that \(\mC = \mC^{\top} = \mSigma^{\nicefrac{1}{2}}\).
This parameterization ensures that a proper projection operator exists onto \(\Lambda_S\), where they proposed to use the singular value decomposition.
This projection operator is quite costly as it imposes a \(\mathcal{O}(d^3)\) complexity.
We will later propose a different parameterization based on triangular matrices, where the projection operator only costs \(\Theta(d)\) while obtaining the same convergence guarantees. 

\subsection{Gradient Estimators}\label{section:gradient_estimators}
The gradient estimators considered in this work are the closed-form entropy (CFE;~\citealp{kucukelbir_automatic_2017,titsias_doubly_2014}) and sticking the landing (STL;~\citealp{roeder_sticking_2017}) estimators.

\vspace{-1ex}
\paragraph{Closed-From Entropy Estimator}
The CFE estimator is the ``standard'' estimator used for BBVI.
\begin{definition}[\textbf{Closed-Form Entropy Estimator}]\label{def:cfe}
The closed-form entropy gradient estimator is
{%
\setlength{\belowdisplayskip}{1.5ex} \setlength{\belowdisplayshortskip}{1.5ex}
\setlength{\abovedisplayskip}{1.5ex} \setlength{\abovedisplayshortskip}{1.5ex}
\[
  \vg\left(\vlambda\right) 
  \triangleq
  \nabla_{\vlambda} \log \ell \left(\mathcal{T}_{\vlambda}\left(\vu\right)\right) + \nabla_{\vlambda} \mathbb{H}\left(q_{\vlambda}\right),
\]
}%
where the gradient of the entropy term is computed deterministically.
\end{definition}
It can be used whenever the entropy \(\mathbb{H}\left(q_{\vlambda}\right)\) is available in a closed form.
For location-scale families, this is always the case up to an additive constant.

\vspace{-1ex}
\paragraph{Sticking-the-Landing Estimator}
On the other hand, the STL estimator estimates the entropy through a
 special Monte Carlo strategy:
\begin{definition}[\textbf{Sticking-the-Landing Estimator; STL}]\label{def:stl}
The sticking-the-landing gradient estimator 
{%
\setlength{\belowdisplayskip}{1.ex} \setlength{\belowdisplayshortskip}{1.ex}
\setlength{\abovedisplayskip}{1.ex} \setlength{\abovedisplayshortskip}{1.ex}
\[
  \rvvg_{\mathrm{STL}}\left(\vlambda\right) 
  \triangleq
  \nabla_{\vlambda} \log \ell \left(\mathcal{T}_{\vlambda}\left(\rvvu\right)\right) - \nabla_{\vlambda} \log q_{\vnu}\left(\mathcal{T}_{\vlambda}\left(\rvvu\right)\right) \Big\lvert_{\vnu = \vlambda}
\]
}%
is given by stopping the gradient from propagating through \(\log q_{\vlambda}\).
\end{definition}
Notice that, the gradient of \(\log q\) is ``stopped'' by the assignment \(\vnu = \vlambda\).
This creates a control variate effect, where the control variate \(\mathrm{cv}\) is implicitly formed as
{%
\setlength{\belowdisplayskip}{1.ex} \setlength{\belowdisplayshortskip}{1.ex}
\setlength{\abovedisplayskip}{1.ex} \setlength{\abovedisplayshortskip}{1.ex}
\[
  \mathrm{cv}\left(\vlambda; \vu\right)
  =
  \nabla_{\vlambda} \mathbb{H}\left(\vlambda\right)
  +
  \nabla_{\vlambda} \log q_{\vnu}\left(\mathcal{T}_{\vlambda}\left(\vu\right)\right) \Big\lvert_{\vnu = \vlambda}.
\]
}%
Subtracting this to the CFE estimator leads to the STL estimator.

\subsection{Quadratic Variance Condition}
The convergence of BBVI has recently been established concurrently by \citet{kim_convergence_2023,domke_provable_2023}.
However, the analysis of \citeauthor{domke_provable_2023} presents a broadly applicable framework based on the \textit{quadratic variance} (QV) condition.
\begin{definition}[\textbf{Quadratic Variance; QV}]
  A gradient estimator \(\rvvg\) is said to satisfy the quadratic variance condition if the following bound holds:
{%
\setlength{\belowdisplayskip}{1.ex} \setlength{\belowdisplayshortskip}{1.ex}
\setlength{\abovedisplayskip}{1.ex} \setlength{\abovedisplayshortskip}{1.ex}
  \[
    \mathbb{E}\norm{\rvvg\left(\vlambda\right)}_2^2 \leq \alpha \norm{\vlambda - \vlambda^*}_2^2 + \beta,
  \] 
  }%
  for any \(\vlambda \in \Lambda_S\) and some \(0 \leq \alpha, \beta < \infty\), where \(\vlambda^*\) is a stationary point.
\end{definition}
This basically assumes that the gradient variance grows no more than a quadratic plus a constant.
For non-asymptotic analysis of SGD, this bound was first used by \citet{moulines_nonasymptotic_2011} as an intermediate step implied by the condition:
{%
\setlength{\belowdisplayskip}{1.5ex} \setlength{\belowdisplayshortskip}{1.5ex}
\setlength{\abovedisplayskip}{1.ex} \setlength{\abovedisplayshortskip}{1.ex}
\[
    \mathbb{E}\norm{ \rvvg\left(\vlambda\right) - \rvvg\left(\vlambda'\right) }_2^2 \leq \mathcal{L} \norm{\vlambda - \vlambda'}_2^2,
\]
}%
for all \(\vlambda, \vlambda' \in \Lambda\) and some \(0 < \mathcal{L} < \infty\).
This, combined with the assumption \(\mathbb{E}\norm{\rvvg\left(\vlambda^*\right)}_2^2 < \infty\), implies the QV condition.
They used this strategy to prove the convergence of SGD on strongly convex functions.
Later on,~\citet[p. 85]{wright_optimization_2021} directly assumed the QV condition to obtain similar results.
A comprehensive convergence analysis of projected and proximal SGD with the QV condition was conducted by \citet{domke_provable_2023}, where they also prove the convergence on general convex functions.
This work will invoke the analysis of \citeauthor{domke_provable_2023} by establishing the QV condition of the considered gradient estimators.

\subsection{Interpolation Condition}
To establish the linear, or more intuitively ``exponential,'' convergence of SGD, \citet{schmidt_fast_2013} have relied on the interpolation condition:
\begin{definition}[\textbf{Interpolation}]
  A gradient estimator \(\rvvg\) is said to satisfy the interpolation condition if  \(\mathbb{E}\norm{\rvvg\left(\vlambda^*\right)}^2 = 0\) for \(\vlambda^* \in \Lambda\) such that \(\norm{\nabla F\left(\vlambda^*\right)} = 0\).
\end{definition}
This assumes that the gradient variance vanishes at a stationary point, gradually retrieving the convergence behavior of deterministic gradient descent.
For the QV condition, this corresponds to \(\beta = 0\).

\vspace{-1ex}
\paragraph{Achieving ``Interpolation''}
Currently, there are two ways where the interpolation condition can be achieved.
The first case is when interpolation is achieved \textit{naturally}.
That is, in ERM, when the model is so overparameterized that certain parameters can ``interpolate'' all of the data points in the train data~\citep{ma_power_2018,vaswani_fast_2019}, the gradient becomes 0.
Otherwise, a control variate approach such as stochastic average gradient (SAG; \citealp{schmidt_minimizing_2017}) or stochastic variance-reduced gradient (SVRG; \citealp{johnson_accelerating_2013}), and their many variants~\citep{gower_variancereduced_2020} can be used.

\vspace{-1ex}
\paragraph{Does STL ``Interpolate?''}
As we previously discussed, the STL estimator is essentially a control variate method.
Thus, an important question is whether it can achieve the same effect, notably linear convergence, as variance-reduced SGD methods. 
While \citet{roeder_sticking_2017} have already shown that the STL estimator achieves interpolation when \(q_{\vlambda^*} = \pi\), our research question is whether this fact can be rigorously used to establish linear convergence of SGD.

\section{MAIN RESULTS}
\vspace{-1ex}
\subsection{Triangular Scale Parameterization}\label{section:triangular_scale}
\vspace{-1ex}

First, we will demonstrate a parameterization that is more computationally efficient than the matrix square-root parameterization considered in \cref{section:scale_parameterization}, while satisfying the constraints \labelcref{item:positive_definite,item:linearity,item:convexity,item:smooth_entropy}.
We first turn our attention to the following domain for the variational parameters:
{%
\setlength{\belowdisplayskip}{1ex} \setlength{\belowdisplayshortskip}{1ex}
\setlength{\abovedisplayskip}{1ex} \setlength{\abovedisplayshortskip}{1ex}
\begin{align*}
  \Lambda_{S} \triangleq
  \left\{\, \left(\vm, \mC\right) \in \mathbb{R}^{d} \times \mathbb{L}_{++}^d \mid \sigma_{\mathrm{min}}\left(\mC\right) \geq
   1/\sqrt{S} \,\right\},
\end{align*}
}%
where \(\mathbb{L}_{++}^d\) is the set of Cholesky factors.
A key special case is the mean-field variational family, which is a strict subset of \(\Lambda_S\), where we restrict \(\mC\) to be diagonal matrices.
With that said, we consider the two following parameterizations:
{
\setlength{\belowdisplayskip}{1.ex} \setlength{\belowdisplayshortskip}{1.ex}
\setlength{\abovedisplayskip}{1.ex} \setlength{\abovedisplayshortskip}{1.ex}
\begin{alignat*}{4}
  \mC &= \mL, 
  &&\qquad
  \text{(full-rank)}
  &&\qquad
  \\
  \mC &= \mathrm{diag}\left(L_{11}, \ldots, L_{dd}\right),
  &&\qquad
  \text{(mean-field)}
\end{alignat*}
}%
where \(\mL\) is a Cholesky factor.
In practice, the triangular matrix parameterization is most commonly used~\citep{kucukelbir_automatic_2017}, and results in lower gradient variance than the square root parameterization \citep{kim_practical_2023}.

\vspace{-1ex}
\paragraph{Projection Operator}
The key advantage of operating with triangular scale matrices is that the entropy is the log-sum of their eigenvalues, which turns out to be their diagonal elements.
This implies that the gradient of the entropy term \(\vlambda \mapsto \mathbb{H}\left(q_{\vlambda}\right)\) only resides on the diagonal subspace of \(\mC\).
Therefore, the ``smoothness'' of \(\vlambda \mapsto \mathbb{H}\left(q_{\vlambda}\right)\) can be achieved by only controlling the eigenvalues (or diagonal elements) of \(\mC\).
This sharply contrasts with the square-root parameterization where the constraint is on the \textit{singular values}, which are much harder to control.
Nevertheless, the canonical Euclidean projection operator is:
\begin{proposition}
    The Euclidean projection operator onto \(\Lambda_S\), 
    \(\mathrm{proj}_{\Lambda_S} : \mathbb{R}^d \times \mathbb{L}^d \to \Lambda_{S} \), is given as
{%
\setlength{\belowdisplayskip}{1ex} \setlength{\belowdisplayshortskip}{1ex}
\setlength{\abovedisplayskip}{1ex} \setlength{\abovedisplayshortskip}{1ex}
    \begin{align*}
       \mathrm{proj}_{\Lambda_S}\left(\vlambda\right) 
       = \argmin_{\vlambda' \in \Lambda_S} \norm{\vlambda - \vlambda'}_2^2
       = \left(\vm, \widetilde{\mC}\right),
    \end{align*}
}
    where \(\widetilde{\mC}\) is the projection of \(\mC\) such that
{%
\setlength{\belowdisplayskip}{1.5ex} \setlength{\belowdisplayshortskip}{1.5ex}
\setlength{\abovedisplayskip}{1.5ex} \setlength{\abovedisplayshortskip}{1.5ex}
    \[
      \widetilde{C}_{ij} = \begin{cases}
        \; \max\left(C_{ii}, \; 1/\sqrt{S}\right) &\text{for } i = j \\
        \; C_{ij} &\text{for } i \neq j.
      \end{cases}
    \]
}%
\end{proposition}
\vspace{-2ex}
\begin{proof}
    Since the eigenvalues of a triangular matrix are its diagonal elements, we notice that \(\Lambda_{S}\) is a constraint only on the diagonal elements of \(\mC\) such that \(C_{ii} \geq 1/\sqrt{S}\).
    Conveniently, this is an element-wise half-space constraint for which the projection follows as
{%
\setlength{\belowdisplayskip}{0ex} \setlength{\belowdisplayshortskip}{0ex}
\setlength{\abovedisplayskip}{1.5ex} \setlength{\abovedisplayshortskip}{1.5ex}
    \[
       \widetilde{C}_{ii} 
       = \argmin_{c \geq 1/\sqrt{S}} \norm{C_{ii} - c}_2^2
       = \max\left(C_{ii}, 1/\sqrt{S}\right).
    \]
}%
\vspace{-1ex}
\end{proof}

\vspace{-1ex}
\paragraph{Theoretical Properties}
We will now prove that our construction is valid.
It is trivial that \labelcref{item:positive_definite,item:linearity,item:convexity} are satisfied.
We formally prove that \(\Lambda_S\) satisfies \labelcref{item:smooth_entropy}:

\begin{theoremEnd}[category=entropysmooth]{proposition}\label{thm:entropy_smoothness}
    The entropy \(\vlambda \mapsto \mathbb{H}\left(q_{\vlambda}\right)\) is \(S\)-Lipschitz smooth on \(\Lambda_S\).
\end{theoremEnd}
\begin{proofEnd}
    From the definition of the entropy of location-scale variational families, we have
    \begin{align*}
        \norm{ 
          \nabla_{\vlambda} \mathbb{H}\left(q_{\vlambda}\right) - \nabla_{\vlambda'} \mathbb{H}\left(q_{\vlambda'}\right) 
        }_2^2
        &=
        \norm{ 
          \nabla_{\mC} \log \mathrm{det} \,\mC - \nabla_{\mC'} \log \mathrm{det} \, \mC'
        }_2^2,
\shortintertext{since \(\mC, \mC' \in \mathbb{L}^d_{++}\),} 
        &=
        \norm{ 
          \nabla_{\mC} \log \mathrm{det} \left(\mathrm{diag}\left(\mC\right)\right) - \nabla_{\mC'} \log \mathrm{det} \left(\mathrm{diag}\left(\mC'\right)\right)
        }_2^2,
\shortintertext{since the determinant of triangular matrices is the product of the diagonal,} 
        &=
        \sum^d_{i=1} \abs{ \frac{\partial \log C_{ii}}{\partial C_{ii}} - \frac{\partial \log C_{ii}'}{\partial C_{ii}'}}^2 
        \\
        &=
        \sum^d_{i=1} \abs{ \frac{1}{C_{ii}} - \frac{1}{C_{ii}'}}^2 
        \\
        &=
        \sum^d_{i=1} C_{ii}^{-2} \abs{ C_{ii}' - C_{ii} }^2 {\left(C_{ii}'\right)}^{-2},
\shortintertext{and since \(\sigma_{\mathrm{min}}\left(\mC\right) \geq S^{\nicefrac{-1}{2}} \Leftrightarrow C_{ii}^{-2} \leq S\) for all \(i=1, \ldots, d\),} 
        &\leq
        S^2 \sum^d_{i=1} \abs{ C_{ii}' - C_{ii} }^2
        \\
        &=
        S^2 \norm{ \mathrm{diag}\left(\mC\right) - \mathrm{diag}\left(\mC'\right) }^2_2
        \\
        &\leq
        S^2 \norm{ \vlambda - \vlambda' }^2_2.
    \end{align*}
\end{proofEnd}

Given these results, we will hereafter assume projected SGD is run on \(\Lambda_S\) with the projection operator \(\mathrm{proj}_{\Lambda_S}\).

\subsection{Theoretical Analysis of the STL Estimator}\label{section:gradient_variance}
Before presenting our analysis on BBVI gradient estimators, we will discuss a notable aspect of our strategy and the key step in our proof.

Our main assumption on the target posterior is that it is \(L\)-log(-Lipschitz) smooth:
\begin{definition}
    \(\pi\) is said to be \(L\)-log-(Lipschitz) smooth if its log-density \(\log \pi : \mathbb{R}^d \to \mathbb{R}\) is \(L\)-Lipschitz smooth such that
    \[
      \norm{ \nabla \log \pi\left(\vz\right) - \nabla \log \pi\left(\vz'\right) }_2 \leq L \norm{\vz - \vz'}_2,
    \]
    for all \(\vz, \vz' \in \mathbb{R}^d\) and some \(0 < L < \infty\).
\end{definition}
If this holds for \(\pi\), the same bound holds for \(\ell\) as well since they are proportional up to a constant such that \(\nabla \log \ell = \nabla \log \pi\).
This assumption has been used by \citet{domke_provable_2019} to establish similar results for the CFE estimator and is also widely used in the analysis of sampling algorithms based on log-concave analysis. 
(See \citet[\S 2.3]{dwivedi_logconcave_2019} for such example.)
For probability measures, log-smoothness implies that the density of \(\pi\) can be upper bounded by some Gaussian.
Naturally, this essentially corresponds to assuming \(\pi\) has sub-Gaussian tails.

\vspace{-1ex}
\paragraph{Adaptive Bounds with the Peter-Paul Inequality}
Unlike the QV bounds obtained by \citet{domke_provable_2023}, our bounds involve a free parameter \(\delta \geq 0\).
We call these bounds \textit{adaptive} QV bounds.
\begin{assumption}[\textbf{Adaptive QV}]\label{assumption:adaptiveqvc}
  The gradient estimator \(\rvvg\) satisfies the bound
{%
\setlength{\belowdisplayskip}{1.ex} \setlength{\belowdisplayshortskip}{1.ex}
\setlength{\abovedisplayskip}{1.ex} \setlength{\abovedisplayshortskip}{1.ex}
  \[
    \mathbb{E}\norm{\rvvg\left(\vlambda\right)}_2^2 \leq (1 + C \delta) \, \widetilde{\alpha} \, \norm{\vlambda - \vlambda^*}_2^2 + (1 + C^{-1} \delta^{-1}) \,\widetilde{\beta},
  \]
  }%
  for any \(\delta > 0\), any \(\vlambda \in \Lambda_S\), and some \(0 < \widetilde{\alpha}, \widetilde{\beta} < \infty\), where \(\vlambda^*\) is a stationary point.
\end{assumption}
This is a consequence of the use of the ``Peter-Paul'' inequality such that
{%
\setlength{\belowdisplayskip}{1.ex} \setlength{\belowdisplayshortskip}{1.ex}
\setlength{\abovedisplayskip}{1.ex} \setlength{\abovedisplayshortskip}{1.ex}
\begin{equation}
  {\left( a + b \right)}^2 \leq \left(1 + \delta\right) \,a^2 + (1 + \delta^{-1}) \,b^2,
  \label{eq:peterpaul}
\end{equation}
}%
and can be seen as a generalization of the usual inequality \({\left(a + b\right)}^2 \leq 2 a^2 + 2 b^2\).
Adjusting \(\delta\) can occasionally tighten the analysis.
In fact, \(\delta\) can be optimized to become \textit{adaptive} to the downstream analysis.
Indeed, in our complexity analysis, \(\delta\) automatically trades-off the influence of \(\widetilde{\alpha}\) and \(\widetilde{\beta}\) according to the accuracy budget \(\epsilon\).

\vspace{-1ex}
\paragraph{Key Lemma}
The key first step in all of our analysis is the following decomposition:

\begin{theoremEnd}[category=gradvarlemmas,all end]{lemma}\label{thm:peterpaul}
  For any \(a, b, c, \in \mathbb{R}\), 
  \[ {(a+b+c)}^2 \leq (2 + \delta ) a^2 + (2 + \delta ) b^2 + (1 + 2 \delta^{-1})  c^2, \]
  for any \(\delta > 0\).
\end{theoremEnd}
\begin{proofEnd}
The Peter-Paul generalization of Young's inequality states that, for \(d, e \geq 0\), we have
\[
   d e \leq \frac{\delta }{2} d^2 + \frac{\delta^{-1}}{2} e^2.
\]
Applying this,
\begin{align*}
  {(a+b+c)}^2
  &\;=
  a^2 + b^2 + c^2 + 2 a b + 2 a c + 2 b c
  \\
  &\;\leq
    a^2 + b^2 + c^2 + 2 \abs{a} \abs{b} + 2 \abs{a} \abs{c} + 2 \abs{b} \abs{c}
  \\
  &\;\leq
  a^2 + b^2 + c^2 
  + 2 \left( \frac{1}{2} a^2 + \frac{1}{2} b^2 \right)
  + 2 \left( \frac{\delta}{2} a^2 + \frac{\delta^{-1}}{2} c^2 \right)
  + 2 \left( \frac{\delta}{2} b^2 + \frac{\delta^{-1}}{2} c^2 \right) 
  \\
  &\;=
  a^2 + b^2 + c^2 + \left( a^2 +  b^2 \right)
  + \left( \delta a^2 + \delta^{-1} c^2 \right)
  + \left( \delta b^2 + \delta^{-1} c^2 \right) 
  \\
  &\;=
  \left(2 + \delta\right) a^2 + \left(2 + \delta\right) b^2 + \left(1 + 2 \delta^{-1}\right) c^2.
\end{align*}
\end{proofEnd}

\begin{theoremEnd}[category=stlupperboundlemma]{lemma}\label{thm:stl_decomposition}
{\setlength{\belowdisplayskip}{1.5ex} \setlength{\belowdisplayshortskip}{1.5ex}
\setlength{\abovedisplayskip}{1.ex} \setlength{\abovedisplayshortskip}{1.ex}
  Assume \cref{assumption:variation_family}.
  The expected-squared norm of STL is bounded as
  {\small
  \begin{align*}
    \mathbb{E} \norm{\rvvg_{\mathrm{STL}}\left(\vlambda\right)}_2^2
    \leq
    \left( 2 + \delta \right) V_{1}
    +
    \left( 2 + \delta \right) V_{2}
    +
    \left( 1 + 2 \delta^{-1} \right) V_{3},
  \end{align*}
  }
  where the terms are
  {\small
  \begin{align*}
    V_{1}
    &=
    \mathbb{E} \,
    J_{\mathcal{T}}\left(\rvvu\right) 
    \lVert
    \nabla \log \ell \left(\mathcal{T}_{\vlambda}\left(\rvvu\right)\right) 
    -
    \nabla \log \ell \left(\mathcal{T}_{\vlambda^*}\left(\rvvu\right)\right) 
    {\rVert}_2^2
    \\
    V_{2}
    &=
    \mathbb{E} \,
    J_{\mathcal{T}}\left(\rvvu\right)
    \lVert
    \nabla \log q_{\vlambda^*}\left(\mathcal{T}_{\vlambda^*}\left(\rvvu\right)\right)
    - 
    \nabla \log q_{\vlambda}\left(\mathcal{T}_{\vlambda}\left(\rvvu\right)\right)
    {\rVert}_2^2
    \\
    V_{3}
    &=
    \mathbb{E} \,
    J_{\mathcal{T}}\left(\rvvu\right)
    \lVert
    \nabla \log \ell\left(\mathcal{T}_{\vlambda^*}\left(\rvvu\right)\right)
    - 
    \nabla \log q_{\vlambda^*}\left(\mathcal{T}_{\vlambda^*}\left(\rvvu\right)\right)
    {\rVert}_2^2,
  \end{align*}
  }
  for any \(\delta > 0\) and \(\vlambda \in \mathbb{R}^p\).
  \(J_{\mathcal{T}} : \mathbb{R}^d \to \mathbb{R}\) is a function depending on the variational family as
  \begin{alignat*}{2}
    J_{\mathcal{T}}\left(\rvvu\right) &= 1 + {\textstyle\sum^{d}_{i=1} \rvu_i^2} \quad                        &&\quad \text{for full-rank and} \\
    J_{\mathcal{T}}\left(\rvvu\right) &= 1 + {\textstyle\sqrt{\sum^{d}_{i=1} \rvu_i^4}} &&\quad \text{for mean-field.}
  \end{alignat*}
 }%
\end{theoremEnd}
\begin{proofEnd}
From the definition of the STL estimator~\cref{def:stl},
\begin{align*}
    \mathbb{E} \norm{\rvvg_{\mathrm{STL}}\left(\vlambda\right)}_2^2
    &=
    \mathbb{E} \norm{
    \nabla_{\vlambda} \log \ell \left(\mathcal{T}_{\vlambda}\left(\rvvu\right)\right) - \nabla_{\vlambda} \log q_{\vnu}\left(\mathcal{T}_{\vlambda}\left(\rvvu\right)\right)
    }_2^2
    \;\Big\lvert_{\vnu = \vlambda},
\shortintertext{by \cref{thm:jacobian_reparam_inner},}
    &=
    \mathbb{E}
    J_{\mathcal{T}}\left(\rvvu\right)
    \norm{
    \nabla \log \ell \left(\mathcal{T}_{\vlambda}\left(\rvvu\right)\right) 
    - \nabla \log q_{\vnu}\left(\mathcal{T}_{\vlambda}\left(\rvvu\right)\right)
    }_2^2
    \;\Big\lvert_{\vnu = \vlambda}
\shortintertext{adding the terms \(\nabla \log \ell \left(\mathcal{T}_{\vlambda^*}\left(\rvvu\right)\right)\) and \(\nabla \log q_{\vlambda^*}\left(\mathcal{T}_{\vlambda^*}\left(\rvvu\right)\right)\) that cancel,}
    &=
    \mathbb{E}
    J_{\mathcal{T}}\left(\rvvu\right)
    \lVert
    \nabla \log \ell \left(\mathcal{T}_{\vlambda}\left(\rvvu\right)\right) 
    -
    \nabla \log \ell \left(\mathcal{T}_{\vlambda^*}\left(\rvvu\right)\right) 
    \\
    &\quad\qquad\qquad+
    \nabla \log \ell\left(\mathcal{T}_{\vlambda^*}\left(\rvvu\right)\right)
    - 
    \nabla \log q_{\vlambda^*}\left(\mathcal{T}_{\vlambda^*}\left(\rvvu\right)\right)
    \\
    &\quad\qquad\qquad+
    \nabla \log q_{\vlambda^*}\left(\mathcal{T}_{\vlambda^*}\left(\rvvu\right)\right)
    - 
    \nabla \log q_{\vlambda}\left(\mathcal{T}_{\vlambda}\left(\rvvu\right)\right)
    {\rVert}_2^2,
\shortintertext{applying \cref{thm:peterpaul},}
    &\leq
    \mathbb{E}
    J_{\mathcal{T}}\left(\rvvu\right)
    \Big(
    \left( 2 + \delta \right)
    \lVert
    \nabla \log \ell \left(\mathcal{T}_{\vlambda}\left(\rvvu\right)\right) 
    -
    \nabla \log \ell \left(\mathcal{T}_{\vlambda^*}\left(\rvvu\right)\right) 
    {\rVert}_2^2
    \\
    &\quad\qquad\qquad+
    \left( 2 + \delta \right)
    \lVert
    \nabla \log q_{\vlambda^*}\left(\mathcal{T}_{\vlambda^*}\left(\rvvu\right)\right)
    - 
    \nabla \log q_{\vlambda}\left(\mathcal{T}_{\vlambda}\left(\rvvu\right)\right)
    {\rVert}_2^2
    \\
    &\quad\qquad\qquad+
    \left(1 + 2 \delta^{-1}\right)
    \lVert
    \nabla \log \ell\left(\mathcal{T}_{\vlambda^*}\left(\rvvu\right)\right)
    - 
    \nabla \log q_{\vlambda^*}\left(\mathcal{T}_{\vlambda^*}\left(\rvvu\right)\right)
    {\rVert}_2^2
    \Big),
\shortintertext{and distributing \(J_{\mathcal{T}}\) and the expectation,}
    &=
    \left( 2 + \delta \right)
    \underbrace{
    \mathbb{E}
    J_{\mathcal{T}}\left(\rvvu\right)
    \lVert
    \nabla \log \ell \left(\mathcal{T}_{\vlambda}\left(\rvvu\right)\right) 
    -
    \nabla \log \ell \left(\mathcal{T}_{\vlambda^*}\left(\rvvu\right)\right) 
    {\rVert}_2^2
    }_{V_{1}}
    \\
    &\quad+
    \left( 2 + \delta \right)  
     \underbrace{
    J_{\mathcal{T}}\left(\rvvu\right)
    \mathbb{E}
    \lVert
    \nabla \log q_{\vlambda^*}\left(\mathcal{T}_{\vlambda^*}\left(\rvvu\right)\right)
    - 
    \nabla \log q_{\vlambda}\left(\mathcal{T}_{\vlambda}\left(\rvvu\right)\right)
    {\rVert}_2^2
    }_{V_{2}}
    \\
    &\quad+
    \left(1 + 2 \delta^{-1}\right)  
    \underbrace{
    \mathbb{E}
    J_{\mathcal{T}}\left(\rvvu\right)
    \lVert
    \nabla \log \ell\left(\mathcal{T}_{\vlambda^*}\left(\rvvu\right)\right)
    - 
    \nabla \log q_{\vlambda^*}\left(\mathcal{T}_{\vlambda^*}\left(\rvvu\right)\right)
    {\rVert}_2^2
    }_{V_{3}}.
\end{align*}
\end{proofEnd}

Here, \(J_{\mathcal{T}}\) is a term that stems from the Jacobian of \(\mathcal{T}\).
Thus, \(J_{\mathcal{T}}\) contains the properties unique to the chosen variational family.
\(V_{1}\) and \(V_{2}\) measure how far the current variational approximation \(q_{\vlambda}\) is from a stationary point \({\vlambda^*}\).
Thus, both terms will eventually reach 0 as BBVI converges, regardless of family specification.
The key is \(V_{3}\), which captures the amount of mismatch between the score of the true posterior \(\pi\) and variational posterior \(q_{\vlambda^*}\).
Establishing the ``interpolation condition'' amounts to analyzing when \(V_{3}\) becomes 0.

\subsubsection{Upper Bounds}
We now present our upper bound on the expected-squared norm of the STL gradient estimator.

\begin{theoremEnd}[category=stlupperboundfr]{theorem}\label{thm:stl_upperbound}
Assume \cref{assumption:variation_family} and that \(\ell\) is \(L\)-log-smooth.
For the full-rank parameterization, the expected-squared norm of the STL estimator is bounded as
{\setlength{\belowdisplayskip}{1.5ex} \setlength{\belowdisplayshortskip}{1.5ex}
\setlength{\abovedisplayskip}{1.5ex} \setlength{\abovedisplayshortskip}{1.5ex}
\begin{align*}
  \mathbb{E} \norm{\rvvg_{\mathrm{STL}}\left(\vlambda\right)}_2^2
  \leq
  \alpha_{\mathrm{STL}} \norm{\vlambda - \vlambda^*}_2^2 + \beta_{\mathrm{STL}}
\end{align*}
}%
where 
{\setlength{\belowdisplayskip}{1.5ex} \setlength{\belowdisplayshortskip}{1.5ex}
\setlength{\abovedisplayskip}{1.5ex} \setlength{\abovedisplayshortskip}{1.5ex}
\begin{align*}
  \alpha_{\mathrm{STL}}
  &=
  (2 + \delta) \left(
  L^2 \, \left(d + k_{\varphi}\right) 
  + 
  S^{2} \left(d + 1\right) \right) 
  \\
  \beta_{\mathrm{STL}}
  &=
  (1 + 2 \delta^{-1}) \left( 2 d + k_{\varphi} \right) \sqrt{\mathrm{D}_{\mathrm{F}^4}\left(q_{\vlambda^*}, \ell\right)}
\end{align*}
}%
for any \(\vlambda, \vlambda^* \in \Lambda_{S}\) and any \(\delta > 0\).
\end{theoremEnd}
\begin{proofEnd}\label{proof:stl_upperbound}
We analyze each of the terms in \cref{thm:stl_decomposition}.

\paragraph{Bound on \(V_{1}\)}
For \(V_{1}\), we obtain the quadratic bound from the optimum as
\begin{alignat}{3}
  V_{1} &= \mathbb{E}
    J_{\mathcal{T}}\left(\rvvu\right)
    \norm{
    \nabla \log \ell \left(\mathcal{T}_{\vlambda}\left(\rvvu\right)\right) 
    -
    \nabla \log \ell \left(\mathcal{T}_{\vlambda^*}\left(\rvvu\right)\right) 
    }_2^2
    \nonumber
    \\
    &\leq
    L^2 \, \mathbb{E} J_{\mathcal{T}}\left(\rvvu\right)
    \norm{ 
    \mathcal{T}_{\vlambda}\left(\rvvu\right) - 
    \mathcal{T}_{\vlambda^*}\left(\rvvu\right) 
    }_2^2
    &&\qquad\text{(\(L\)-log-smoothness)}
    \label{eq:stl_upperbound_t1_inter}
    \\
    &\leq
    L^2 \left(d + k_{\varphi}\right) \norm{\vlambda - \vlambda^*}_2^2.
    &&\qquad\text{(\cref{thm:u_normsquared_marginalization})}
    \label{eq:stl_upperbound_t181}
\end{alignat}

\paragraph{Bound on \(V_{2}\)}
Now for,
\begin{align*}
  V_{2}
  =
  \mathbb{E}
  J_{\mathcal{T}}\left(\rvvu\right)
  \norm{
    \nabla \log q_{\vlambda^*}\left(\mathcal{T}_{\vlambda^*}\left(\rvvu\right)\right)
    - 
    \nabla \log q_{\vlambda}\left(\mathcal{T}_{\vlambda}\left(\rvvu\right)\right)
  }_2^2,
\end{align*}
we use the fact that, for location-scale family distributions, the log-probability density is
\[
  \log q_{\vlambda}\left( \vz \right) = \log \varphi\left( \mC^{-1}\left(\vz - \vm\right) \right) - \log \abs{\mC}.
\]
Considering reparameterization, 
\begin{align*}
  \log q_{\vlambda}\left( \mathcal{T}_{\vlambda}\left(\vu\right) \right) 
  &=
  \log \varphi\left( \mC^{-1}\left( \mathcal{T}_{\vlambda}\left(\vu\right)  - \vm\right) \right) - \log \abs{\mC}
  \\
  &=
  \log \varphi\left( \mC^{-1}\left( \left( \mC \vu + \vm \right) - \vm\right) \right) - \log \abs{\mC}
  \\
  &=
  \log \varphi\left( \vu \right) - \log \abs{\mC}.
\end{align*}
This implies
\begin{align*}
  \nabla \log q_{\vlambda}\left( \mathcal{T}_{\vlambda}\left(\vu\right) \right) 
  &=
  \nabla_{\vlambda} \log \phi\left( \vu \right) - \nabla \log \abs{\mC}
  \\
  &=
  - \nabla \log \abs{\mC}
  \\
  &=
  \nabla_{\vlambda} \mathbb{H}\left(q_{\vlambda}\right).
\end{align*}

Thus, 
\begin{alignat}{2}
  V_2
  &=
  \mathbb{E}
  J_{\mathcal{T}}\left(\rvvu\right)
  \norm{
    \nabla \log q_{\vlambda^*}\left(\mathcal{T}_{\vlambda^*}\left(\rvvu\right)\right)
    - 
    \nabla \log q_{\vlambda}\left(\mathcal{T}_{\vlambda}\left(\rvvu\right)\right)
  }_2^2
  \nonumber
  \\
  &\leq
  S^{2} 
  \mathbb{E}
  J_{\mathcal{T}}\left(\rvvu\right)
  \norm{
    \vlambda
    -
    \vlambda^*
  }^2_{2}
  &&\qquad\text{(\cref{thm:entropy_smoothness})}
  \label{eq:stl_upperbound_t2_inter}
  \\
  &=
  S^{2} 
  \left(1 + \mathbb{E} \sum^{d}_{i=1} \rvu_i^2 \right)
  \norm{
    \vlambda
    -
    \vlambda^*
  }^2_{2}
  \nonumber
  &&\qquad\text{(definition of \(J_{\mathcal{T}}\) in \cref{thm:jacobian_reparam_inner})}
  \\
  &=
  S^2
  \left( 
   1 + d
  \right)
  \norm{
    \vlambda
    -
    \vlambda^*
  }^2_{2}
  &&\qquad\text{(\cref{assumption:symmetric_standard})}
  \label{eq:stl_upperbound_t182}
\end{alignat}

\paragraph{Bound on \(V_3\)}
Finally, for \(V_3\), 
\begin{align}
  V_3 
  &=
  \mathbb{E}
  J_{\mathcal{T}}\left(\rvvu\right)
  \lVert
  \nabla \log \ell\left(\mathcal{T}_{\vlambda^*}\left(\rvvu\right)\right)
  - 
  \nabla \log q_{\vlambda^*}\left(\mathcal{T}_{\vlambda^*}\left(\rvvu\right)\right)
  {\rVert}_2^2,
  \nonumber
\shortintertext{by the definition of \(J_{\mathcal{T}}\) in \cref{thm:jacobian_reparam_inner},}
  &=
  \mathbb{E}
  \left(1 + \sum^{d}_{i=1} \rvu_i^2 \right)
  \lVert
  \nabla \log \ell\left(\mathcal{T}_{\vlambda^*}\left(\rvvu\right)\right)
  - 
  \nabla \log q_{\vlambda^*}\left(\mathcal{T}_{\vlambda^*}\left(\rvvu\right)\right)
  {\rVert}_2^2
  \nonumber
  \\
  &=
  \mathbb{E}
  \left(1 + \norm{\rvvu}_2^2 \right)
  \lVert
  \nabla \log \ell\left(\mathcal{T}_{\vlambda^*}\left(\rvvu\right)\right)
  - 
  \nabla \log q_{\vlambda^*}\left(\mathcal{T}_{\vlambda^*}\left(\rvvu\right)\right)
  {\rVert}_2^2,
  \nonumber
\shortintertext{through the Cauchy-Schwarz inequality,}
  &\leq
  \sqrt{
  \mathbb{E}
  {\left(1 + \norm{\rvvu}_2^2 \right)}^2
  }
  \sqrt{
  \mathbb{E}
  \lVert
  \nabla \log \ell\left(\mathcal{T}_{\vlambda^*}\left(\rvvu\right)\right)
  - 
  \nabla \log q_{\vlambda^*}\left(\mathcal{T}_{\vlambda^*}\left(\rvvu\right)\right)
  {\rVert}_2^4
  }
  \nonumber
  \\
  &=
  \sqrt{
    \left( 1 + 2 \mathbb{E} \norm{\rvvu}_2^2 + \mathbb{E} \norm{\rvvu}_2^4 \right)} \,
  \sqrt{
  \mathbb{E}
  \lVert
  \nabla \log \ell\left(\mathcal{T}_{\vlambda^*}\left(\rvvu\right)\right)
  - 
  \nabla \log q_{\vlambda^*}\left(\mathcal{T}_{\vlambda^*}\left(\rvvu\right)\right)
  {\rVert}_2^4
  },
  \nonumber
\shortintertext{by \cref{thm:u_identities,assumption:symmetric_standard}, the \(2 \, \mathbb{E} \norm{\rvvu}_2^2\) term becomes}
  &=
  \sqrt{
     1 + 2 d + \mathbb{E} \norm{\rvvu}_2^4 \,
  }
  \,
  \sqrt{
  \mathbb{E}
  \lVert
  \nabla \log \ell\left(\mathcal{T}_{\vlambda^*}\left(\rvvu\right)\right)
  - 
  \nabla \log q_{\vlambda^*}\left(\mathcal{T}_{\vlambda^*}\left(\rvvu\right)\right)
  {\rVert}_2^4
  }.
  \label{eq:stl_upperbound_t183}
\end{align}

Meanwhile, \(\mathbb{E} \norm{\rvvu}_2^4\) follows as
\begin{align*}
  \mathbb{E}
  \norm{\rvvu}_2^4 
  &=
  \mathbb{E}
  {\left( \norm{\rvvu}_2^2 \right)}^2
  =
  \mathbb{E}
  {\left( \sum^{d}_{i=1} \rvu_i^2 \right)}^2
  =
  \mathbb{E}
  \left(
  \sum^{d}_{i=1} \rvu_i^4
  +
  \sum_{i \neq j} \rvu_i^2 \rvu_j^2
  \right),
\shortintertext{while from \cref{assumption:symmetric_standard}, \(u_i\) and \(u_j\) are independent for \(i \neq j\). Thus}
  &=
  \sum^{d}_{i=1} \mathbb{E} \rvu_i^4
  +
  \sum_{i \neq j} \mathbb{E}\rvu_i^2 \, \mathbb{E}\rvu_j^2,
\shortintertext{and by \cref{assumption:symmetric_standard}, we have \(\mathbb{E} \rvu_i^4 = k_{\varphi}\), \(\mathbb{E}\rvu_i^2 = 1\), and \(\mathbb{E}\rvu_j^2 = 1\). Therefore,}
  &=
  d k_{\varphi}
  +
  2 {d \choose 2},
\shortintertext{and applying the well-known upper bound on the binomial coefficient \({d \choose 2} \leq {\left(\frac{\mathrm{e} d}{2}\right)}^2\),}
  &\leq
  d k_{\varphi}
  +
  2 \, {\left(\frac{\mathrm{e}}{2} d\right)}^2
  =
  d k_{\varphi}
  +
  \frac{\mathrm{e}^2}{2} d^2,
\end{align*}
where \(\mathrm{e}\) is Euler's constant.

Applying this to first term in \cref{eq:stl_upperbound_t183},
\begin{align}
  \sqrt{
    1 + 2 d + \mathbb{E}\norm{\rvvu}_2^4
  }
  &\leq
  \sqrt{
    1 + 2 d + k_{\varphi} d +  \frac{\mathrm{e}^2}{2} d^2
  }
  =
  \sqrt{
    \frac{\mathrm{e}^2}{2} d^2 
    + 
    \left( 2 +  k_{\varphi}\right) d
    + 1
  },
  \nonumber
\shortintertext{and since \(k_{\varphi} \geq 1\), \quad \(d \geq 1\), and \quad \(\mathrm{e}^2/2 \leq 4 \),}
  &\leq
  \sqrt{
    4 d^2 
    + 
    \left( 2 k_{\varphi} +  k_{\varphi}\right) d
    + 
    k_{\varphi}
  }
  =
  \sqrt{
    4 d^2 
    + 
    3 k_{\varphi} d
    + k_{\varphi}^2
  }
  \nonumber
  \\
  &\leq
  \sqrt{
    4 d^2 
    + 
    4 d k_{\varphi}
    + 
    k_{\varphi}^2
  }
  \nonumber
  \\
  &=
  \left( 2 d + k_{\varphi} \right)
  \label{eq:stl_upperbound_t183_eq2}
\end{align}

Thus, \(V_3\) can be bounded as
\begin{align}
  V_3
  &\leq
  \sqrt{
     1 + 2 d + \mathbb{E} \norm{\rvvu}_2^4 \,
  }
  \sqrt{
  \mathbb{E}
  \lVert
  \nabla \log \ell\left(\mathcal{T}_{\vlambda^*}\left(\rvvu\right)\right)
  - 
  \nabla \log q_{\vlambda^*}\left(\mathcal{T}_{\vlambda^*}\left(\rvvu\right)\right)
  {\rVert}_2^4
  },
  \nonumber
\shortintertext{applying \cref{eq:stl_upperbound_t183_eq2},}
  &\leq
  \left( 2 d + k_{\varphi}\right)
  \sqrt{
  \mathbb{E}
  \lVert
  \nabla \log \ell\left(\mathcal{T}_{\vlambda^*}\left(\rvvu\right)\right)
  - 
  \nabla \log q_{\vlambda^*}\left(\mathcal{T}_{\vlambda^*}\left(\rvvu\right)\right)
  {\rVert}_2^4
  }
  \nonumber
\shortintertext{applying Change-of-Variable on the score term,}
  &=
  \left( 2 d + k_{\varphi}\right)
  \sqrt{
  \mathbb{E}_{\rvvz \sim q_{\vlambda^*}}
  \lVert
  \nabla \log \ell\left( \rvvz \right)
  - 
  \nabla \log q_{\vlambda^*}\left( \rvvz \right)
  {\rVert}_2^4
  }
  \nonumber
  \\
  &=
  \left( 2 d + k_{\varphi}\right)
  \sqrt{
  \mathbb{E}_{\rvvz \sim q_{\vlambda^*}}
  \lVert
  \nabla \log \pi\left( \rvvz \right)
  - 
  \nabla \log q_{\vlambda^*}\left( \rvvz \right)
  {\rVert}_2^4
  },
  \nonumber
\shortintertext{and by the definition of the 4th order Fisher-Hyv\"arinen divergence,}
  &\leq
  \left( 
     2 d + k_{\varphi}
  \right)
  \sqrt{\mathrm{D}_{\mathrm{F}^4}\left(q_{\vlambda^*}, \ell\right)}.
  \label{eq:stl_upperbound_t183_eq3}
\end{align}

Combining \cref{eq:stl_upperbound_t181,eq:stl_upperbound_t182,eq:stl_upperbound_t183_eq3} with \cref{thm:stl_decomposition},
\begin{align*}
  \mathbb{E} \norm{\rvvg_{\mathrm{STL}}\left(\vlambda\right)}_2^2
  &\leq
  (2 + \delta) V_1 + (2 + \delta) V_2 + (1 + 2 \delta^{-1}) V_3
  \\
  &\leq
  L^2 (2 + \delta) \left(d + k_{\varphi}\right) \norm{\vlambda - \vlambda^*}_2^2
  +
  S^{2} (2 + \delta)
  \left( 
   d + 1
  \right)
  \norm{
    \vlambda - \vlambda^*
  }^2_{2}
  \\
  &\qquad+
  (1 + 2 \delta^{-1}) \left( 2 d + k_{\varphi} \right) \sqrt{\mathrm{D}_{\mathrm{F}^4}\left(q_{\vlambda^*}, \ell\right)}
  \\
  &\;=
  (2 + \delta) \left(
  L^2 \, \left(d + k_{\varphi}\right) 
  + 
  S^{2} \left(d + 1\right) \right) \norm{\vlambda - \vlambda^*}_2^2
  \\
  &\qquad+
  (1 + 2 \delta^{-1}) \left( 2 d + k_{\varphi} \right) \sqrt{\mathrm{D}_{\mathrm{F}^4}\left(q_{\vlambda^*}, \ell\right)}.
\end{align*}
\end{proofEnd}

\begin{theoremEnd}[all end, category=stlupperboundmf]{theorem}\label{thm:stl_upperbound_mf}
Assume \cref{assumption:variation_family} and that \(\ell\) is \(L\)-log-smooth.
For the mean-field parameterization, the expected-squared norm of the STL estimator is bounded as
{
\begin{align*}
  \mathbb{E} \norm{\rvvg_{\mathrm{STL}}\left(\vlambda\right)}_2^2
  \leq
  (2 + \delta)
  \left(
    L^2 \, \left( 2 k_{\varphi} \sqrt{d} + 1 \right) 
    + 
    S^{2} \left( \sqrt{ d k_{\varphi} } + 1 \right)
  \right)
  \norm{\vlambda - \vlambda^*}_2^2
  +
  (1 + 2 \delta^{-1}) \left( 1 + \sqrt{ d k_{\varphi} } \right) \sqrt{\mathrm{D}_{\mathrm{F}^4}\left(q_{\vlambda^*}, \ell\right)}
\end{align*}
}%
for any \(\vlambda, \vlambda^* \in \Lambda_S\) and any \(\delta > 0\).
\end{theoremEnd}
\begin{proofEnd}\label{proof:stl_upperbound_mf}
Similarly with \cref{thm:stl_upperbound}, we analyze each term in \cref{thm:stl_decomposition}.

\paragraph{Bound on \(V_1\)}
The process for \(V_1\) is more or less identical to \cref{thm:stl_upperbound}.
Starting from~\cref{eq:stl_upperbound_t1_inter},
\begin{alignat}{3}
  V_1
  &\leq
  L^2 \, 
  \mathbb{E} J_{\mathcal{T}}\left(\rvvu\right)
  \norm{ 
  \mathcal{T}_{\vlambda}\left(\rvvu\right) - 
  \mathcal{T}_{\vlambda^*}\left(\rvvu\right) 
  }_2^2,
  \nonumber
  \\
  &\leq
  \left( 2 k_{\varphi} \sqrt{d} + 1 \right) \norm{\vlambda - \vlambda^*}_2^2.
  \label{eq:stl_upperbound_mf_182}
  &&\qquad\text{(\cref{thm:u_normsquared_marginalization})}
\end{alignat}

\paragraph{Bound on \(V_2\)}
This is also identical to \cref{thm:stl_upperbound} apart from \(J_{\mathcal{T}}\).
Resuming from~\cref{eq:stl_upperbound_t2_inter},
\begin{alignat}{2}
  V_2
  &\leq
  S^{2} 
  \mathbb{E} 
  J_{\mathcal{T}}
  \norm{
    \vlambda
    -
    \vlambda^*
  }^2_{2}
  \nonumber
  \\
  &=
  S^{2} 
  \left( 
   1 + \mathbb{E} \sqrt{ \sum_{i=1}^d \rvu_i^4 }
  \right)
  \norm{
    \vlambda
    -
    \vlambda^*
  }^2_{2},
  \nonumber
  &&\qquad\text{(definition of \(J_{\mathcal{T}}\) in \cref{thm:jacobian_reparam_inner})}
  \\
  &\leq
  S^{2} 
  \left( 
   1 + \sqrt{ \sum_{i=1}^d \mathbb{E} \rvu_i^4 }
  \right)
  \norm{
    \vlambda
    -
    \vlambda^*
  }^2_{2},
  \nonumber
  &&\qquad\text{(Jensen's inequality)}
  \\
  &\leq
  S^{2} 
  \left( 
   1 + \sqrt{ d k_{\varphi} }
  \right)
  \norm{
    {\vlambda}
    -
    {\vlambda^*}
  }^2_{2}.
  \label{eq:stl_upperbound_mf_183}
  &&\qquad\text{(\cref{assumption:symmetric_standard})}.
\end{alignat}

\paragraph{Bound on \(V_3\)}
The derivation for \(V_3\) is less technical than the full-rank case.
Denoting \(\rvmU = \mathrm{diag}\left(\rvu_1, \ldots, \rvu_d\right)\) for clarity, we have
\begin{equation}
  \textstyle{\sqrt{\sum_{i=1}^d \rvu_i^4 } = {\lVert \rvmU^2 \rVert}_{\mathrm{F}}}.
  \label{eq:stl_upperbound_meanfield_J}
\end{equation}
Then,
\begin{align*}
  V_3
  &=
  \mathbb{E}
  J_{\mathcal{T}}\left(\rvvu\right)
  \lVert
  \nabla \log \ell\left(\mathcal{T}_{\vlambda^*}\left(\rvvu\right)\right)
  - 
  \nabla \log q_{\vlambda^*}\left(\mathcal{T}_{\vlambda^*}\left(\rvvu\right)\right)
  {\rVert}_2^2
\shortintertext{by the definition of \(J_{\mathcal{T}}\) in \cref{thm:jacobian_reparam_inner} and \cref{eq:stl_upperbound_meanfield_J},}
  &=
  \mathbb{E}
  \left( 1 + {\lVert \rvmU^2 \rVert}_{\mathrm{F}} \right)
  \lVert
  \nabla \log \ell\left(\mathcal{T}_{\vlambda^*}\left(\rvvu\right)\right)
  - 
  \nabla \log q_{\vlambda^*}\left(\mathcal{T}_{\vlambda^*}\left(\rvvu\right)\right)
  {\rVert}_2^2,
\shortintertext{through the Cauchy-Schwarz inequality,}
  &\leq
  \underbrace{
  \sqrt{
  \mathbb{E}
  {\left( 1 + 2 \norm{\rvmU^2}_{\mathrm{F}} + \norm{\rvmU^2}_{\mathrm{F}}^2 \right)}
  }
  }_{T_{\text{\ding{172}}}}
  \sqrt{
  \mathbb{E}
  \lVert
  \nabla \log \ell\left(\mathcal{T}_{\vlambda^*}\left(\rvvu\right)\right)
  - 
  \nabla \log q_{\vlambda^*}\left(\mathcal{T}_{\vlambda^*}\left(\rvvu\right)\right)
  {\rVert}_2^4
  }.
\end{align*}

\(T_{\text{\ding{172}}}\) follows as
\begin{align}
  T_{\text{\ding{172}}}
  &=
  \sqrt{
  \mathbb{E}
  {\left( 1 + 2 {\lVert \rvmU^2 \rVert}_{\mathrm{F}} + {\lVert \rvmU^2 \rVert}_{\mathrm{F}}^2 \right)}
  }
  \nonumber
  \\
  &=
\textstyle{
  \sqrt{
  \mathbb{E}
  {\left( 1 + 2 \sqrt{ \sum_{i=1}^d \rvu_i^4 } + \sum_{i=1}^d \rvu_i^4 \right)}
  },
}
  \nonumber
\shortintertext{distributing the expectation,}
  &=
\textstyle{
  \sqrt{
  1 + 2 \mathbb{E} \left( \sqrt{ \sum_{i=1}^d \rvu_i^4 } \right) + \sum_{i=1}^d \mathbb{E} \rvu_i^4
  },
}
  \nonumber
\shortintertext{applying Jensen's inequality to the middle term,}
  &\leq
\textstyle{
  \sqrt{
  1 + 2 \sqrt{ \sum_{i=1}^d \mathbb{E} \rvu_i^4 } + \sum_{i=1}^d \mathbb{E} \rvu_i^4 
  },
}
  \nonumber
\shortintertext{and from \cref{assumption:symmetric_standard},}
  &=
  \sqrt{
    1 + 2 \sqrt{ d k_{\varphi} } + d k_{\varphi}
  }
  =
  \sqrt{
    {\left( 1 + \sqrt{ d k_{\varphi} }\right)}^2
  }
  \nonumber
  \\
  &=
  1 + \sqrt{ d k_{\varphi} }.
  \label{eq:stl_upperbound_mf_184}
\end{align}
As in the proof of \cref{thm:stl_decomposition}, we obtain the 4th order Fisher-Hyv\"arinen divergence after Change-of-Variable.
Combining this fact with \cref{eq:stl_upperbound_mf_182,eq:stl_upperbound_mf_183,eq:stl_upperbound_mf_184,thm:stl_decomposition}, 
\begin{align*}
  \mathbb{E} \norm{\rvvg_{\mathrm{STL}}\left(\vlambda\right)}_2^2
  &\leq
  (2 + \delta) V_1 + (2 + \delta) V_2 + (1 + 2 \delta^{-1}) V_3
  \\
  &\leq
  L^2 (2 + \delta) \left( 2 k_{\varphi} \sqrt{d} + 1 \right) \norm{\vlambda - \vlambda^*}_2^2
  +
  S^{2} (2 + \delta)
  \left( 
    \sqrt{ d k_{\varphi} } + 1
  \right)
  \norm{
    \vlambda - \vlambda^*
  }^2_{2}
  \\
  &\qquad+
  (1 + 2 \delta^{-1}) \left(\sqrt{ d k_{\varphi} } + 1\right) \sqrt{\mathrm{D}_{\mathrm{F}^4}\left(q_{\vlambda^*}, \ell\right)}
  \\
  &=
  (2 + \delta) \left(
  L^2 \, \left( 2 k_{\varphi} \sqrt{d} + 1 \right) 
  + 
  S^{2} \left( \sqrt{ d k_{\varphi} } + 1 \right) \right) \norm{\vlambda - \vlambda^*}_2^2
  \\
  &\qquad+
  (1 + 2 \delta^{-1}) \left( \sqrt{ d k_{\varphi} } + 1 \right) \sqrt{\mathrm{D}_{\mathrm{F}^4}\left(q_{\vlambda^*} \ell\right)}.
\end{align*}
\end{proofEnd}


\newpage
\begin{remark}[\textbf{Mean-Field Variational Family}]
  We prove an equivalent result for the mean-field variational family, \cref{thm:stl_upperbound_mf} in \cref{section:stl_meanfield}, which has an \(\mathcal{O}\left(\sqrt{d}\right)\) dimensional dependence.
\end{remark}
\vspace{0.5ex}
\begin{remark}[\textbf{Interpolation Condition}]
  \cref{thm:stl_upperbound} encompasses both settings where the variational family is well-specified and misspecified.
  That is, when the variational family is well specified, \textit{i.e.}, \( \mathrm{D}_{\mathrm{F}^4}\left(q_{\vlambda^*}, \pi\right) = 0 \), we obtain interpolation such that \(\beta_{\mathrm{STL}} = 0\).
\end{remark}
\vspace{0.5ex}
\begin{remark}[\textbf{Adaptivity of Bound}]
  When the variational family is well specified such that \( \mathrm{D}_{\mathrm{F}^4}\left(q_{\vlambda^*}, \pi\right) = 0 \), we can adaptively tighten the bound by setting \(\delta = 0\), where \(\alpha_{\mathrm{STL}}\) is reduced by a constant factor.
\end{remark}

\subsubsection{Lower Bounds}
We also obtain lower bounds on the expected-squared norm of the STL estimator to analyze its best-case behavior and the tightness of the bound.

\vspace{-1ex}
\paragraph{Necessary Conditions for Interpolation}
First, we obtain lower bounds that generally hold for all \(\vlambda \in \Lambda_L\) and any \(\pi\).
Our analysis relates the gradient variance with the Fisher-Hyv\"arinen divergence.
This can be related back to the KL divergence through an assumption on the posterior \(\pi\) known as the log-Sobolev inequality.
The general form of the log-Sobolev inequality was originally proposed by \citet{gross_logarithmic_1975} to study diffusion processes.
In this work, we use the form used by \citet{otto_generalization_2000}:
\begin{assumption}[\textbf{Log-Sobolev Inequality; LSI}]
\(\pi\) is said to satisfy the log-Sobolev inequality if, for any variational family \(\mathcal{Q}\) and all \(q_{\vlambda} \in \mathcal{Q}\), the following inequality holds:
{%
\setlength{\belowdisplayskip}{1.ex} \setlength{\belowdisplayshortskip}{1.ex}
\setlength{\abovedisplayskip}{0ex} \setlength{\abovedisplayshortskip}{0ex}
\[
  \DKL{q}{\pi}
  \leq
  \frac{C_{\mathrm{LSI}}}{2} \, \DHF{q}{\pi}.
\]
}%
\end{assumption}
\vspace{-1ex}
Strongly log-concave distributions are known to satisfy the LSI, where the strong log-concavity constant becomes the (inverse) LSI constant (see also~\citealp[Theorem 9.9]{villani_topics_2016}):

\vspace{1ex}
\begin{remark}[\citealp{bakry_diffusions_1985}]
    Let \(\pi\) be \(\mu\)-strongly log-concave.
    Then, LSI holds with \(C_{\mathrm{LSI}}^{-1} = \mu\).
\end{remark}

We now present a lower bound which holds for all \(\vlambda \in \Lambda_S\) and any log-differentiable \(\pi\):

\begin{theoremEnd}[category=stllowerbound]{theorem}\label{thm:stl_lowerbound}
  Assume \cref{assumption:variation_family}.
  The expected-squared norm of the STL estimator is lower bounded as
{%
\setlength{\belowdisplayskip}{1.ex} \setlength{\belowdisplayshortskip}{1.ex}
\setlength{\abovedisplayskip}{1.ex} \setlength{\abovedisplayshortskip}{1.ex}
  \[
    \mathbb{E} \norm{\rvvg_{\mathrm{STL}}\left(\vlambda\right)}_2^2 
    \geq
    \DHF{q_{\vlambda}}{\pi}
    \geq
    \frac{2}{C_{\mathrm{LSI}}} \DKL{q_{\vlambda}}{\pi},
  \]
  }
  for all \(\vlambda \in \Lambda_{S}\) and any \(0 < S < \infty\), where the last inequality holds if \(\pi\) is LSI.
\end{theoremEnd}
\vspace{-1ex}
\begin{proofEnd}\label{proof:stl_lowerbound}
\begin{align*}
  \mathbb{E} \norm{\rvvg_{\mathrm{STL}}\left(\vlambda\right)}_2^2
  &=
  \mathbb{E} \norm{  
  \nabla_{\vlambda} \log \ell \left(\mathcal{T}_{\vlambda}\left(\rvvu\right)\right) - \nabla_{\vlambda} \log q_{\vnu}\left(\mathcal{T}_{\vlambda}\left(\rvvu\right)\right)
  }_2^2
  \;\Bigg\lvert_{\vnu = \vlambda},
\shortintertext{by \cref{thm:jacobian_reparam_inner},}
  &=
  \mathbb{E} 
  J_{\mathcal{T}}\left(\rvvu\right)
  \norm{
  \nabla \log \ell \left(\mathcal{T}_{\vlambda}\left(\rvvu\right)\right) 
  - \nabla \log q_{\vnu}\left(\mathcal{T}_{\vlambda}\left(\rvvu\right)\right)
  }_2^2
  \;\Big\lvert_{\vnu = \vlambda}
  \\
  &=
  \mathbb{E} 
  J_{\mathcal{T}}\left(\rvvu\right)
  \norm{
  \nabla \log \ell \left(\mathcal{T}_{\vlambda}\left(\rvvu\right)\right) 
  - \nabla \log q_{\vlambda}\left(\mathcal{T}_{\vlambda}\left(\rvvu\right)\right)
  }_2^2,
\shortintertext{since \(J_{\mathcal{T}}\left(\rvvu\right) \geq 1\) for both the full-rank and mean-field parameterizations,}
  &\geq
  \mathbb{E} 
  \norm{
  \nabla \log \ell \left(\mathcal{T}_{\vlambda}\left(\rvvu\right)\right) 
  - \nabla \log q_{\vlambda}\left(\mathcal{T}_{\vlambda}\left(\rvvu\right)\right)
  }_2^2,
\shortintertext{after Change-of-Variable,}
  &=
  \mathbb{E}_{\rvvz \sim q_{\vlambda}} 
  \norm{
  \nabla \log \ell \left(\rvvz\right) 
  - \nabla \log q_{\vlambda}\left(\rvvz\right)
  }_2^2,
\shortintertext{and since \(\log \pi(\vz) = \log \ell\left(\vz\right) + \log Z\) for some constant \(Z > 0\),} 
  &=
  \mathbb{E}_{\rvvz \sim q_{\vlambda}} 
  \norm{
  \nabla \log \pi\left(\rvvz\right) 
  - \nabla \log q_{\vlambda}\left(\rvvz\right)
  }_2^2
  \\
  &=
  \DHF{q_{\vlambda}}{\ell}.
\shortintertext{Finally, when the log-Sobolev inequality applies,}
  &\geq
  \frac{2}{C_{\mathrm{LSI}}} \DKL{q_{\vlambda}}{\pi}.
\end{align*}
\end{proofEnd}


\begin{corollary}[\textbf{Necessary Conditions for Interpolation}]
For the STL estimator, the interpolation condition does not hold if
\begin{enumerate}[label=\textbf{(\roman*)}]
  \vspace{-1.5ex}
  \setlength\itemsep{0.ex}
    \item \(\DHF{q_{\vlambda^*_{\mathrm{F}}}}{\pi} > 0\), or,
    \item when \(\pi\) is LSI, \(\DKL{q_{\vlambda^*_{\mathrm{KL}}}}{\pi} > 0\),
  \vspace{-1.5ex}
\end{enumerate}
  \begin{center}
   {\begingroup
    \setlength\tabcolsep{10pt} 
  \begin{tabular}{ll}
    \text{where }
    &
    \(\vlambda_{\mathrm{F}}^* \in \argmin_{\vlambda \in \Lambda_S} \DHF{q_{\vlambda}}{\pi} \),\; 
    \text{and} \\
    & \(\vlambda_{\mathrm{KL}}^* \in \argmin_{\vlambda \in \Lambda_S} \DKL{q_{\vlambda}}{\pi} \),
  \end{tabular}
  \endgroup}
  \end{center}
  \vspace{-1.5ex}
  for any \(0 < S < \infty\).
\end{corollary}
\vspace{-1ex}

\vspace{-1.ex}
\paragraph{Tightness Analysis}
The bound in \cref{thm:stl_lowerbound} is unfortunately not tight regarding the constants.
It, however, holds for all \(\vlambda\) and \(\pi\).
Instead, we establish an alternative lower bound that holds for some \(\vlambda\) and \(\pi\) but is tight regarding the dependence on \(d\) and \(L\).
\vspace{1ex}

\begin{theoremEnd}[all end, category=gradvarlemmas]{lemma}\label{lemma:unorm_times_reparam}
  Let \(\mathcal{T}_{\vlambda}: \mathbb{R}^p \times \mathbb{R}^d \rightarrow \mathbb{R}^d\) be the location-scale reparameterization function (\cref{def:reparam}) and \(\rvvu \sim \varphi\) satisfy \cref{assumption:symmetric_standard}.
  Then,
  \[
  \mathbb{E} 
  \left(1 + \textstyle{\sum^{d}_{i=1} \rvu_i^2} \right)
  \left(
  \mathcal{T}_{\vlambda}\left(\rvvu\right) 
  +
  \vz
  \right)
  =
  \left(d+1\right)
  \left( \vm + \vz \right)
  \]
  for any \(\vz \in \mathbb{R}^d\).
\end{theoremEnd}
\vspace{-1ex}
\begin{proofEnd}
  \begin{align*}
  \mathbb{E} 
  \left(1 + \textstyle{\sum^{d}_{i=1} \rvu_i^2} \right)
  \left(
  \mathcal{T}_{\vlambda}\left(\rvvu\right) + \vz
  \right)
  &=
  \mathbb{E} 
  \left(1 + \norm{\rvvu}_2^2 \right)
  \left( \mC \rvvu + \vm + \vz \right)
  \\
  &=
  \mC \mathbb{E} \left(1 + \norm{\rvvu}_2^2 \right) \rvvu +  \mathbb{E}\left(1 + \norm{\rvvu}_2^2 \right) \left( \vm + \vz \right)
  \\
  &=
  \left(d+1\right) \left( \vm + \vz \right),
  \end{align*}
  where the last equality follows from \cref{thm:u_identities,assumption:symmetric_standard}.
\end{proofEnd}

\begin{theoremEnd}[all end, category=gradvarlemmas]{lemma}\label{thm:lowerbound_matrix_innerproduct_lemma}
  Let \(\mA = \mathrm{diag}\left(A_1, \ldots, A_d\right) \in \mathbb{R}^{d \times d}\) be some diagonal matrix, define
{%
\setlength{\belowdisplayskip}{1.ex} \setlength{\belowdisplayshortskip}{1.ex}
\setlength{\abovedisplayskip}{1.ex} \setlength{\abovedisplayshortskip}{1.ex}
  \[
  \mB = \begin{bmatrix}
    L^{-1/2} &   &        &   \\
             & L^{1/2} &        &   \\
             &   & \ddots &   \\
             &   &        & L^{1/2} \\
  \end{bmatrix},
  \qquad
  \mC = L^{-1/2} \, \boldupright{I},
  \]
  }%
  some \(\vu \in \mathbb{R}^d\), \(\vm \in \mathbb{R}^d\), and \(\vz \in \mathbb{R}^d \) such that \(m_1 = z_1\).
  For \(\vlambda = (\vm, \mC)\), the expression
{%
\setlength{\belowdisplayskip}{1.ex} \setlength{\belowdisplayshortskip}{1.ex}
\setlength{\abovedisplayskip}{1.ex} \setlength{\abovedisplayshortskip}{1.ex}
  \[
    \norm{\mB^{-1} \mC^{-1} \left( \mA \rvvu + \vm - \vz \right)}_2^2.
  \]
}%
  can be bounded for the following instances of \(\mA\):
  \begin{enumerate}[label=\textbf{(\roman*)}]
    \item If \(\mA = \mC\),  
    {%
    \setlength{\belowdisplayskip}{1.ex} \setlength{\belowdisplayshortskip}{1.ex}
    \setlength{\abovedisplayskip}{1.ex} \setlength{\abovedisplayshortskip}{1.ex}
      \begin{align*}
      &{\lVert
        \mB^{-1} \mC^{-1} \left(\mC \vu + \vm - \vz\right)
      \rVert}_2^2
      =
      \norm{\mC \vu + \vm - \vz}_2^2
      + {\left(L - L^{-1} \right)} \, u_1^2,
      \end{align*}
    }
    \item while if \(\mA = \boldupright{O}\),  \\
    {%
    \setlength{\belowdisplayskip}{1.ex} \setlength{\belowdisplayshortskip}{1.ex}
    \setlength{\abovedisplayskip}{-1.ex} \setlength{\abovedisplayshortskip}{-1.ex}
      \[
      {\lVert
        \mB^{-1} \mC^{-1} \left(\vm - \vz\right)
      \rVert}_2^2
      =
      \norm{\vm - \vz}_2^2.
      \]
    }
  \end{enumerate}
\end{theoremEnd}
\begin{proofEnd}
  First notice that 
  \begin{align*}
    \mB^{-1} \mC^{-1}
    =
    \begin{bmatrix}
      L &       &        &   \\
        & 1     &        &   \\
        &       & \ddots &   \\
        &       &        & 1 \\
    \end{bmatrix}.
  \end{align*}

  Denoting the 1st coordinate of \(\mA \vu + \vm \) as \({[\mA \vu + \vm]}_1 = A_1 \rvu_1 + m_1\), we have
  \begin{align}
    &\norm{
      \mB^{-1} \mC^{-1} \left(\mA \vu + \vm - \vz\right)
    }_2^2
    \\
    &\;=
    \norm{
    \begin{bmatrix}
      L &       &        &   \\
        & 1     &        &   \\
        &       & \ddots &   \\
        &       &        & 1 \\
    \end{bmatrix} 
    \left(\mA \vu + \vm - \vz\right)
    }_2^2
    \nonumber
    \\
    &\;=
    \norm{\mA \vu + \vm - \vz}_2^2
    + {\left(L^{2} - 1 \right)} {\left( {[\mA \vu + \vm]}_1 - z_1 \right)}^2
    \nonumber
    \\
    &\;=
    \norm{\mA \rvvu + \vm - \vz}_2^2
    + {\left(L^{2} - 1 \right)} {\left(A_1 u_1 + m_1 - z_1 \right)}^2,
    \nonumber
\shortintertext{and using the fact that \(m_1 = z_1\)}
    &\;=
    \norm{\mA \vu + \vm - \vz}_2^2
    + {\left(L^{2} - 1 \right)} \, A_1^2  \, u_1^2.
    \label{eq:unimprovability_key_lemma_eq1}
  \end{align}

  \paragraph{Proof of (i)}
  If \(\mA = \mC = L^{-1/2} \boldupright{I}\), \cref{eq:unimprovability_key_lemma_eq1} yields,
  \begin{align*}
    \norm{
      \mB^{-1} \mC^{-1} \left(\mA \vu + \vm - \vz\right)
    }_2^2
    &=
    \norm{\mC \vu + \vm - \vz}_2^2
    + {\left(L^{2} - 1 \right)} \, L^{-1} u_1^2
    \\
    &=
    \norm{\mC \vu + \vm - \vz}_2^2
    + {\left(L - L^{-1} \right)} \, u_1^2
  \end{align*}

  \paragraph{Proof of (ii)}
  If \(\mA = \boldupright{O}\), \cref{eq:unimprovability_key_lemma_eq1} yields,
  \begin{align*}
    \norm{
      \mB^{-1} \mC^{-1} \left(\mA \vu + \vm - \vz\right)
    }_2^2
    &=
    \norm{\vm - \vz}_2^2.
  \end{align*}

\end{proofEnd}

\begin{theoremEnd}[category=stllowerboundunimprovability]{theorem}\label{thm:stl_lowerbound_unimprovability}
 Assume \cref{assumption:variation_family}.
  There exists a strongly-convex, \(L\)-log-smooth posterior and some variational parameter \(\widetilde{\vlambda} \in \Lambda_{L}\) for all \(L \geq 1\) such that
{%
\setlength{\belowdisplayskip}{1.ex} \setlength{\belowdisplayshortskip}{1.ex}
\setlength{\abovedisplayskip}{1.ex} \setlength{\abovedisplayshortskip}{1.ex}
  {
  \begin{align*}
    \mathbb{E} {\lVert \rvvg_{\mathrm{STL}}\left(\widetilde{\vlambda}\right) \rVert}_2^2
    &\geq
    \left(
    L^2
    \left( d + k_{\varphi} \right) 
    -2
    \left(d + 1 \right) 
    \right)
    {\lVert \widetilde{\mC} \rVert}_{\mathrm{F}}^2
    \\
    &\qquad
    - 2
    \left(k_{\varphi} - 1\right) \norm{ \widetilde{\vm} - \bar{\vz} }_2^2,
  \end{align*}
  }%
  }%
  where \(\widetilde{\vlambda} = (\widetilde{\vm}, \widetilde{\mC})\) and \(\bar{\vz}\) is a stationary point of the said log posterior.
\end{theoremEnd}
\vspace{-1ex}
\begin{proofEnd}\label{proof:stl_lowerbound_unimprovability}
  The worst case is achieved by the following:
  \begin{enumerate}[label=\textbf{(\roman*)}]
  \item \textbf{\(\log \ell\) is ill-conditioned such that the smoothness constant is large.} 
    This results in the domain \(\Lambda_{L}\) to include ill-conditioned \(\mC\)s, which has the largest impact on the gradient variance. Furthermore, 
  \item \textbf{\(\pi\) and \(q_{\vlambda}\) need to have the least overlap in probability volume.} 
    This means the variance reduction effect will be minimal.
  \end{enumerate}
  For Gaussians, this is equivalent to minimizing 
  \(
    {\lVert \mS^{-1} \mSigma^{-1} \rVert}_{\mathrm{F}}^2
  \)
  while maximizing \({\lVert \mSigma^{-1} \rVert}_{\mathrm{F}}^2\) and \({\lVert \mS^{-1} \rVert}_{\mathrm{F}}^2\).

  We therefore choose
  \[
  \pi = \mathcal{N}\left(\bar{\vz}, \mSigma\right)
  \qquad
  q_{\vlambda} = \mathcal{N}\left(\widetilde{\vm}, \widetilde{\mS}\right),
  \]
  where 
  \[
  \mSigma = \begin{bmatrix}
    L^{-1}  &   &        &   \\
           & L &        &   \\
           &   & \ddots &   \\
           &   &        & L \\
  \end{bmatrix},
  \qquad
  \widetilde{\mS} = L^{-1} \boldupright{I},
  \;\quad\text{and}\quad
  \widetilde{\vm} = \begin{bmatrix}
    \bar{z}_1 \\ m_2 \\ \vdots \\ m_d
  \end{bmatrix},
  \]
  where \(\bar{z}_1\) is the 1st element of \(\bar{\vz}\) such that \(\widetilde{m}_1 = \bar{z}_1\).
  The choice of \(\widetilde{m}_1 = \bar{z}_1\) is purely for clarifying the derivation.
  Notice that \(\mSigma\) has \(d-1\) entries set as \(L\), only one entry set as \(L^{-1}\), and \(\widetilde{\mS} = \widetilde{\mC}\widetilde{\mC}\).
  Here, \(\pi\) is \(L^{-1}\)-strongly log-concave, \(L\)-log smooth, and \(\widetilde{\vlambda} = \left(\widetilde{\vm}, \widetilde{\mC}\right) \in \Lambda_{L}\).

  \paragraph{General Gaussian \(\pi\) Lower Bound}
  As usual, we start from the definition of the STL estimator as
  \begin{align*}
    \mathbb{E} \norm{\rvvg_{\mathrm{STL}}\left(\vlambda\right)}_2^2
    &=
    \mathbb{E} \norm{
    \nabla_{\vlambda} \log \ell \left(\mathcal{T}_{\vlambda}\left(\rvvu\right)\right) - \nabla_{\vlambda} \log q_{\vnu}\left(\mathcal{T}_{\vlambda}\left(\rvvu\right)\right)
  }_2^2
    \;\Big\lvert_{\vnu = \vlambda}
  \shortintertext{by \cref{thm:jacobian_reparam_inner},}
    &=
    \mathbb{E} 
    J_{\mathcal{T}}\left(\rvvu\right)
    \norm{
    \nabla \log \ell \left(\mathcal{T}_{\vlambda}\left(\rvvu\right)\right) 
    - \nabla \log q_{\vnu}\left(\mathcal{T}_{\vlambda}\left(\rvvu\right)\right)
    }_2^2
    \;\Big\lvert_{\vnu = \vlambda},
\shortintertext{since both \(\pi\) and \(q_{\vlambda}\) are Gaussians,}
    &=
    \mathbb{E} 
    \left(1 + \textstyle{\sum^{d}_{i=1} \rvu_i^2} \right)
    \norm{
    \nabla \log \ell \left(\mathcal{T}_{\vlambda}\left(\rvvu\right)\right) 
    - \nabla \log q_{\vlambda}\left(\mathcal{T}_{\vlambda}\left(\rvvu\right)\right)
    }_2^2
    \\
    &=
    \mathbb{E} 
    \left(1 + \textstyle{\sum^{d}_{i=1} \rvu_i^2} \right)
    \norm{
      \mSigma^{-1} \left(\mathcal{T}_{\vlambda}\left(\rvvu\right) - \bar{\vz}\right)
      -
      \mS^{-1} \left(\mathcal{T}_{\vlambda}\left(\rvvu\right) - \vm\right)
    }_2^2
    \\
    &=
    \mathbb{E} 
    \left(1 + \textstyle{\sum^{d}_{i=1} \rvu_i^2} \right)
    \norm{
      \mSigma^{-1} \left(\mathcal{T}_{\vlambda}\left(\rvvu\right) - \bar{\vz}\right)
      -
      \mS^{-1} \left(\mathcal{T}_{\vlambda}\left(\rvvu\right) - \bar{\vz}\right)
      +
      \mS^{-1} \left(\vm - \bar{\vz}\right)
    }_2^2
    \\
    &=
    \mathbb{E} 
    \left(1 + \textstyle{\sum^{d}_{i=1} \rvu_i^2} \right)
    \Big(
    {\lVert
      \mSigma^{-1} \left(\mathcal{T}_{\vlambda}\left(\rvvu\right) - \bar{\vz}\right)
    \rVert}_2^2
    +
    {\lVert
      \mS^{-1} \left(\mathcal{T}_{\vlambda}\left(\rvvu\right) - \bar{\vz}\right)
    \rVert}_2^2
    +
    {\lVert
      \mS^{-1} \left(\vm - \bar{\vz}\right)
    \rVert}_2^2
    \\
    &\quad\qquad\qquad\qquad\qquad
    -2
    \inner{
      \mSigma^{-1} \left(\mathcal{T}_{\vlambda}\left(\rvvu\right) - \bar{\vz}\right)
    }{
      \mS^{-1} \left(\mathcal{T}_{\vlambda}\left(\rvvu\right) - \bar{\vz}\right)
    }
    \\
    &\quad\qquad\qquad\qquad\qquad
    +2
    \inner{
      \mSigma^{-1} \left(\mathcal{T}_{\vlambda}\left(\rvvu\right) - \bar{\vz}\right)
    }{
      \mS^{-1} \left(\vm - \bar{\vz}\right)
    }
    \\
    &\quad\qquad\qquad\qquad\qquad
    -2
    \inner{
      \mS^{-1} \left(\mathcal{T}_{\vlambda}\left(\rvvu\right) - \bar{\vz}\right)
    }{
      \mS^{-1} \left(\vm - \bar{\vz}\right)
    }
    \Big),
\shortintertext{distributing the expectation and \(1 + \textstyle{\sum^{d}_{i=1} \rvu_i^2}\),}
    &=
    \mathbb{E} 
    \left(1 + \textstyle{\sum^{d}_{i=1} \rvu_i^2} \right)
    \Big(
    {\lVert
      \mSigma^{-1} \left(\mathcal{T}_{\vlambda}\left(\rvvu\right) - \bar{\vz}\right)
    \rVert}_2^2
    +
    {\lVert
      \mS^{-1} \left(\mathcal{T}_{\vlambda}\left(\rvvu\right) - \bar{\vz}\right)
    \rVert}_2^2
    \Big)
    \\
    &\quad+
    \mathbb{E} 
    \left(1 + \textstyle{\sum^{d}_{i=1} \rvu_i^2} \right)
    {\lVert
      \mS^{-1} \left(\vm - \bar{\vz}\right)
    \rVert}_2^2
    \\
    &\quad
    -2 \,
    \mathbb{E} 
    \left(1 + \textstyle{\sum^{d}_{i=1} \rvu_i^2} \right)
    \inner{
      \mSigma^{-1} \left(
      \mathcal{T}_{\vlambda}\left(\rvvu\right) - \bar{\vz}\right)
    }{
      \mS^{-1} \left(\mathcal{T}_{\vlambda}\left(\rvvu\right) - \bar{\vz}\right)
    }
    \\
    &\quad
    +2 \,
    \inner{
      \mSigma^{-1} 
      \mathbb{E} 
      \left(1 + \textstyle{\sum^{d}_{i=1} \rvu_i^2}  \right)
      \left(
      \mathcal{T}_{\vlambda}\left(\rvvu\right) - \bar{\vz}\right)
    }{
      \mS^{-1} \left(\vm - \bar{\vz}\right)
    }
    \\
    &\quad
    -2 \,
    \inner{
      \mS^{-1} 
      \mathbb{E} 
      \left(1 + \textstyle{\sum^{d}_{i=1} \rvu_i^2} \right)
      \left(
      \mathcal{T}_{\vlambda}\left(\rvvu\right) - \bar{\vz}\right)
    }{
      \mS^{-1} \left(\vm - \bar{\vz}\right)
    },
\shortintertext{applying \cref{lemma:unorm_times_reparam,thm:u_identities} to the second term and the last two inner product terms,}
    &=
    \mathbb{E} 
    \left(1 + \textstyle{\sum^{d}_{i=1} \rvu_i^2} \right)
    \Big(
    {\lVert
      \mSigma^{-1} \left(\mathcal{T}_{\vlambda}\left(\rvvu\right) - \bar{\vz}\right)
    \rVert}_2^2
    +
    {\lVert
      \mS^{-1} \left(\mathcal{T}_{\vlambda}\left(\rvvu\right) - \bar{\vz}\right)
    \rVert}_2^2
    \Big)
    \\
    &\quad+
    \left(d + 1\right)
    {\lVert
      \mS^{-1} \left(\vm - \bar{\vz}\right)
    \rVert}_2^2
    \\
    &\quad
    -2 \,
    \mathbb{E} 
    \left(1 + \textstyle{\sum^{d}_{i=1} \rvu_i^2} \right)
    \inner{
      \mSigma^{-1} \left(
      \mathcal{T}_{\vlambda}\left(\rvvu\right) - \bar{\vz}\right)
    }{
      \mS^{-1} \left(\mathcal{T}_{\vlambda}\left(\rvvu\right) - \bar{\vz}\right)
    }
    \\
    &\quad
    +2 \,
    \left(d + 1\right) 
    \inner{
      \mSigma^{-1} \left( \vm - \bar{\vz}\right)
    }{
      \mS^{-1} \left(\vm - \bar{\vz}\right)
    }
    \\
    &\quad
    -2 \,
    \left(d + 1\right) 
    \inner{
      \mS^{-1} \left(\vm - \bar{\vz}\right)
    }{
      \mS^{-1} \left(\vm - \bar{\vz}\right)
    }.
\shortintertext{The last two inner products can be denoted as norms such that}
    &\;=
    \mathbb{E} 
    \left(1 + \textstyle{\sum^{d}_{i=1} \rvu_i^2} \right)
    \Big(
    {\lVert
      \mSigma^{-1} \left(\mathcal{T}_{\vlambda}\left(\rvvu\right) - \bar{\vz}\right)
    \rVert}_2^2
    +
    {\lVert
      \mS^{-1} \left(\mathcal{T}_{\vlambda}\left(\rvvu\right) - \bar{\vz}\right)
    \rVert}_2^2
    \Big)
    \\
    &\quad
    +
    \left(d + 1\right)
    {\lVert
      \mS^{-1} \left(\vm - \bar{\vz}\right)
    \rVert}_2^2
    \\
    &\quad
    -2 \,
    \mathbb{E} 
    \left(1 + \textstyle{\sum^{d}_{i=1} \rvu_i^2} \right)
    {\lVert
      \mB^{-1} \mC^{-1} \left( \mathcal{T}_{\vlambda}\left(\rvvu\right) - \bar{\vz}\right)
    \rVert}_2^2
    \\
    &\quad
    +2 
    \left(d + 1\right)
    {\lVert
      \mB^{-1} \mC^{-1} \left(\vm - \bar{\vz}\right)
    \rVert}_2^2
    -2
    \left(d + 1\right)
    {\lVert
      \mS^{-1} \left(\vm - \bar{\vz}\right)
    \rVert}_2^2,
  \end{align*}
  where \(\mB\) is the matrix square root of \(\mSigma\) such that \(\mB^{-1} \mB^{-1} = \mSigma^{-1}\).
  The derivation so far applies to any Gaussian \(\pi, q_{\vlambda}\) and \(\vlambda \in \Lambda_{S}\) for any \(S > 0\).

  \paragraph{Worst-Case Lower Bound}
  Now, for our worst-case example, 
  \begin{align*}
    \mathbb{E} {\lVert \rvvg_{\mathrm{STL}}\left(\widetilde{\vlambda}\right) \rVert}_2^2
    &=
    \mathbb{E} 
    \left(1 + \textstyle{\sum^{d}_{i=1} \rvu_i^2} \right)
    \Big(
    {\lVert
      \mSigma^{-1} \left(\mathcal{T}_{\widetilde{\vlambda}}\left(\rvvu\right) - \bar{\vz}\right)
    \rVert}_2^2
    +
    {\lVert
      \widetilde{\mS}^{-1} \left(\mathcal{T}_{\widetilde{\vlambda}}\left(\rvvu\right) - \bar{\vz}\right)
    \rVert}_2^2
    \Big)
    \\
    &\quad
    +
    \left(d + 1\right)
    {\lVert
      \widetilde{\mS}^{-1} \left(\widetilde{\vm} - \bar{\vz}\right)
    \rVert}_2^2
    \\
    &\quad
    -2 \,
    \mathbb{E} 
    \left(1 + \textstyle{\sum^{d}_{i=1} \rvu_i^2} \right)
    {\lVert
      \mB^{-1} \widetilde{\mC}^{-1} \left( \mathcal{T}_{\widetilde{\vlambda}}\left(\rvvu\right) - \bar{\vz}\right)
    \rVert}_2^2
    \\
    &\quad
    +2 
    \left(d + 1\right)
    {\lVert
      \mB^{-1} \widetilde{\mC}^{-1} \left(\widetilde{\vm} - \bar{\vz}\right)
    \rVert}_2^2
    -2 
    \left(d + 1\right)
    {\lVert
      \widetilde{\mS}^{-1} \left(\widetilde{\vm} - \bar{\vz}\right)
    \rVert}_2^2,
\shortintertext{since \(\pi\) is \(\mu\)-strongly log-concave and \(\widetilde{\mS}^{-1} = L \boldupright{I}\),}
    &\geq
    \mathbb{E} 
    \left(1 + \textstyle{\sum^{d}_{i=1} \rvu_i^2} \right)
    \Big(
    L^{-2} \norm{\mathcal{T}_{\widetilde{\vlambda}}\left(\rvvu\right) - \bar{\vz}}_2^2
    +
    L^2 \norm{\mathcal{T}_{\widetilde{\vlambda}}\left(\rvvu\right) - \bar{\vz}}_2^2
    \Big)
    \\
    &\quad
    + \left(d + 1\right) L^2 \norm{\widetilde{\vm} - \bar{\vz}}_2^2
    \\
    &\quad
    -2 \,
    \mathbb{E} 
    \left(1 + \textstyle{\sum^{d}_{i=1} \rvu_i^2} \right)
    {\lVert
      \mB^{-1} \widetilde{\mC}^{-1} \left( \mathcal{T}_{\widetilde{\vlambda}}\left(\rvvu\right) - \bar{\vz}\right)
    \rVert}_2^2
    \\
    &\quad
    +2 \left(d + 1\right) \,
    {\lVert
      \mB^{-1} \widetilde{\mC}^{-1} \left(\widetilde{\vm} - \bar{\vz}\right)
    \rVert}_2^2
    -2 \left(d + 1\right) L^2 \norm{ \widetilde{\vm} - \bar{\vz} }_2^2,
\shortintertext{and grouping the terms,}
    &=
    \underbrace{
    \left( L^{-2} + L^2 \right) \,
    \mathbb{E} 
    \left(1 + \textstyle{\sum^{d}_{i=1} \rvu_i^2} \right)
    \norm{\mathcal{T}_{\widetilde{\vlambda}}\left(\rvvu\right) - \bar{\vz}}_2^2
    -
    \left(d + 1\right) L^2 \norm{\widetilde{\vm} - \bar{\vz}}_2^2
    }_{T_{\text{\ding{172}}}}
    \\
    &\qquad
    \underbrace{
    -2 \,
    \mathbb{E} 
    \left(1 + \textstyle{\sum^{d}_{i=1} \rvu_i^2} \right)
    {\lVert
      \mB^{-1} \widetilde{\mC}^{-1} \left( \mathcal{T}_{\widetilde{\vlambda}}\left(\rvvu\right) - \bar{\vz}\right)
    \rVert}_2^2
    }_{T_{\text{\ding{173}}}}
    \\
    &\qquad
    \underbrace{
    +2 \left(d + 1\right) \,
    {\lVert
      \mB^{-1} \widetilde{\mC}^{-1} \left(\widetilde{\vm} - \bar{\vz}\right)
    \rVert}_2^2.
    }_{T_{\text{\ding{174}}}}
  \end{align*}

\paragraph{Lower Bound on \(T_{\text{\ding{172}}}\)}
For \(T_{\text{\ding{172}}}\), we have
  \begin{align}
    T_{\text{\ding{172}}}
    &=
    \left(L^{-2} + L^2 \right)
    \mathbb{E} 
    \left(1 + \textstyle{\sum^{d}_{i=1} \rvu_i^2} \right)
    \norm{\mathcal{T}_{\widetilde{\vlambda}}\left(\rvvu\right) - \bar{\vz}}_2^2
    - 
    \left(d + 1\right) L^2 \norm{\widetilde{\vm} - \bar{\vz}}_2^2,
    \nonumber
\shortintertext{applying \cref{thm:normdist_1pnormu},}
    &=
    \left(L^{-2} + L^2 \right)
    \left(
      \left( d + 1 \right) \norm{\widetilde{\vm} - \bar{\vz}}_2^2
      +
      \left( d + k_{\varphi} \right) {\lVert \widetilde{\mC} \rVert}_{\mathrm{F}}^2
    \right)
    -
    \left(d + 1\right) L^2 \norm{\widetilde{\vm} - \bar{\vz}}_2^2,
    \nonumber
\shortintertext{and since \(L^{-2} > 0\) and is negligible for large \(L\)s,}
    &\geq
    L^2
    \left(
      \left( d + 1 \right) \norm{\widetilde{\vm} - \bar{\vz}}_2^2
      +
      \left( d + k_{\varphi} \right) {\lVert \widetilde{\mC} \rVert}_{\mathrm{F}}^2
    \right)
    -
    \left(d + 1\right) L^2 \norm{\widetilde{\vm} - \bar{\vz}}_2^2
    \nonumber
    \\
    &=
    L^2
    \left( d + k_{\varphi} \right) {\lVert \widetilde{\mC} \rVert}_{\mathrm{F}}^2.
    \label{eq:unimprovability_term1}
  \end{align}

\paragraph{Lower Bound on \(T_{\text{\ding{173}}}\)}
For \(T_{\text{\ding{173}}}\), we now use the covariance structures of our worst case through \cref{thm:lowerbound_matrix_innerproduct_lemma}.
That is,
  \begin{align*}
    T_{\text{\ding{173}}}
    &=
    -2 \,
    \mathbb{E} 
    \left(1 + \textstyle{\sum^{d}_{i=1} \rvu_i^2} \right)
    {\lVert
      \mB^{-1} \widetilde{\mC}^{-1} \left( \mathcal{T}_{\widetilde{\vlambda}}\left(\rvvu\right) - \bar{\vz}\right)
    \rVert}_2^2.
\shortintertext{Noting that \(\mathcal{T}_{\widetilde{\vlambda}}\left(\rvvu\right) = \widetilde{\mC} \rvvu + \widetilde{\vm}\) by definition, we can apply \cref{thm:lowerbound_matrix_innerproduct_lemma} Item (i) as}
    &=
    -2 \,
    \mathbb{E} 
    \left(1 + \textstyle{\sum^{d}_{i=1} \rvu_i^2} \right)
    \left(
      \norm{\mathcal{T}_{\widetilde{\vlambda}}\left(\rvvu\right) - \bar{\vz}}_2^2 + {\left(L - L^{-1} \right)} \rvu_1^2 
    \right),
\shortintertext{distributing the expectation and \(1 + \textstyle{\sum^{d}_{i=1} \rvu_i^2}\),}
    &=
    -2 \,
    \Big(
    \mathbb{E} 
    \left(1 + \textstyle{\sum^{d}_{i=1} \rvu_i^2} \right)
    \norm{\mathcal{T}_{\widetilde{\vlambda}}\left(\rvvu\right) - \bar{\vz}}_2^2
    +
    \underbrace{
    \mathbb{E} 
    \left(1 + \textstyle{\sum^{d}_{i=1} \rvu_i^2} \right)
    {\left(L - L^{-1} \right)} \rvu_1^2 
    }_{T_{\text{\ding{175}}}}
    \Big),
  \end{align*}

  \(T_{\text{\ding{175}}}\) follows as
  \begin{align}
    T_{\text{\ding{175}}}
    &=
    \mathbb{E} 
    \left(1 + \textstyle{\sum^{d}_{i=1} \rvu_i^2} \right)
    {\left(L - L^{-1} \right)} \rvu_1^2 
    \nonumber
    \\
    &=
    {\left(L^1 - L^{-1} \right)}
    \mathbb{E} 
    \left(1  + \textstyle{\sum^{d}_{i=1} \rvu_i^2} \right)  \rvu_1^2
    \nonumber
    \\
    &=
    {\left(L - L^{-1} \right)}
    \left(\mathbb{E} \rvu_1^2 + \mathbb{E} \rvu_1^4 + \textstyle{\sum^{d}_{i=2} \mathbb{E} \rvu_i^2 \mathbb{E} \rvu_1^2} \right)  ,
    \nonumber
\shortintertext{applying \cref{thm:u_identities},}
    &=
    {\left(L - L^{-1} \right)}
    \left(1 + k_{\varphi} + d - 1 \right)  
    \nonumber
    \\
    &=
    {\left(L - L^{-1} \right)} 
    \left(d + k_{\varphi}\right).
    \label{eq:stl_lowerbound_173}
  \end{align}

  Then,
  \begin{align}
    T_{\text{\ding{173}}}
    &=
    -2 \,
    \Big(
    \mathbb{E} 
    \left(1 + \textstyle{\sum^{d}_{i=1} \rvu_i^2} \right)
    \norm{\mathcal{T}_{\widetilde{\vlambda}}\left(\rvvu\right) - \bar{\vz}}_2^2
    +
    T_{\text{\ding{175}}}
    \Big),
    \nonumber
\shortintertext{bringing \cref{eq:stl_lowerbound_173} in,}
    &=
    -2 \,
    \Big(
    \mathbb{E} 
    \left(1 + \textstyle{\sum^{d}_{i=1} \rvu_i^2} \right)
    \norm{\mathcal{T}_{\widetilde{\vlambda}}\left(\rvvu\right) - \bar{\vz}}_2^2
    +
    {\left(L - L^{-1} \right)} 
    \left(d + k_{\varphi}\right)
    \Big),
    \nonumber
\shortintertext{applying \cref{thm:normdist_1pnormu},}
    &=
    -2 \,
    \Big(
    \left(d + k_{\varphi}\right) \norm{ \widetilde{\vm} - \bar{\vz} }_2^2
    +
    \left(d + 1 \right) {\lVert \widetilde{\mC} \rVert}_{\mathrm{F}}^2
    +
    \left(d + k_{\varphi}\right)
    {\left(L - L^{-1} \right)} 
    \Big)
    \label{eq:unimprovability_term2}
  \end{align}

\paragraph{Lower Bound on \(T_{\text{\ding{174}}}\)}
  Similarly for \(T_{\text{\ding{174}}}\), we can apply \cref{thm:lowerbound_matrix_innerproduct_lemma} Item (ii) as
  \begin{align}
    T_{\text{\ding{174}}}
    =
    2 \left(d+1\right)
    {\lVert
      \mB^{-1} \widetilde{\mC}^{-1} \left( \widetilde{\vm} - \bar{\vz}\right)
    \rVert}_2^2
    =
    2 \left(d+1\right) \norm{\widetilde{\vm} - \bar{\vz}}_2^2.
    \label{eq:unimprovability_term3}
  \end{align}

  Combining \cref{eq:unimprovability_term1,eq:unimprovability_term2,eq:unimprovability_term3},
  \begin{align*}
    \mathbb{E} {\lVert\rvvg_{\mathrm{STL}}\left(\widetilde{\vlambda}\right) \rVert}_2^2
    &\geq
    T_{\text{\ding{172}}} + T_{\text{\ding{173}}} + T_{\text{\ding{174}}}
    \\
    &\geq
    L^2
    \left( d + k_{\varphi} \right) {\lVert \widetilde{\mC} \rVert}_{\mathrm{F}}^2
    -2 \,
    \Big(
    \left(d + k_{\varphi}\right) \norm{ \widetilde{\vm} - \bar{\vz} }_2^2
    +
    \left(d + 1 \right) {\lVert \widetilde{\mC} \rVert}_{\mathrm{F}}^2
    +
    \left(d + k_{\varphi}\right)
    {\left(L - L^{-1} \right)} 
    \Big)
    + 2 \left(d + 1\right) \norm{\widetilde{\vm} - \bar{\vz}}_2^2
    \\
    &=
    \left(
    L^2
    \left( d + k_{\varphi} \right) 
    -2
    \left(d + 1 \right) 
    \right)
    {\lVert \widetilde{\mC} \rVert}_{\mathrm{F}}^2
    -
    2 \left(k_{\varphi} - 1\right) \norm{ \widetilde{\vm} - \bar{\vz} }_2^2
    +
    \left(d + k_{\varphi}\right)
    {\left(L - L^{-1} \right)} ,
\shortintertext{and when \(L \geq 1\), we have \(L - L^{-1} > 0\). Therefore, we can simply the bound as}
    &\geq
    \left(
    L^2
    \left( d + k_{\varphi} \right) 
    -2
    \left(d + 1 \right) 
    \right)
    {\lVert \widetilde{\mC} \rVert}_{\mathrm{F}}^2
    -
    2 \left(k_{\varphi} - 1\right) \norm{ \widetilde{\vm} - \bar{\vz} }_2^2.
  \end{align*}

\end{proofEnd}


%
\vspace{0.5ex}
\begin{remark}\label{remark:stl_tightness}
  \cref{thm:stl_lowerbound_unimprovability} implies that \cref{thm:stl_upperbound}  with \(S = L\) is tight with respect to the dimension dependence \(d\) and the log-smoothness \(L\) except for a factor of 4.
\end{remark}
\vspace{1ex}
\begin{remark}[\textbf{Room for Improvement}]
  Part of the factor of \(4\) looseness is due to the extreme worst case: when \(\nabla \log \pi\) and \(\nabla \log q_{\vlambda}\) are anti-correlated.
  This worst case is unlikely to appear in practice, thus making a tighter lower bound challenging to obtain.
  But at the same time, we were unsuccessful at seeking a general assumption that would rule out these worst cases in the upper bound.
  Specifically, we tried very hard to apply coercivity/gradient monotonicity of log-concave distributions, but to no avail, leaving this to future works.
\end{remark}

\subsection{Theoretical Analysis of the CFE Estimator}
We now present the analysis of the CFE estimator.
While the CFE estimator has been studied in-depth by \citet{domke_provable_2019,kim_practical_2023,domke_provable_2023}, we slightly improve the latest analysis of \citet[Theorem 3]{domke_provable_2023}.
Specifically, we improve the constants and obtain an adaptive bound.
This ensures that we have a fair comparison with the STL estimator.

\begin{theoremEnd}[category=cfeupperbound]{theorem}\label{thm:cfe_upperbound}
  Assume \cref{assumption:variation_family} and that \(\ell\) is \(L\)-log-smooth.
  For the full-rank parameterization, the expected-squared norm of the CFE estimator is bounded as
{\setlength{\belowdisplayskip}{1.5ex} \setlength{\belowdisplayshortskip}{1.5ex}
\setlength{\abovedisplayskip}{1.5ex} \setlength{\abovedisplayshortskip}{1.5ex}
  \begin{align*}
    \mathbb{E} \norm{ \rvvg_{\mathrm{CFE}}\left(\vlambda\right) }_2^2
    \leq \alpha_{\mathrm{CFE}} {\lVert \vlambda - \vlambda^* \rVert}_2^2 + \beta_{\mathrm{CFE}}
  \end{align*}
  }%
  where 
{\setlength{\belowdisplayskip}{1.5ex} \setlength{\belowdisplayshortskip}{1.5ex}
\setlength{\abovedisplayskip}{1.5ex} \setlength{\abovedisplayshortskip}{1.5ex}
  \begin{align*}
    \alpha_{\mathrm{CFE}} &= L^2 \left( d + k_{\varphi} \right) \left(1 + \delta\right) + {\left(L + S \right)}^{2} \\
    \beta_{\mathrm{CFE}} &=  L^2 \left( d + k_{\varphi} \right) \left(1 + \delta^{-1} \right) {\lVert \vlambda^* - \bar{\vlambda} \rVert}_2^2
  \end{align*}
  }%
  for any \(\vlambda, \vlambda^* \in \Lambda_S\) and \(\delta > 0\), where \(\bar{\vlambda} = \left(\bar{\vz}, \mathbf{0}\right)\) and \(\bar{\vz}\) is any stationary point of \(f\).
\end{theoremEnd}
\vspace{-1ex}
\begin{proofEnd}\label{proof:cfe_upperbound}
  Following the notation of \citet{domke_provable_2023}, we denote \( \log \ell = f \).
  Then, starting from the definition of the variance,
  \begin{align}
    \mathbb{E} \norm{ \rvvg_{\mathrm{CFE}}\left(\vlambda\right) }_2^2
    &=
    \mathrm{tr}\mathbb{V}\rvvg\left(\vlambda\right)
    +
    \norm{ \mathbb{E} \rvvg_{\mathrm{CFE}}\left(\vlambda\right) }_2^2,
    \nonumber
\shortintertext{and by the unbiasedness of \(\rvvg_{\mathrm{CFE}}\),}
    &=
    \mathrm{tr}\mathbb{V}\rvvg\left(\vlambda\right)
    +
    \norm{ \nabla F\left(\vlambda\right) }_2^2,
    \nonumber
\shortintertext{by the definition of \(\rvvg_{\mathrm{CFE}}\) (\cref{def:cfe}),}
    &=
    \mathrm{tr}\mathbb{V}_{\rvvz \sim q_{\vlambda}} \left( \nabla_{\vlambda} f\left( \rvvz \right) + \nabla \mathbb{H}\left(q_{\vlambda}\right) \right)
    +
    \norm{ \nabla F\left(\vlambda\right) }_2^2.
    \nonumber
\shortintertext{We now apply the property of the variance: the deterministic components are neglected as}
    &=
    \mathrm{tr}\mathbb{V}_{\rvvz \sim q_{\vlambda}} \nabla_{\vlambda} f\left( \rvvz \right) 
    +
    \norm{ \nabla F\left(\vlambda\right) }_2^2
    \nonumber
    \\
    &\leq
    \mathbb{E}_{\rvvz \sim q_{\vlambda}} \norm{ \nabla_{\vlambda} f\left( \rvvz \right) }_2^2
    +
    \norm{ \nabla F\left(\vlambda\right) }_2^2.
    \label{eq:thm_cfe}
  \end{align}

  For \(L\)-log-smooth posteriors (\(L\)-smooth \(f\)), \citet[Theorem 3]{domke_provable_2019} show that
  \begin{align*}
    \mathbb{E}_{\rvvz \sim q_{\vlambda}} \norm{ \nabla_{\vlambda} f\left( \rvvz \right) }_2^2
    &\leq
    L^2  \left( 
    \left( d + k_{\varphi} \right) \norm{ \vm - \bar{\vz} }_2^2
    +
    \left( d + 1 \right) \norm{ \mC }_{\mathrm{F}}^2
    \right),
\shortintertext{and since \(k_{\varphi} \geq 1\),}
    &\leq
    L^2  \left( 
    \left( d + k_{\varphi} \right) \norm{ \vm - \bar{\vz} }_2^2
    +
    \left( d + k_{\varphi} \right) \norm{ \mC }_{\mathrm{F}}^2
    \right)
    \\
    &=
    L^2 \left( d + k_{\varphi} \right) {\lVert \vlambda - \bar{\vlambda} \rVert}_2^2,
  \end{align*}
  which is tight.

  Applying \cref{eq:peterpaul}, we have
  \begin{align}
    \mathbb{E}_{\rvvz \sim q_{\vlambda}} {\lVert \nabla_{\vlambda} f\left( \rvvz \right) \rVert}_2^2
    &\leq
    L^2 \left( d + k_{\varphi} \right) {\lVert \vlambda - \bar{\vlambda} \rVert}_2^2
    \nonumber
    \\
    &=
    L^2 \left( d + k_{\varphi} \right) {\lVert \vlambda - \vlambda^* + \vlambda^* - \bar{\vlambda} \rVert}_2^2
    \nonumber
    \\
    &\leq
    L^2 \left( d + k_{\varphi} \right) \left( \left(1 + \delta \right) {\lVert \vlambda - \vlambda^* \rVert}_2^2 + \left(1 + \delta^{-1} \right) {\lVert \vlambda^* - \bar{\vlambda} \rVert}_2^2 \right).
    \label{eq:thm_cfe_energy}
  \end{align}
  
  Now, for \(\vlambda \in \Lambda_S\),~\citet[Theorem 1 \& Lemma 12]{domke_provable_2020} show that the negative ELBO \(F\) is (\(L + S\))-smooth as
  \begin{align}
    \norm{ \nabla F\left(\vlambda\right) }_2^2
    =
    \norm{ \nabla F\left(\vlambda\right) - \nabla F\left(\vlambda^*\right) }_2^2
    \leq 
    {\left(L + S \right)}^{2} \norm{ \vlambda - \vlambda^* }_2^2.
    \label{eq:thm_cfe_entropy}
  \end{align}

  Now back to \cref{eq:thm_cfe},
  \begin{align*}
    \mathbb{E} \norm{ \rvvg_{\mathrm{CFE}}\left(\vlambda\right) }_2^2
    &\leq
    \mathbb{E}_{\rvvz \sim q_{\vlambda}} \norm{ \nabla_{\vlambda} f\left( \rvvz \right) }_2^2
    +
    \norm{ \nabla F\left(\vlambda\right) }_2^2
\shortintertext{applying \cref{eq:thm_cfe_energy},}
    &\leq
    L^2 \left( d + k_{\varphi} \right) \left( \left(1 + \delta \right) {\lVert \vlambda - \vlambda^* \rVert}_2^2 + \left(1 + \delta^{-1} \right) {\lVert \vlambda^* - \bar{\vlambda} \rVert}_2^2 \right)
    +
    \norm{ \nabla F\left(\vlambda\right) }_2^2
\shortintertext{and \cref{eq:thm_cfe_entropy},}
    &\leq
    L^2 \left( d + k_{\varphi} \right) \left( \left(1 + \delta \right) {\lVert \vlambda - \vlambda^* \rVert}_2^2 + \left(1 + \delta^{-1} \right) {\lVert \vlambda^* - \bar{\vlambda} \rVert}_2^2 \right)
    +
    {\left(L + S \right)}^{2} \norm{ \vlambda - \vlambda^* }_2^2
    \\
    &=
    \left( L^2 \left( d + k_{\varphi} \right) \left(1 + \delta\right) + {\left(L + S \right)}^{2} \right) {\lVert \vlambda - \vlambda^* \rVert}_2^2 
    +
    L^2 \left( d + k_{\varphi} \right) \left(1 + \delta^{-1} \right) {\lVert \vlambda^* - \bar{\vlambda} \rVert}_2^2.
  \end{align*}
\end{proofEnd}

\begin{theoremEnd}[all end, category=cfeupperboundmf]{theorem}\label{thm:cfe_upperbound_mf}
  Assume \cref{assumption:variation_family} and that \(\ell\) is \(L\)-log-smooth.
  For the mean-field parameterization, the expected-squared norm of the CFE estimator is bounded as
  \begin{align*}
    \mathbb{E} \norm{ \rvvg_{\mathrm{CFE}}\left(\vlambda\right) }_2^2
    &\leq
    \big(
    \left(2 k_{\varphi} \sqrt{d} + 1 \right) 
    \left(1 + \delta \right)
    +
    {\left(L + S \right)}^{2}
    \big)
    {\lVert \vlambda - \vlambda^* \rVert}_2^2
    +
    \left(2 k_{\varphi} \sqrt{d} + 1 \right) \left(1 + \delta^{-1} \right) {\lVert \vlambda^* - \bar{\vlambda} \rVert}_2^2.
  \end{align*}
  for any \(\vlambda \in \Lambda_S\) and \(\delta \geq 0\), where \(\bar{\vlambda} = \left(\bar{\vz}, \mathbf{0}\right)\) and \(\bar{\vz}\) is any stationary point of \(f\).
\end{theoremEnd}
\begin{proofEnd}\label{proof:cfe_upperbound_mf}
  For the mean-field case, the only difference with \cref{thm:cfe_upperbound} is the upper bound on the energy term.
  The key step is the mean-field part of \cref{thm:normdist_1pnormu}, first proven by \citet{kim_practical_2023}.
  The remaining steps are similar to Theorem 1 of \citet{kim_practical_2023}.
  That is,
  \begin{align}
    \mathbb{E}_{\rvvz \sim q_{\vlambda}} \norm{ \nabla_{\vlambda} f\left( \rvvz \right) }_2^2
    &=
    \mathbb{E} \norm{ \nabla_{\vlambda} f\left( \mathcal{T}_{\vlambda}\left(\rvvu\right) \right) }_2^2,
    \nonumber
\shortintertext{applying \cref{thm:jacobian_reparam_inner},}
    &=
    \mathbb{E} J_{\mathcal{T}}\left(\rvvu\right) \norm{ \nabla f\left( \mathcal{T}_{\vlambda}\left(\rvvu\right) \right) }_2^2
    \nonumber
    \\
    &=
    \mathbb{E} J_{\mathcal{T}}\left(\rvvu\right) \norm{ \nabla f\left( \mathcal{T}_{\vlambda}\left(\rvvu\right) \right) - \nabla f\left(\bar{\vz}\right) }_2^2,
    \nonumber
\shortintertext{from \(L\)-smoothness of \(f = \log \ell\),}
    &\leq
    L^2 \, J_{\mathcal{T}}\left(\rvvu\right) \mathbb{E} \norm{ \mathcal{T}_{\vlambda}\left(\rvvu\right) - \bar{\vz} }_2^2,
    \nonumber
\shortintertext{applying \cref{thm:normdist_1pnormu},}
    &\leq
    L^2 \left(\sqrt{dk_{\varphi}} + k_{\varphi} \sqrt{d} + 1 \right) \norm{ \vm - \bar{\vz} }_2^2
    + L^2 \left(2 k_{\varphi} \sqrt{d} + 1\right) \norm{\mC}_{\mathrm{F}}^2.
    \nonumber
\shortintertext{and since \(k_{\varphi} \geq 1\), we have \(k_{\varphi} > \sqrt{k_{\varphi}}\), and thus}
    &\leq
    L^2 \left(2 k_{\varphi} \sqrt{d} + 1 \right) \left( \norm{ \vm - \bar{\vz} }_2^2 +  \norm{\mC}_{\mathrm{F}}^2 \right)
    \nonumber
    \\
    &=
    L^2 \left(2 k_{\varphi} \sqrt{d} + 1 \right) {\lVert \vlambda - \bar{\vlambda} \rVert}_2^2.
    \nonumber
\shortintertext{We finally apply \cref{eq:peterpaul} as}
    &\leq
    L^2 \left(2 k_{\varphi} \sqrt{d} + 1 \right)
    \left(
    \left(1 + \delta \right) {\lVert \vlambda - \vlambda^* \rVert}_2^2
      + \left(1 + \delta^{-1} \right) {\lVert \vlambda^* - \bar{\vlambda} \rVert}_2^2
    \right).
    \label{eq:cfe_meanfield_energy}
  \end{align}

  Combining this with \cref{eq:thm_cfe,eq:thm_cfe_entropy}, we have
  \begin{align*}
    \mathbb{E} \norm{ \rvvg_{\mathrm{CFE}}\left(\vlambda\right) }_2^2
    &=
    \mathbb{E}_{\rvvz \sim q_{\vlambda}} \norm{ \nabla_{\vlambda} f\left( \rvvz \right) }_2^2
    +
    \norm{ \nabla F\left(\vlambda\right) }_2^2
\shortintertext{and applying \cref{eq:cfe_meanfield_energy},}
    &\leq
    \mathbb{E}_{\rvvz \sim q_{\vlambda}} \norm{ \nabla_{\vlambda} f\left( \rvvz \right) }_2^2
    +
    {\left(L + S \right)}^{2} \norm{ \vlambda - \vlambda^* }_2^2
    \\
    &\leq
    \left(2 k_{\varphi} \sqrt{d} + 1 \right)
    \left(
      L^2 \left(1 + \delta \right) {\lVert \vlambda - \vlambda^* \rVert}_2^2
      + L^2 \left(1 + \delta^{-1} \right) {\lVert \vlambda^* - \bar{\vlambda} \rVert}_2^2
    \right)
    +
    {\left(L + S \right)}^{2} \norm{ \vlambda - \vlambda^* }_2^2
    \\
    &=
    \left(
    \left(2 k_{\varphi} \sqrt{d} + 1 \right) 
    L^2 \left(1 + \delta \right)
    +
    {\left(L + S \right)}^{2}
    \right)
    {\lVert \vlambda - \vlambda^* \rVert}_2^2
    +
    L^2 \left(2 k_{\varphi} \sqrt{d} + 1 \right) \left(1 + \delta^{-1} \right) {\lVert \vlambda^* - \bar{\vlambda} \rVert}_2^2.
  \end{align*}

\end{proofEnd}


\vspace{1ex}
\begin{remark}[\textbf{Comparison with STL}]\label{remark:variance_comparison}
    Compared to the STL estimator, the constant \(\alpha\) of the CFE estimator is tighter by a factor of \(4\).
    Considering \cref{thm:stl_lowerbound_unimprovability}, the constant factor difference should be marginal in practice.
\end{remark}

\vspace{1ex}
\begin{remark}[\textbf{Intuitions on \({\lVert \bar{\vlambda} - \vlambda^* \rVert}_2^2\)}]
  The quantity \({\lVert \bar{\vlambda} - \vlambda^* \rVert}_2^2\) can be expressed in the Wasserstein-2 distance as
{%
\setlength{\abovedisplayskip}{.5ex} \setlength{\abovedisplayshortskip}{.5ex}
\setlength{\belowdisplayskip}{1.ex} \setlength{\belowdisplayshortskip}{1.ex}
  \[
    \mathrm{d}_{\mathcal{W}_2}\left(q_{\vlambda^*}, \delta_{\bar{\vz}}\right) = \sqrt{ {\lVert \vm^* - \bar{\vz}\rVert}_2^2 + \norm{\mC^*}_{\mathrm{F}}^2} =  {\lVert \bar{\vlambda} - \vlambda^* \rVert}_2,
  \]
}
  where \(\delta_{\bar{\vz}}\) is a delta measure centered on the posterior mode \(\bar{\vz}\).
   Also, when the variational posterior mean \(\vm^*\) is close to \(\bar{\vz}\) such that  \({\lVert \vm^* - \bar{\vz}\rVert}_2^2 \approx 0\), \({\lVert \bar{\vlambda} - \vlambda^* \rVert}_2^2\) corresponds to the variational posterior variance as
{%
\setlength{\abovedisplayskip}{1ex} \setlength{\abovedisplayshortskip}{1ex}
\setlength{\belowdisplayskip}{1.ex} \setlength{\belowdisplayshortskip}{1.ex}
  \[
     {\lVert \bar{\vlambda} - \vlambda^* \rVert}_2^2 \approx \norm{\mC^*}_{\mathrm{F}}^2 = \mathrm{tr}\, \Vsub{\rvvz \sim q_{\vlambda^*}}{\rvvz}.
  \]
}
\end{remark}


\begin{theoremEnd}[all end, category=complexityprojsgdqvcfixed]{theorem}[\textbf{Strongly convex \(F\) with a fixed stepsize}]\label{thm:projsgd_stronglyconvex_fixedstepsize}
  For a \(\mu\)-strongly convex \(F : \Lambda \to \mathbb{R}\) on a convex set \(\Lambda\) with a unique global minimizer \(\vlambda^* \in \Lambda\), the last iterate \(\vlambda_{T}\) of projected SGD with a fixed stepsize satisfies \( \mathbb{E}{\lVert \vlambda_T - \vlambda^* \rVert}_2^2 \leq \epsilon \) if
{%
\setlength{\abovedisplayskip}{.5ex} \setlength{\abovedisplayshortskip}{.5ex}
\setlength{\belowdisplayskip}{1.ex} \setlength{\belowdisplayshortskip}{1.ex}
  \begin{align*}
    \gamma = \min\left( \frac{\epsilon \mu}{4 \beta}, \frac{\mu}{2 \alpha}, \frac{2}{\mu} \right)  \quad\text{and}\quad
    T \geq \max\left( \frac{ 4 \beta }{\mu^2 } \frac{1}{\epsilon}, \frac{2 \alpha}{\mu^2}, \frac{1}{2} \right) \log \left( 2 \norm{\vlambda_0 - \vlambda^*}_2^2 \, \frac{1}{\epsilon} \right).
  \end{align*}
}
\end{theoremEnd}
\begin{proofEnd}\label{proof:projsgd_stronglyconvex_fixedstepsize}
  Theorem 6 of \citet{domke_provable_2023} utilizes the two-stage stepsize of \citep{gower_sgd_2019}.
  The anytime convergence of the first stage, 
  \[
     \norm{ \vlambda_{T} - \vlambda^* }_2^2
     \leq
     {(1 - \gamma \mu)}^{T} \norm{ \vlambda_0 - \vlambda^* }_2^2
     +
     \frac{2 \gamma \beta}{\mu}
  \]
  corresponds to the SGD with only a fixed stepsize \(\gamma < \frac{\mu}{2 \alpha}\).

  Here, the result follows from Lemma A.2 of \citet{garrigos_handbook_2023} by plugging the constants
  \[
    \alpha_0 = \norm{ \vlambda_0 - \vlambda^* }_2^2, \quad
    A = \frac{2 \beta}{\mu}, 
    \;\text{and}\quad
    C = \frac{2 \alpha}{\mu}, \frac{\mu}{2}.
  \]
\end{proofEnd}

\begin{theoremEnd}[all end, category=complexityprojsgdqvcdec]{theorem}[\textbf{Strongly convex \(F\) with a decreasing stepsize schedule}]\label{thm:projsgd_stronglyconvex_decstepsize}
  For a \(\mu\)-strongly convex \(F : \Lambda \to \mathbb{R}\) on a convex set \(\Lambda\) with a unique global minimizer \(\vlambda^* \in \Lambda\), the last iterate \(\vlambda_{T}\) of projected SGD with a descreasing stepsize satisfies \( \mathbb{E}{\lVert \vlambda_T - \vlambda^* \rVert}_2^2 \leq \epsilon \) if
{%
\setlength{\abovedisplayskip}{.5ex} \setlength{\abovedisplayshortskip}{.5ex}
\setlength{\belowdisplayskip}{1.ex} \setlength{\belowdisplayshortskip}{1.ex}
  \begin{align*}
    \gamma_t = \min\left( \frac{\mu}{2 \alpha}, \frac{4 t + 2 }{ \mu \, {\left( t + 1\right)}^2 } \right)
    \quad\text{and}\quad
    T \geq \frac{16 \beta}{ \mu^2 } \frac{1}{\epsilon} + \frac{8 \alpha \, \norm{\vlambda_0 - \vlambda^*}_2 }{ \mu^2} \frac{1}{\sqrt{\epsilon}}.
  \end{align*}
}
\end{theoremEnd}
\begin{proofEnd}\label{proof:projsgd_stronglyconvex_decstepsize}
  Theorem 6 of \citet{domke_provable_2023} utilizes the two-stage stepsize of \citet{gower_sgd_2019}.
  After \(T\) steps, with a carefully tuned stepsize of 
  \[
    \gamma_t = \min\left( \frac{\mu}{2 \alpha}, \frac{4 t + 2 }{ \mu \, {\left( t + 1\right)}^2 } \right)
  \]
  projected SGD achieves
  \[
    \norm{\vlambda_{T} - \vlambda^*}_2^2
    \leq
    \frac{64 \alpha}{\mu^2} \frac{ \norm{\vlambda_0 - \vlambda^*}_2^2 }{T^2} 
    +
    \frac{32 \beta}{\mu^2} \frac{1}{T}.
  \]
  Following a similar strategy to \citet{kim_convergence_2023}, we can obtain a computational complexity by solving for the smallest \(T\) that achieves
  \[
    \frac{64 \alpha}{\mu^2} \frac{ \norm{\vlambda_0 - \vlambda^*}_2^2 }{T^2} 
    +
    \frac{16 \beta}{\mu^2} \frac{1}{T}
    \leq
    \epsilon.
  \]
  After re-organizing, we solve for
  \[
     A \, T^2  + B \, T + C = 0,
  \]
  where 
  \[
    A = \epsilon, \quad
    B = -\frac{16 \beta}{\mu^2},\; \text{and} \quad
    C = - \frac{64 \alpha^2}{\mu^4} \norm{\vlambda_0 - \vlambda^*}_2^2.
  \]
  Since \(T > 0\), the equation has a unique root
  \begin{align*}
    T 
    &= \frac{ - B + \sqrt{ B^2 - 4 A C  } }{ 2 A },
\shortintertext{applying the inequality \(\sqrt{a + b} \leq \sqrt{a} + \sqrt{b}\) for \(a, b \geq 0\),}
    &\leq \frac{ - B + \sqrt{ B^2 } +  \sqrt{  4 A \left( -C \right)  } }{ 2 A }
    = \frac{2 \left(-B\right) }{2 A} + \frac{ \sqrt{ 4 A  \left( -C \right)  } }{ 2 A }
    = \frac{\left(-B\right)}{A} + \frac{ \sqrt{ \left( -C \right) }  } { \sqrt{A} }
    \\
    &= \frac{ 16 \beta }{ \mu^2 \epsilon} + \frac{ \sqrt{  \frac{64 \alpha^2}{\mu^4} \norm{\vlambda_0 - \vlambda^*}_2^2  }  } { \sqrt{\epsilon} }
    \\
    &= \frac{ 16 \beta }{ \mu^2 \epsilon} + \frac{ 8 \alpha \norm{\vlambda_0 - \vlambda^*}_2 } { \mu^2 \sqrt{\epsilon} }.
  \end{align*}
\end{proofEnd}





\subsection{Non-Asymptotic Complexity of Black-Box Variational Inference}\label{section:bbvicomplexity}
\paragraph{Strongly Log-Concave Posteriors}
First, let us define the following:
\begin{definition}
    \(\pi\) is said to be \(\mu\)-strongly log-concave if its negative log-density \(V = -\log \pi\) satisfies the inequality
{%
\setlength{\abovedisplayskip}{.5ex} \setlength{\abovedisplayshortskip}{.5ex}
\setlength{\belowdisplayskip}{1.ex} \setlength{\belowdisplayshortskip}{1.ex}
    \[
      \inner{ \nabla V (\vz) }{ \vz - \vz' }
      \geq 
      V \left(\vz\right)
      -
      V \left(\vz'\right) 
      +
      \frac{\mu}{2} \norm{ \vz - \vz' }^2_2
    \]
}%
    for all \(\vz, \vz' \in \mathbb{R}^d\) and some \(\mu > 0\).
\end{definition}
Essentially, this assumes that the negative log-density of \(\pi\), $V$, is \(\mu\)-strongly convex.
It also implies that the density of \(\pi\) is lower bounded by some Gaussian.
A consequence of \(\mu\)-strong log-concavity is that, combined with the constraints on the variational parameterization in \cref{section:scale_parameterization}, the ELBO is also \(\mu\)-strongly convex \citep{domke_provable_2023,kim_convergence_2023,challis_gaussian_2013}.
\(\mu\)-strongly log-concave posteriors can easily be constructed by combining a log-concave likelihood with a Gaussian prior, and are popularly used to analyze BBVI and sampling algorithms.

\vspace{-1.ex}
\paragraph{Theoretical Setup}
We now apply the general complexity results for projected SGD established in \cref{section:projsgdcomplexity} to BBVI.
\begin{enumerate*}[label=\textbf{(\roman*)}]
    \item strongly log-concave posteriors,
    \item SGD run with fixed stepsizes, and 
    \item the full-rank variational family.
\end{enumerate*}
This is because the convergence analyses for \textbf{(ii)} \(\cap\) \textbf{(iii)} are the tightest.
Although the bounds for the mean-field parameterization have better dependences on \(d\), so far, it is unknown whether they are tight~\citep{kim_practical_2023}. (See also \citealp[Conjecture 1]{kim_convergence_2023}.)

\begin{theoremEnd}[all end, category=complexityprojsgdadaptiveqvcfixed]{lemma}[\textbf{Strongly convex \(F\) with adaptive QV and Fixed Stepsize}]\label{thm:projsgd_stronglyconvex_adaptive_complexity}
  For a \(\mu\)-strongly convex \(F : \Lambda \to \mathbb{R}\) on a convex set \(\Lambda\) the last iterate \(\vlambda_T\) of projected SGD with a gradient estimator satisfying an adaptive QV bound (\cref{assumption:adaptiveqvc}) is \(\epsilon\)-close to \(\vlambda^* = \argmin_{\vlambda \in \Lambda} F\left(\vlambda\right)\) such that \(\mathbb{E}\norm{\vlambda_T - \vlambda^{*}}_2^2 < \epsilon\) if
{%
\setlength{\abovedisplayskip}{.5ex} \setlength{\abovedisplayshortskip}{.5ex}
\setlength{\belowdisplayskip}{1.ex} \setlength{\belowdisplayshortskip}{1.ex}
  \begin{align*}
    \gamma &=
    \min\left(
      \frac{1}{2}
      \frac{
        \mu
      }{
        \widetilde{\alpha} + 2 \widetilde{\beta} \epsilon^{-1} 
      }\, ,\,
      \frac{2}{\mu}
    \right)  \quad\text{and}
    \\
    T &\geq
    \frac{2}{\mu^2} \max\left(\widetilde{\alpha} + 2 \widetilde{\beta} \frac{1}{\epsilon}, \; \frac{\mu^2}{4}\right) \log \left( 2 \norm{\vlambda_0 - \vlambda^*}_2^2 \, \frac{1}{\epsilon} \right).
  \end{align*}
}
\end{theoremEnd}
\vspace{-1ex}
\begin{proofEnd}\label{proof:projsgd_stronglyconvex_adaptive_complexity}
  Recall that, for a stepsize \(\gamma\) and a number of steps \(T\) satisfying 
  \begin{align*}
    \gamma \leq \min\left( \frac{\epsilon \mu}{4 \beta}, \frac{\mu}{2 \alpha}, \frac{2}{\mu} \right)
    \quad\text{and}\quad
    T \geq \max\left( \frac{ 4 \beta }{\mu^2 } \frac{1}{\epsilon}, \frac{2 \alpha}{\mu^2}, \frac{1}{2} \right) \log \left( 2 \norm{\vlambda_0 - \vlambda^*}_2^2 \, \frac{1}{\epsilon} \right),
  \end{align*}
  we can guarantee that the iterate \(\vlambda_t\) can guarantee \(\mathbb{E} \norm{\vlambda^* - \vlambda_T}_2^2 \leq \epsilon\).

  We optimize the parameter \(\delta\) to minimize the number of steps.
  That is,
  \begin{align*}
    \max\left( \frac{ 4 \beta }{\mu^2 } \frac{1}{\epsilon}, \frac{2 \alpha}{\mu^2}, \frac{1}{2} \right) \log \left( 2 \norm{\vlambda_0 - \vlambda^*}_2^2 \, \frac{1}{\epsilon} \right)
    =
    \frac{2}{\mu^2} \max\left( 2 (1 + C^{-1} \delta^{-1}) \,\widetilde{\beta} \frac{1}{\epsilon}, (1 + C \delta) \widetilde{\alpha}, \frac{\mu^2}{4}\right)
    \log \left( 2 \norm{\vlambda_0 - \vlambda^*}_2^2 \, \frac{1}{\epsilon} \right).
  \end{align*}
  Since the first and second arguments of the max function are monotonic with respect to \(\delta\), the optimum is unique, and achieved when the two terms are equal.
  That is,
  \begin{alignat*}{2}
    & &
    2 (1 + C^{-1} \delta^{-1}) \,\widetilde{\beta} \frac{1}{\epsilon}
    &=
    (1 + C \delta) \widetilde{\alpha}
    \\
    &\Leftrightarrow&\qquad
    \frac{2 \widetilde{\beta}}{\epsilon} + \frac{2 \widetilde{\beta} C^{-1}}{\epsilon} \delta^{-1} \,
    &=
    \widetilde{\alpha} + \widetilde{\alpha} C \delta 
    \\
    &\Leftrightarrow&\qquad
    \frac{2 \widetilde{\beta}}{\epsilon} \delta +  \frac{2 \widetilde{\beta} C^{-1}}{\epsilon}  \,
    &=
    \widetilde{\alpha} \delta + \widetilde{\alpha} C \delta^2
    \\
    &\Leftrightarrow&\qquad
    \widetilde{\alpha} C \delta^2 + \left( \widetilde{\alpha} - \frac{2 \widetilde{\beta}}{\epsilon} \right) \delta - \frac{2 \widetilde{\beta} C^{-1}}{\epsilon}
    &=
    0
    \\
    &\Leftrightarrow&\qquad
    \left(
    \widetilde{\alpha} \delta
    - 
    \frac{2 \widetilde{\beta} C^{-1}}{\epsilon}
    \right)
    \left(
      C \delta + 1
    \right)
    &=
    0.
  \end{alignat*}
  Conveniently, we have a unique feasible solution
  \begin{alignat*}{2}
    \delta
    =
    2 
    \frac{
      \widetilde{\beta}
    }{
      \widetilde{\alpha} 
    }
    C^{-1}
    \epsilon^{-1}.
  \end{alignat*}

  Thus, the optimal bound is obtained by setting
  \(
    \delta
    = 2 
    \frac{
      \widetilde{\beta}
    }{
      \widetilde{\alpha} 
    }
    C^{-1}
    \epsilon^{-1},
  \)
  such that
  \begin{align*}
    T
    &\geq
    \frac{2}{\mu^2} \max\left(  2 \beta \frac{1}{\epsilon}, \alpha, \frac{\mu^2}{4} \right) \log \left( 2 \norm{\vlambda_0 - \vlambda^*}_2^2 \, \frac{1}{\epsilon} \right)
    \\
    &=
    \frac{2}{\mu^2} \max\left( 2 \left( 1 + C^{-1} \delta^{-1} \right) \widetilde{\beta}  \frac{1}{\epsilon},  \left( 1 + C \delta \right) \widetilde{\alpha}, \frac{\mu^2}{4} \right)
    \log \left( 2 \norm{\vlambda_0 - \vlambda^*}_2^2 \, \frac{1}{\epsilon} \right)
    \\
    &=
    \frac{2}{\mu^2} \max\left( \widetilde{\alpha} + 2 \widetilde{\beta} \frac{1}{\epsilon}, \frac{\mu^2}{4} \right) \log \left( 2 \norm{\vlambda_0 - \vlambda^*}_2^2 \, \frac{1}{\epsilon} \right).
  \end{align*}
  The stepsize with the optimal \(\delta\) is consequently
  \begin{align*}
    \gamma
    &\leq
    \min\left( \frac{\epsilon \mu}{4 \beta}, \frac{\mu}{2 \alpha}, \frac{2}{\mu} \right)
    =
    \min\left( \frac{\epsilon \mu}{4 (1 + C^{-1} \delta^{-1}) \widetilde{\beta}} \,, \; \frac{\mu}{2 (1 + C \delta) \widetilde{\alpha}}, \frac{2}{\mu} \right)
    =
    \min\left(
      \frac{1}{2}
      \frac{
        \mu
      }{
        \widetilde{\alpha} + 2 \widetilde{\beta} \epsilon^{-1} 
      }\, ,\;
      \frac{2}{\mu}
    \right).
  \end{align*}
\end{proofEnd}

\begin{theoremEnd}[all end, category=complexityprojsgdadaptiveqvcdec]{lemma}[\textbf{Strongly convex \(F\) with adaptive QV and Decreasing Stepsize}]\label{thm:projsgd_stronglyconvex_decstepsize_adaptive_complexity}
  For a \(\mu\)-strongly convex \(F : \Lambda \to \mathbb{R}\) on a convex set \(\Lambda\) with a unique global minimizer \(\vlambda^* \in \Lambda\), the last iterate \(\vlambda_T\) of projected SGD with a gradient estimator satisfying an adaptive QVC bound (\cref{assumption:adaptiveqvc}) and a decreasing stepsize satisfies a suboptimality of \(\mathbb{E}\norm{\vlambda_T - \vlambda_{*}}_2^2 < \epsilon\) if
  {
  \begin{align*}
    \gamma_t
    &=
    \min\Bigg(
    \frac{
      \mu
    }{
      2 \widetilde{\alpha}
      +
      \sqrt{2 \norm{\vlambda_0 - \vlambda^*}_2 }
      \,
      \epsilon^{1/4}
      \,
      \widetilde{\alpha}^{3/2}
      \widetilde{\beta}^{-1/2}
    },
    \frac{4 t + 2 }{ \mu \, {\left( t + 1\right)}^2 }
    \Bigg)
    \\
    T
    &\geq
    \frac{16 \widetilde{\beta} }{ \mu^2 } \frac{1}{\epsilon} 
    +
    \frac{16 \sqrt{2}}{ \mu^2 }
    \sqrt{
      \norm{\vlambda_0 - \vlambda^*}_2
    }
    \,
    \sqrt{
      \widetilde{\alpha} \widetilde{\beta}
    }
    \,
    \frac{1}{\epsilon^{3/4}}
    +
    \frac{8 \widetilde{\alpha} \, \norm{\vlambda_0 - \vlambda^*}_2 }{ \mu^2}
    \frac{1}{\sqrt{\epsilon}}.
  \end{align*}
  }%
\end{theoremEnd}
\begin{proofEnd}\label{proof:projsgd_stronglyconvex_decstepsize_adaptive_complexity}
  Recall that, for a stepsize \(\gamma\) and a number of steps \(T\) such that
  \begin{align*}
    \gamma_t = \min\left( \frac{\mu}{2 \alpha}, \frac{4 t + 2 }{ \mu \, {\left( t + 1\right)}^2 } \right)
    \quad\text{and}\quad
    T \geq \frac{16 \beta}{ \mu^2 } \frac{1}{\epsilon} + \frac{8  \alpha \, \norm{\vlambda_0 - \vlambda^*}_2 }{ \mu^2} \frac{1}{\sqrt{\epsilon}},
  \end{align*}
  we can guarantee that the iterate \(\vlambda_t\) can guarantee \(\mathbb{E} \norm{\vlambda^* - \vlambda_T}_2^2 \leq \epsilon\).

  We optimize the parameter \(\delta\) to minimize the required number of steps \(T\).
  That is, we maximize
  \begin{align*}
    \frac{16 \beta }{ \mu^2 } \frac{1}{\epsilon} + \frac{8 \sqrt{2} \, \alpha \, \norm{\vlambda_0 - \vlambda^*}_2 }{ \mu^2} \frac{1}{\sqrt{\epsilon}}
    = \frac{16 \left(1 + C \delta\right) \widetilde{\beta} }{ \mu^2 } \frac{1}{\epsilon} + \frac{8 \left(1 + C^{-1} \delta^{-1}\right) \widetilde{\alpha} \, \norm{\vlambda_0 - \vlambda^*}_2 }{ \mu^2} \frac{1}{\sqrt{\epsilon}}.
  \end{align*}
  This is clearly a convex function with respect to \(\delta\).
  Thus, we only need to find a first-order stationary point 
  {\small%
  \begin{align*}
    \frac{\mathrm{d}}{\mathrm{d} \delta}
    \left(
      \frac{16 \left(1 + C \delta\right) \widetilde{\beta} }{ \mu^2 } \frac{1}{\epsilon} + \frac{8 \left(1 + C^{-1} \delta^{-1}\right) \widetilde{\alpha} \, \norm{\vlambda_0 - \vlambda^*}_2 }{ \mu^2} \frac{1}{\sqrt{\epsilon}}
    \right) &= 0.
  \end{align*}
  }%
  Differentiating, we have
  \begin{alignat*}{3}
    &&
    \frac{16 C \widetilde{\beta} }{ \mu^2 } \frac{1}{\epsilon} - \frac{8  \, C^{-1} \delta^{-2} \widetilde{\alpha} \, \norm{\vlambda_0 - \vlambda^*}_2 }{ \mu^2} \frac{1}{\sqrt{\epsilon}}
    &= 0,
\shortintertext{multiplying \(\delta^2\) to both sides,}
    &\Leftrightarrow&\qquad
    \delta^2 \frac{16 C \widetilde{\beta} }{ \mu^2 } \frac{1}{\epsilon} - \frac{8 \sqrt{2} \, C^{-1} \widetilde{\alpha} \, \norm{\vlambda_0 - \vlambda^*}_2 }{ \mu^2} \frac{1}{\sqrt{\epsilon}}
    &= 0.
  \end{alignat*}
  Reorganizing,
  \begin{alignat*}{3}
    &\Leftrightarrow&\qquad
    \delta^2  \frac{16 C \widetilde{\beta} }{ \mu^2 } \frac{1}{\epsilon} 
    &=
    \frac{8 \sqrt{2} \, C^{-1} \widetilde{\alpha} \, \norm{\vlambda_0 - \vlambda^*}_2 }{ \mu^2} \frac{1}{\sqrt{\epsilon}}
    \\
    &\Leftrightarrow&\qquad
    \delta^2  
    &=
    \left( \frac{\mu^2 \epsilon}{ 16 C \widetilde{\beta} } \right)
    \left(
    \frac{8  C^{-1} \widetilde{\alpha} \, \norm{\vlambda_0 - \vlambda^*}_2 }{ \mu^2} \frac{1}{\sqrt{\epsilon}}
    \right)
    \\
    &\Leftrightarrow&\qquad
    \delta^2  
    &=
    \frac{C^{-2} \widetilde{\alpha} \, \norm{\vlambda_0 - \vlambda^*}_2}{ 2 \widetilde{\beta}} \sqrt{\epsilon},
\shortintertext{and taking the square-root of both sides,}
    &\Leftrightarrow&\qquad
    \delta  
    &=
    \frac{
      \sqrt{ \norm{\vlambda_0 - \vlambda^*}_2 }
      \,
      \epsilon^{1/4}
      \,
      \sqrt{\widetilde{\alpha}}
    }{
      \sqrt{2} \,
      C 
      \sqrt{\widetilde{\beta}}
    }.
  \end{alignat*}
  Recall that the required number of iterations is
  \begin{align*}
    T
    &\geq
    \frac{16 \left(1 + C \delta\right) \widetilde{\beta} }{ \mu^2 } \frac{1}{\epsilon} 
    +
      \frac{8 \left(1 + C^{-1} \delta^{-1}\right) \widetilde{\alpha} \, \norm{\vlambda_0 - \vlambda^*}_2 }{ \mu^2} \frac{1}{\sqrt{\epsilon}}
    \\
    &=
    \underbrace{
      \frac{16 \widetilde{\beta} }{ \mu^2 } \frac{1}{\epsilon} 
      +
      \frac{16 \widetilde{\beta} }{ \mu^2 } \frac{1}{\epsilon} 
      C \delta
    }_{T_{\text{\ding{172}}}}
    +
    \underbrace{
      \frac{8 \widetilde{\alpha} \, \norm{\vlambda_0 - \vlambda^*}_2 }{ \mu^2} \frac{1}{\sqrt{\epsilon}}
      +
      \frac{8 \widetilde{\alpha} \, \norm{\vlambda_0 - \vlambda^*}_2 }{ \mu^2} \frac{1}{\sqrt{\epsilon}}
      C^{-1} \delta^{-1}
    }_{T_{\text{\ding{173}}}}.
  \end{align*}
  Plugging \(\delta\) in, we have
  \begin{align*}
    T_{\text{\ding{172}}}
    &=
    \frac{16 \widetilde{\beta} }{ \mu^2 } \frac{1}{\epsilon} 
    +
    \frac{16 \widetilde{\beta} }{ \mu^2 } \frac{1}{\epsilon}
    C 
    \left(
    \frac{
      \sqrt{ \norm{\vlambda_0 - \vlambda^*}_2 }
      \,
      \epsilon^{1/4}
      \,
      \sqrt{\widetilde{\alpha}}
    }{
      \sqrt{2} \, C
      \sqrt{\widetilde{\beta}}
    }
    \right)
    \\
    &=
    \frac{16 \widetilde{\beta} }{ \mu^2 } \frac{1}{\epsilon} 
    +
    \frac{8 \sqrt{2} }{ \mu^2 }
    \sqrt{
      \widetilde{\alpha} \widetilde{\beta}
    } \,
    \sqrt{ \norm{\vlambda_0 - \vlambda^*}_2 }
    \,
    \epsilon^{-3/4}
    \\
    \\
    T_{\text{\ding{173}}}
    &=
    \frac{8 \, \widetilde{\alpha} \, \norm{\vlambda_0 - \vlambda^*}_2 }{ \mu^2} \frac{1}{\sqrt{\epsilon}}
    +
    \frac{8 \, \widetilde{\alpha} \, \norm{\vlambda_0 - \vlambda^*}_2 }{ \mu^2} \frac{1}{\sqrt{\epsilon}}
    C^{-1} 
    \left(
    \frac{
      \sqrt{2} \, C
      \sqrt{\widetilde{\beta}}
    }{
      \sqrt{ \norm{\vlambda_0 - \vlambda^*}_2 }
      \,
      \epsilon^{1/4}
      \,
      \sqrt{\widetilde{\alpha}}
    }
    \right)
    \\
    &=
    \frac{8 \widetilde{\alpha} \, \norm{\vlambda_0 - \vlambda^*}_2 }{ \mu^2} \frac{1}{\sqrt{\epsilon}}
    +
    \frac{8 \sqrt{2}}{ \mu^2}
    \sqrt{ \norm{\vlambda_0 - \vlambda^*}_2 }
    \,
    \sqrt{
      \widetilde{\alpha}
      \widetilde{\beta}
    }
    \,
    \epsilon^{-3/4}.
  \end{align*}
  Combining the results, 
  \begin{align*}
    T
    \geq
    T_{\text{\ding{172}}}
    +
    T_{\text{\ding{173}}}
    &=
    \frac{16 \widetilde{\beta} }{ \mu^2 } \frac{1}{\epsilon} 
    +
    \frac{8 \sqrt{2} }{ \mu^2 }
    \sqrt{
      \widetilde{\alpha} \widetilde{\beta}
    } \,
    \sqrt{ \norm{\vlambda_0 - \vlambda^*}_2 }
    \,
    \epsilon^{-3/4}
    \\
    &\qquad+
    \frac{8 \widetilde{\alpha} \, \norm{\vlambda_0 - \vlambda^*}_2 }{ \mu^2} \frac{1}{\sqrt{\epsilon}}
    +
    \frac{8 \sqrt{2}}{ \mu^2}
    \sqrt{ \norm{\vlambda_0 - \vlambda^*}_2 }
    \,
    \sqrt{
      \widetilde{\alpha}
      \widetilde{\beta}
    }
    \,
    \epsilon^{-3/4}
    \\
    &=
    \frac{16 \widetilde{\beta} }{ \mu^2 } \frac{1}{\epsilon} 
    +
    \frac{16 \sqrt{2}}{ \mu^2 }
    \sqrt{
      \norm{\vlambda_0 - \vlambda^*}_2
    }
    \,
    \sqrt{
      \widetilde{\alpha} \widetilde{\beta}
    }
    \,
    \epsilon^{-3/4}
    +
    \frac{8 \widetilde{\alpha} \, \norm{\vlambda_0 - \vlambda^*}_2 }{ \mu^2}
    \frac{1}{\sqrt{\epsilon}}.
  \end{align*}

  For the stepsize
  \begin{align*}
    \gamma
    =
    \min\left( \frac{\mu}{2 \alpha}, \frac{4 t + 2 }{ \mu \, {\left( t + 1\right)}^2 } \right)
    =
    \min\left(
      \frac{\mu}{2 \left(1 + C \delta \right) \widetilde{\alpha}},
      \frac{4 t + 2 }{ \mu \, {\left( t + 1\right)}^2 }
    \right),
  \end{align*}
  we have 
  \begin{align*}
    2 \left(1 + C \delta \right) \widetilde{\alpha}
    &=
    2 \widetilde{\alpha} + 2 \widetilde{\alpha} C \delta 
    \\
    &=
    2 \widetilde{\alpha}
    +
    2 \widetilde{\alpha}
    C
    \left(
    \frac{
      \sqrt{ \norm{\vlambda_0 - \vlambda^*}_2 }
      \,
      \epsilon^{1/4}
      \,
      \sqrt{\widetilde{\alpha}}
    }{
      \sqrt{2} \, C
      \sqrt{\widetilde{\beta}}
    }
    \right)
    \\
    &=
    2 \widetilde{\alpha}
    +
    \sqrt{2}
    \sqrt{ \norm{\vlambda_0 - \vlambda^*}_2 }
    \,
    \epsilon^{1/4}
    \,
    \widetilde{\alpha}^{3/2}
    \widetilde{\beta}^{-1/2}.
  \end{align*}
  Therefore, 
  \begin{align*}
    \gamma
    =
    \min\left(
      \frac{\mu}{2 \left(1 + C \delta \right) \widetilde{\alpha}},
      \frac{4 t + 2 }{ \mu \, {\left( t + 1\right)}^2 }
    \right)
    =
    \min\left(
    \frac{
      \mu
    }{
      2 \widetilde{\alpha}
      +
      \sqrt{2 \norm{\vlambda_0 - \vlambda^*}_2 }
      \,
      \epsilon^{1/4}
      \,
      \widetilde{\alpha}^{3/2}
      \widetilde{\beta}^{-1/2}
    },
    \frac{4 t + 2 }{ \mu \, {\left( t + 1\right)}^2 }
    \right).
  \end{align*}
\end{proofEnd}

\paragraph{Complexity of BBVI on Strongly-Log-Concave \(\pi\)}
We can now plug in the constants obtained in \cref{section:gradient_variance}.
This immediately establishes the iteration complexity resulting from the use of different gradient estimators.
\vspace{0ex}

\begin{theoremEnd}[category=complexitybbvicfefixed]{theorem}[\textbf{Complexity of Fixed Stepsize BBVI with CFE}]\label{thm:projsgd_bbvicfe_complexity}
  The last iterate \(\vlambda_T \in \Lambda_L\) of BBVI with the CFE estimator and projected SGD with a fixed stepsize applied to a \(\mu\)-strongly log-concave and \(L\)-log-smooth posterior is \(\epsilon\)-close to \(\vlambda^* = \argmin_{\vlambda \in \Lambda_L} F\left(\vlambda\right)\) such that \(\mathbb{E}\norm{ \vlambda_T - \vlambda^* }_2^2 \leq \epsilon\) if
{
\setlength{\abovedisplayskip}{.5ex} \setlength{\abovedisplayshortskip}{.5ex}
\setlength{\belowdisplayskip}{1.ex} \setlength{\belowdisplayshortskip}{1.ex}
  \begin{align*}
    T
    &\geq 
    2 \kappa^2 \left(d + k_{\varphi} + 4\right) \left(1  + 2 {\lVert \bar{\vlambda} - \vlambda^* \rVert}_2^2 \frac{1}{\epsilon}\right) 
    \log \left( 2 \Delta^2 \, \frac{1}{\epsilon} \right)
  \end{align*}
}%
  for some fixed stepsize \(\gamma\), where \(\Delta = {\lVert \vlambda_0 - \vlambda^*\rVert}_2\), and \(\kappa = L/\mu\) is the condition number.
\end{theoremEnd}
\vspace{-1ex}
\begin{proofEnd}\label{proof:projsgd_bbvicfe_complexity}
  From \cref{thm:cfe_upperbound} with \(S = L\), the CFE estimator satisfies adaptive QV with the constants
  \begin{alignat*}{2}
    \alpha_{\mathrm{CFE}} 
    = L^2 \left( d + k_{\varphi} + 4 \right) \left(1 + \delta\right)
    \qquad\text{and}\qquad
    \beta_{\mathrm{CFE}}  
    = L^2 \left( d + k_{\varphi} \right) \left( 1 + \delta^{-1}  \right) {\lVert \bar{\vlambda} - \vlambda^* \rVert}_2^2.
  \end{alignat*}
  Furthermore, for a \(\mu\)-strongly log-concave posterior and our variational parameterization,~\citet[Theorem 9]{domke_provable_2020} show that the ELBO is \(\mu\)-strongly convex.

  We can thus invoke \cref{thm:projsgd_stronglyconvex_adaptive_complexity} with 
  \begin{align*}
    \widetilde{\alpha} = L^2 \left(d + k_{\varphi} + 4\right),\qquad
    \widetilde{\beta}  = L^2 \left(d + k_{\varphi}\right) {\lVert \bar{\vlambda} - \vlambda^* \rVert}_2^2,\quad\text{and}\quad
    C = 1.
  \end{align*}
  This yields a lower bound on the number of iteration 
  \begin{align*}
    &\frac{2}{\mu^2} \max\left(\widetilde{\alpha} + 2 \widetilde{\beta} \frac{1}{\epsilon}, \; \frac{\mu^2}{4} \right) 
    \log \left( 2 \norm{\vlambda_0 - \vlambda^*}_2^2 \, \frac{1}{\epsilon} \right)
    \\
    &\;=
    \frac{2}{\mu^2} \max\left(L^2 \left(d + k_{\varphi} + 4\right) + 2 L^2 \left(d + k_{\varphi}\right) {\lVert \bar{\vlambda} - \vlambda^* \rVert}_2^2 \frac{1}{\epsilon},\; \frac{\mu^2}{4}\right) 
    \log \left( 2 {\lVert \vlambda_0 - \vlambda^*\rVert}_2^2 \, \frac{1}{\epsilon} \right),
\shortintertext{pulling out \(L\),}
    &\;=
    \frac{2 L^2}{\mu^2} \max\left( \left(d + k_{\varphi} + 4\right) + 2 \left(d + k_{\varphi}\right) {\lVert \bar{\vlambda} - \vlambda^* \rVert}_2^2 \frac{1}{\epsilon},\; \frac{\mu^2}{4 L^2}\right) 
    \log \left( 2 {\lVert \vlambda_0 - \vlambda^*\rVert}_2^2 \, \frac{1}{\epsilon} \right),
\shortintertext{and since \(\frac{\mu^2}{4 L^2} < \frac{1}{4}\) and the first argument is larger than 1, the max operation is redundant that}
    &\;=
    \frac{2 L^2}{\mu^2} \left( \left(d + k_{\varphi} + 4\right) + 2 \left(d + k_{\varphi}\right) {\lVert \bar{\vlambda} - \vlambda^* \rVert}_2^2 \frac{1}{\epsilon}\right) 
    \log \left( 2 {\lVert \vlambda_0 - \vlambda^*\rVert}_2^2 \, \frac{1}{\epsilon} \right).
\shortintertext{Now, using the trivial fact \(d + k_{\varphi} < d + k_{\varphi} + 4\) simplifies the bound as,}
    &\;<
    \frac{2 L^2}{\mu^2} \left(d + k_{\varphi} + 4\right) \left(1  + 2 {\lVert \bar{\vlambda} - \vlambda^* \rVert}_2^2 \frac{1}{\epsilon} \right) 
    \log \left( 2 {\lVert \vlambda_0 - \vlambda^*\rVert}_2^2 \, \frac{1}{\epsilon} \right)
    \\
    &\;=
    2 \kappa^2 \left(d + k_{\varphi} + 4\right) \left(1  + 2 {\lVert \bar{\vlambda} - \vlambda^* \rVert}_2^2 \frac{1}{\epsilon}\right) 
    \log \left( 2 {\lVert \vlambda_0 - \vlambda^*\rVert}_2^2 \, \frac{1}{\epsilon} \right).
  \end{align*}
  The optimal \(\delta\) is given as
  \begin{align*}
    \delta 
    = \frac{2}{\epsilon} \frac{\widetilde{\beta}}{\widetilde{\alpha}} C^{-1} 
    = \frac{2}{\epsilon} 
    \frac{
      L^2 \left(d + k_{\varphi}\right) {\lVert \bar{\vlambda} - \vlambda^* \rVert}_2^2
    }{
      L^2 \left(d + k_{\varphi} + 4\right)
    } 
    C^{-1} 
    =
    \frac{2}{\epsilon} \, \frac{d + k_{\varphi}}{d + k_{\varphi} + 4} \, {\lVert \bar{\vlambda} - \vlambda^* \rVert}_2^2 .
  \end{align*}
\end{proofEnd}

\begin{theoremEnd}[all end, category=complexitybbvicfedec]{theorem}[\textbf{Complexity of Decreasing Stepsize BBVI with CFE}]\label{thm:projsgd_bbvicfe_decstepsize_complexity}
  The last iterate \(\vlambda_T \in \Lambda_L\) of BBVI with the CFE estimator and projected SGD with a decreasing stepsize schedule applied to a \(\mu\)-strongly log-concave and \(L\)-log-smooth posterior is \(\epsilon\)-close to \(\vlambda^* = \argmin_{\vlambda \in \Lambda_L} F\left(\vlambda\right) \) such that \(\mathbb{E}\norm{ \vlambda_T - \vlambda^* }_2^2 \leq \epsilon\) if
  \begin{align*}
    T
    &\geq 
    16 \kappa^2 \left(d + k_{\varphi} + 4\right) 
    \bigg(
    {\lVert \bar{\vlambda} - \vlambda^* \rVert}_2^2 \frac{1}{\epsilon} 
    +
    2
    \sqrt{
      \norm{\vlambda_0 - \vlambda^*}_2
    }
    \,
    {\lVert \bar{\vlambda} - \vlambda^* \rVert}_2
    \,
    \frac{1}{\epsilon^{3/4}}
    +
    \norm{\vlambda_0 - \vlambda^*}_2 
    \frac{1}{\sqrt{\epsilon}}
    \bigg).
  \end{align*}
  for some decreasing stepsize schedule \(\gamma_1, \ldots, \gamma_T\), where \(\kappa = L/\mu\) is the condition number and \(\vlambda^* \in \Lambda\) is the optimal variational parameter.
\end{theoremEnd}
\begin{proofEnd}\label{proof:projsgd_bbvicfe_decstepsize_complexity}
  From \cref{thm:cfe_upperbound}, the CFE estimator with \(S = L\) satisfies adaptive QV with the constants
  \begin{align*}
    \alpha_{\mathrm{CFE}} 
    = L^2 \left( d + k_{\varphi} + 4 \right) \left(1 + \delta\right)\qquad\text{and}\qquad
    \beta_{\mathrm{CFE}}  
    = L^2 \left( d + k_{\varphi} \right) \left( 1 + \delta^{-1}  \right) {\lVert \bar{\vlambda} - \vlambda^* \rVert}_2^2.
  \end{align*}
  Furthermore, for a \(\mu\)-strongly log-concave posterior and our variational parameterization,~\citet[Theorem 9]{domke_provable_2020} show that the ELBO is \(\mu\)-strongly convex.

  We thus invoke \cref{thm:projsgd_stronglyconvex_decstepsize_adaptive_complexity} with 
  \begin{alignat*}{3}
    \widetilde{\alpha} = L^2 \left(d + k_{\varphi} + 4\right),
    \qquad\quad
    \widetilde{\beta}  = L^2 \left(d + k_{\varphi}\right) {\lVert \bar{\vlambda} - \vlambda^* \rVert}_2^2, \qquad\text{and}
    \qquad
    C = 1.
  \end{alignat*}
  This yields a lower bound on the number of iterations:
  \begin{align*}
    &\frac{16 \widetilde{\beta} }{ \mu^2 } \frac{1}{\epsilon} 
    +
    \frac{16 \sqrt{2}}{ \mu^2 }
    \sqrt{
      \norm{\vlambda_0 - \vlambda^*}_2
    }
    \,
    \sqrt{
      \widetilde{\alpha} \widetilde{\beta}
    }
    \,
    \frac{1}{\epsilon^{3/4}}
    +
    \frac{8 \widetilde{\alpha} \, \norm{\vlambda_0 - \vlambda^*}_2 }{ \mu^2}
    \frac{1}{\sqrt{\epsilon}}
    \\
    &=
    \frac{16 L^2 \left(d + k_{\varphi}\right) {\lVert \bar{\vlambda} - \vlambda^* \rVert}_2^2 }{ \mu^2 } \frac{1}{\epsilon} 
    +
    \frac{16 \sqrt{2}}{ \mu^2 }
    \sqrt{
      \norm{\vlambda_0 - \vlambda^*}_2
    }
    \,
    \sqrt{
       \left( L^2 \left(d + k_{\varphi} + 4\right) \right)
       \left( L^2 \left(d + k_{\varphi}\right) {\lVert \bar{\vlambda} - \vlambda^* \rVert}_2^2 \right)
    }
    \,
    \frac{1}{\epsilon^{3/4}}
    \\
    &\qquad+
    \frac{8 L^2 \left(d + k_{\varphi} + 4\right) \, \norm{\vlambda_0 - \vlambda^*}_2 }{ \mu^2}
    \frac{1}{\sqrt{\epsilon}},
\shortintertext{using the trivial bound \(d + k_{\varphi} < d + k_{\varphi} + 4\),}
    &<
    \frac{16 L^2 \left(d + k_{\varphi} + 4\right) {\lVert \bar{\vlambda} - \vlambda^* \rVert}_2^2 }{ \mu^2 } \frac{1}{\epsilon} 
    +
    \frac{16 \sqrt{2}}{ \mu^2 }
    \sqrt{
      \norm{\vlambda_0 - \vlambda^*}_2
    }
    \,
    \sqrt{
       L^4 \left(d + k_{\varphi} + 4\right)
       \left(d + k_{\varphi} + 4\right) {\lVert \bar{\vlambda} - \vlambda^* \rVert}_2^2 
    }
    \,
    \frac{1}{\epsilon^{3/4}}
    \\
    &\qquad+
    \frac{8 L^2 \left(d + k_{\varphi} + 4\right) \, \norm{\vlambda_0 - \vlambda^*}_2 }{ \mu^2}
    \frac{1}{\sqrt{\epsilon}},
\shortintertext{pulling out the \(16 \left( d + k_{\varphi} + 4 \right) L^2 / \mu^2 \) factors,}
    &=
    16 \left(d + k_{\varphi} + 4\right) 
    \frac{L^2}{ \mu^2 } 
    \bigg(
    {\lVert \bar{\vlambda} - \vlambda^* \rVert}_2^2 \frac{1}{\epsilon} 
    +
    \sqrt{2}
    \sqrt{
      \norm{\vlambda_0 - \vlambda^*}_2
    }
    \,
    {\lVert \bar{\vlambda} - \vlambda^* \rVert}_2
    \,
    \frac{1}{\epsilon^{3/4}}
    +
    \frac{1}{2}
    \norm{\vlambda_0 - \vlambda^*}_2 
    \frac{1}{\sqrt{\epsilon}}
    \bigg)
    \\
    &=
    16 \kappa^2 \left(d + k_{\varphi} + 4\right) 
    \bigg(
    {\lVert \bar{\vlambda} - \vlambda^* \rVert}_2^2 \frac{1}{\epsilon} 
    +
    \sqrt{2}
    \sqrt{
      \norm{\vlambda_0 - \vlambda^*}_2
    }
    \,
    {\lVert \bar{\vlambda} - \vlambda^* \rVert}_2
    \,
    \frac{1}{\epsilon^{3/4}}
    +
    \frac{1}{2}
    \norm{\vlambda_0 - \vlambda^*}_2 
    \frac{1}{\sqrt{\epsilon}}
    \bigg).
  \end{align*}
  The optimal \(\delta\) is given as
  \begin{align*}
    \delta
    &=
    \frac{
      \sqrt{ \norm{\vlambda_0 - \vlambda^*}_2 }
      \,
      \epsilon^{1/4}
      \,
      \sqrt{\widetilde{\alpha}}
    }{
      \sqrt{2} \, C
      \sqrt{\widetilde{\beta}}
    }
    =
    \frac{
      \sqrt{ \norm{\vlambda_0 - \vlambda^*}_2 }
      \,
      \epsilon^{1/4}
      \,
      \sqrt{ L^2 \left(d + k_{\varphi} + 4\right) }
    }{
      \sqrt{2} 
      \sqrt{ L^2 \left(d + k_{\varphi}\right) {\lVert \bar{\vlambda} - \vlambda^* \rVert}_2^2 }
    }
    =
    \frac{1}{\sqrt{2} } \,
    \frac{\sqrt{ \norm{\vlambda_0 - \vlambda^*}_2 }}{{\lVert \bar{\vlambda} - \vlambda^* \rVert}_2}
    \,
    \sqrt{
    \frac{
       d + k_{\varphi} + 4
    }{
      d + k_{\varphi}
    }
    } \, \epsilon^{-1/4}.
  \end{align*}
\end{proofEnd}

\vspace{1.ex}

In particular, the following theorem establishes that BBVI with the STL estimator can achieve linear convergence under perfect variational family specification.
\vspace{0ex}

\begin{theoremEnd}[category=complexitybbvistl]{theorem}[\textbf{Complexity of Fixed Stepsize BBVI with STL}]\label{thm:projsgd_bbvistl_complexity}
  The last iterate \(\vlambda_T \in \Lambda_L\) of BBVI with the STL estimator and projected SGD with a fixed stepsize applied to a \(\mu\)-strongly log-concave and \(L\)-log-smooth posterior is \(\epsilon\)-close to \(\vlambda^* = \argmin_{\vlambda \in \Lambda_L} F\left(\vlambda\right)\) such that \(\mathbb{E}\norm{ \vlambda_T - \vlambda^* }_2^2 \leq \epsilon\) if
{%
\setlength{\abovedisplayskip}{.5ex} \setlength{\abovedisplayshortskip}{.5ex}
\setlength{\belowdisplayskip}{1.ex} \setlength{\belowdisplayshortskip}{1.ex}
  \begin{align*}
    T
    &\geq
    8 \kappa^2 \left(d + k_{\varphi} \right) \left(1 + \frac{1}{L^2} \sqrt{ \mathrm{D}_{\mathrm{F}^4}\left(q_{\vlambda^*}, \pi\right) } \frac{1}{\epsilon}\right)
    \log \left( 2 \Delta^2 \, \frac{1}{\epsilon} \right)
  \end{align*}
}%
  for some fixed stepsize \(\gamma\), where \(\Delta = {\lVert \vlambda_0 - \vlambda^*\rVert}_2\) is the distance to the optimum and \(\kappa = L / \mu\) is the condition number.
\end{theoremEnd}
\vspace{-1ex}
\begin{proofEnd}\label{proof:projsgd_bbvistl_complexity}
  As shown by \cref{thm:stl_upperbound}, the STL estimator with \(S = L\) satisfies an adaptive QV bound with the constants
  \begin{align*}
    \alpha_{\mathrm{STL}} &= 2 \left(d + k_{\varphi}\right) \left(2 + \delta\right)  L^2
      = 4 L^2 \left(d + k_{\varphi}\right) \left(1 + \frac{1}{2}\delta\right)  \\
    \beta_{\mathrm{STL}}  &= \left(2 d + k_{\varphi}\right) \left(1 + 2 \delta^{-1}\right)  \sqrt{\mathrm{D}_{\mathrm{F}^4}\left(q_{\vlambda^*}, \pi\right)}.
  \end{align*}
  Furthermore, for a \(\mu\)-strongly log-concave posterior and our variational parameterization,~\citet[Theorem 9]{domke_provable_2020} show that the ELBO is \(\mu\)-strongly convex.
  Thus, we can fully invoke \cref{thm:projsgd_stronglyconvex_adaptive_complexity} with 
  \begin{align*}
    \widetilde{\alpha} = 4 L^2 \left(d + k_{\varphi}\right), \qquad
    \widetilde{\beta}  = \left(2 d + k_{\varphi}\right) \sqrt{ \mathrm{D}_{\mathrm{F}^4}\left(q_{\vlambda^*}, \pi\right) },\quad\text{and}\quad
    C = \frac{1}{2}.
  \end{align*}
  This yields a lower bound on the number of iteration 
  \begin{align*}
    &\frac{2}{\mu^2} \max\left(\widetilde{\alpha} + 2 \widetilde{\beta} \frac{1}{\epsilon}, \; \frac{\mu^2}{4}\right) \log \left( 2 \norm{\vlambda_0 - \vlambda^*}_2^2 \, \frac{1}{\epsilon} \right)
    \\
    &=
    \frac{2}{\mu^2} \max\left(4 L^2 \left(d + k_{\varphi} \right) + 2 \left(2 d + k_{\varphi}\right) \sqrt{ \mathrm{D}_{\mathrm{F}^4}\left(q_{\vlambda^*}, \pi\right) } \frac{1}{\epsilon}, \; \frac{\mu^2}{4}\right) 
    \log \left( 2 {\lVert \vlambda_0 - \vlambda^*\rVert}_2^2 \, \frac{1}{\epsilon} \right),
\shortintertext{pulling out the \(L^2\) factor,}
    &=
    \frac{2 L^2}{\mu^2} \max\left(4 \left(d + k_{\varphi} \right) + 2 \frac{1}{L^2} \left(2 d + k_{\varphi}\right) \sqrt{ \mathrm{D}_{\mathrm{F}^4}\left(q_{\vlambda^*}, \pi\right) } \frac{1}{\epsilon}, \; \frac{\mu^2}{4 L^2}\right) 
    \log \left( 2 {\lVert \vlambda_0 - \vlambda^*\rVert}_2^2 \, \frac{1}{\epsilon} \right),
\shortintertext{and since \(\frac{\mu^2}{4 L^2} < \frac{1}{4}\) and the first argument is larger than 1 due to \(k_{\varphi} \geq 1\), the max operation is redundant such that}
    &=
    \frac{2 L^2}{\mu^2} \left(4 \left(d + k_{\varphi} \right) + 2  \frac{1}{L^2} \left(2 d + k_{\varphi}\right) \sqrt{ \mathrm{D}_{\mathrm{F}^4}\left(q_{\vlambda^*}, \pi\right) } \frac{1}{\epsilon}\right) 
    \log \left( 2 {\lVert \vlambda_0 - \vlambda^*\rVert}_2^2 \, \frac{1}{\epsilon} \right).
\shortintertext{Now, using the trivial fact \(2 d + k_{\varphi} < 2 d + 2 k_{\varphi}\) simplifies the bound as,}
    &<
    \frac{8 L^2}{\mu^2} \left(d + k_{\varphi} \right) \left(1 + \frac{1}{L^2} \sqrt{ \mathrm{D}_{\mathrm{F}^4}\left(q_{\vlambda^*}, \pi\right) } \frac{1}{\epsilon}\right)
    \log \left( 2 {\lVert \vlambda_0 - \vlambda^*\rVert}_2^2 \, \frac{1}{\epsilon} \right)
    \\
    &=
    8 \kappa^2 \left(d + k_{\varphi} \right) \left(1 + \frac{1}{L^2} \sqrt{ \mathrm{D}_{\mathrm{F}^4}\left(q_{\vlambda^*}, \pi\right) } \frac{1}{\epsilon}\right) 
    \log \left( 2 {\lVert \vlambda_0 - \vlambda^*\rVert}_2^2 \, \frac{1}{\epsilon} \right).
  \end{align*}
  
  The optimal \(\delta\) is given as
  \begin{align*}
    \delta 
    = \frac{2}{\epsilon} \frac{\widetilde{\beta}}{\widetilde{\alpha}} C^{-1}
    = 
    \frac{2}{\epsilon}  
    \frac{
      \left(2 d + k_{\varphi}\right) \sqrt{ \mathrm{D}_{\mathrm{F}^4}\left(q_{\vlambda^*}, \pi\right) }
    }{
      4 L^2 \left(d + k_{\varphi}\right)
    } \,
    2
    =
    \frac{
      4
    }{
      \epsilon
    }
    \frac{
      \sqrt{ \mathrm{D}_{\mathrm{F}^4}\left(q_{\vlambda^*}, \pi\right) } 
    }{
      L^2
    }
    \frac{2 d + k_{\varphi} }{d + k_{\varphi}}.
  \end{align*}
\end{proofEnd}

\begin{theoremEnd}[all end, category=complexitybbvistldec]{theorem}[\textbf{Complexity of Decreasing Stepsize BBVI with STL}]\label{thm:projsgd_bbvistl_decstepsize_complexity}
  The last iterate \(\vlambda_T \in \Lambda_L\) of BBVI with the STL estimator and projected SGD with a decreasing stepsize schedule applied to a \(\mu\)-strongly log-concave and \(L\)-log-smooth posterior is \(\epsilon\)-close to \(\vlambda^* = \argmin_{\vlambda \in \Lambda_L} F\left(\vlambda\right)\) such that \(\mathbb{E}\norm{ \vlambda_T - \vlambda^* }_2^2 \leq \epsilon\) if
  \begin{align*}
    T
    \geq
    32
    \kappa^2
    \left(d + k_{\varphi}\right)
    \Bigg(
    &\frac{
      \sqrt{ \mathrm{D}_{\mathrm{F}^4}\left(q_{\vlambda^*},\pi\right) }
    }{
      L^2
    }
    \frac{1}{\epsilon} 
    +
    \frac{1}{\sqrt{2}}
    \sqrt{
      \norm{\vlambda_0 - \vlambda^*}_2
    }
    \,
    \frac{
      {\left( \mathrm{D}_{\mathrm{F}^4}\left(q_{\vlambda^*},\pi\right) \right)}^{1/4}
    }{
      L
    }
    \,
    \frac{1}{\epsilon^{3/4}}
    +
    \norm{\vlambda_0 - \vlambda^*}_2 
    \frac{1}{\sqrt{\epsilon}}
    \Bigg),
  \end{align*}
  for some decreasing stepsize schedule \(\gamma_1 \geq \ldots \geq \gamma_T\), where \(\kappa = L / \mu\) is the condition number.
\end{theoremEnd}
\begin{proofEnd}\label{proof:projsgd_bbvistl_decstepsize_complexity}
  As shown by \cref{thm:stl_upperbound}, the STL estimator with \(S = L\) satisfies an adaptive QV bound with the constants
  \begin{align*}
    \alpha_{\mathrm{STL}} &= 2 \left(d + k_{\varphi}\right) \left(2 + \delta\right)  L^2 
      = 4 L^2 \left(d + k_{\varphi}\right) \left(1 + \frac{1}{2}\delta\right)  \\
    \beta_{\mathrm{STL}}  &= \left(2 d + k_{\varphi}\right) \left(1 + 2 \delta^{-1}\right)  \sqrt{\mathrm{D}_{\mathrm{F}^4}\left(q_{\vlambda^*}, \pi\right)}.
  \end{align*}
  Furthermore, for a \(\mu\)-strongly log-concave posterior and our variational parameterization,~\citet[Theorem 9]{domke_provable_2020} show that the ELBO is \(\mu\)-strongly convex.
  Thus, we can invoke \cref{thm:projsgd_stronglyconvex_decstepsize_adaptive_complexity} with 
  \begin{align*}
    \widetilde{\alpha} = 4 L^2 \left(d + k_{\varphi}\right),
    \qquad
    \widetilde{\beta}  = \left(2 d + k_{\varphi}\right) \sqrt{ \mathrm{D}_{\mathrm{F}^4}\left(q_{\vlambda^*}, \pi\right) },
    \qquad\text{and}\qquad
    C = \frac{1}{2}.
  \end{align*}
  This yields a lower bound on the number of iterations: 
  \begin{align*}
    &
    \frac{16 \widetilde{\beta} }{ \mu^2 } \frac{1}{\epsilon} 
    +
    \frac{16 \sqrt{2}}{ \mu^2 }
    \sqrt{
      \norm{\vlambda_0 - \vlambda^*}_2
    }
    \,
    \sqrt{
      \widetilde{\alpha} \widetilde{\beta}
    }
    \,
    \frac{1}{\epsilon^{3/4}}
    +
    \frac{8 \widetilde{\alpha} \, \norm{\vlambda_0 - \vlambda^*}_2 }{ \mu^2}
    \frac{1}{\sqrt{\epsilon}}
    \\
    &=
    \frac{16
      \left(2 d + k_{\varphi}\right) \sqrt{ {D}_{\mathrm{F}^4}^* }
    }{ \mu^2 } \frac{1}{\epsilon} 
    +
    \frac{16 \sqrt{2}}{ \mu^2 }
    \sqrt{
      \norm{\vlambda_0 - \vlambda^*}_2
    }
    \,
    \sqrt{
      4 L^2 \left(d + k_{\varphi}\right)
      \left(2 d + k_{\varphi}\right) \sqrt{ {D}_{\mathrm{F}^4}^* }
    }
    \,
    \frac{1}{\epsilon^{3/4}}
    \\
    &\qquad
    +
    \frac{32 L^2 \left(d + k_{\varphi}\right) \, \norm{\vlambda_0 - \vlambda^*}_2 }{ \mu^2}
    \frac{1}{\sqrt{\epsilon}},
\shortintertext{using the the trivial bound \(2 d + k_{\varphi} < 2 d + 2 k_{\varphi}\),}
    &<
    \frac{32
      \left(d + k_{\varphi}\right) \sqrt{ {D}_{\mathrm{F}^4}^* }
    }{ \mu^2 } \frac{1}{\epsilon} 
    +
    \frac{16 \sqrt{2}}{ \mu^2 }
    \sqrt{
      \norm{\vlambda_0 - \vlambda^*}_2
    }
    \,
    \sqrt{
      8 L^2 \left(d + k_{\varphi}\right) \left(d + k_{\varphi}\right) \sqrt{ {D}_{\mathrm{F}^4}^* }
    }
    \,
    \frac{1}{\epsilon^{3/4}}
    \\
    &\qquad
    +
    \frac{32 L^2 \left(d + k_{\varphi}\right) \, \norm{\vlambda_0 - \vlambda^*}_2 }{ \mu^2}
    \frac{1}{\sqrt{\epsilon}},
\shortintertext{pulling out the \(32 \left(d + k_{\varphi}\right) L^2 / \mu^2\) factors,}
    &=
    32
    \frac{L^2}{\mu^2} 
    \left(d + k_{\varphi}\right)
    \left(
    \frac{
      \sqrt{ {D}_{\mathrm{F}^4}^* }
    }{
      L^2
    }
    \frac{1}{\epsilon} 
    +
    \frac{1}{\sqrt{2}}
    \sqrt{
      \norm{\vlambda_0 - \vlambda^*}_2
    }
    \,
    \frac{
      {\left({D}_{\mathrm{F}^4}^*\right)}^{1/4}
    }{
      L
    }
    \,
    \frac{1}{\epsilon^{3/4}}
    +
    \norm{\vlambda_0 - \vlambda^*}_2 
    \frac{1}{\sqrt{\epsilon}}
    \right)
    \\
    &=
    32
    \kappa^2
    \left(d + k_{\varphi}\right)
    \left(
    \frac{
      \sqrt{ D_{\mathrm{F}^4}^* }
    }{
      L^2
    }
    \frac{1}{\epsilon} 
    +
    \frac{1}{\sqrt{2}}
    \sqrt{
      \norm{\vlambda_0 - \vlambda^*}_2
    }
    \,
    \frac{
      {\left(D_{\mathrm{F}^4}^*\right)}^{1/4}
    }{
      L
    }
    \,
    \frac{1}{\epsilon^{3/4}}
    +
    \norm{\vlambda_0 - \vlambda^*}_2 
    \frac{1}{\sqrt{\epsilon}}
    \right),
  \end{align*}
  where we have denoted \(D_{\mathrm{F}^4}^* = \mathrm{D}_{\mathrm{F}^4}\left(q_{\vlambda^*},\pi\right)\).
  
  Also, the optimal \(\delta\) is given as
  \begin{align*}
    \delta 
    &=
    \frac{
      \sqrt{ \norm{\vlambda_0 - \vlambda^*}_2 }
      \,
      \epsilon^{1/4}
      \,
      \sqrt{\widetilde{\alpha}}
    }{
      \sqrt{2} \, C 
      \sqrt{\widetilde{\beta}}
    }
    \\
    &=
    \frac{
      \sqrt{ \norm{\vlambda_0 - \vlambda^*}_2 }
      \,
      \epsilon^{1/4}
      \,
      \sqrt{4 L^2 \left(d + k_{\varphi}\right)}
    }{
      2^{-1} \, \sqrt{2}
      \sqrt{ \left(2 d + k_{\varphi}\right) \sqrt{ D_{\mathrm{F}^4}^* } }
    }
    \\
    &=
    2 \sqrt{2} L
    \sqrt{ \norm{\vlambda_0 - \vlambda^*}_2 }
    \,
    \sqrt{\frac{d + k_{\varphi}}{2 d + k_{\varphi}}}
    \,
    {\left(D_{\mathrm{F}^4}^*\right)}^{-1/2}
    \,
    \epsilon^{1/4}
    \\
    &=
    2 \sqrt{2} L
    \,
    \sqrt{ 
    \frac{
      \norm{\vlambda_0 - \vlambda^*}_2
    }{
      \mathrm{D}_{\mathrm{F}^4}\left(q_{\vlambda^*},\pi\right)
    }
    }
    \sqrt{\frac{d + k_{\varphi}}{2 d + k_{\varphi}}}
    \,
    \epsilon^{1/4}.
  \end{align*}
\end{proofEnd}

\vspace{1ex}
\begin{corollary}[\textbf{Linear Convergence of BBVI with STL}]
  If the variational family is perfectly specified such that \( \mathrm{D}_{\mathrm{F}^4}\left(q_{\vlambda}^*, \pi\right) = 0\) for \(\vlambda^* = \argmin_{\vlambda \in \Lambda_L} F\left(\vlambda\right)\), then BBVI with the STL estimator converges linearly with a complexity of \(\mathcal{O}\left(d \kappa^2 \log \left( 1 / \epsilon \right) \right)\).
\end{corollary}


\vspace{1ex}
\begin{remark}
  Convergence is slowed when using a decreasing step size schedule, as shown in \cref{thm:projsgd_bbvistl_decstepsize_complexity}.
  Thus, one does not achieve a linear convergence rate under this schedule even if the variational family is perfectly specified. 
  However, when the variational family is misspecified, this achieves a better rate of \(\mathcal{O}\left(1/\epsilon\right)\) compared to the \(\mathcal{O}\left(1/\epsilon \log 1/\epsilon\right)\) of \cref{thm:projsgd_bbvistl_complexity}.
\end{remark}

\vspace{1ex}
\begin{remark}[\textbf{Variational Family Misspecification}]\label{remark:misspecification}
  Under variational family misspecification, STL has an \(\mathcal{O}\left(1/\epsilon\right)\) dependence on the 4th order Fisher divergence \(\mathrm{D}_{\mathrm{F}^4}\left(q_{\vlambda^*}, \pi\right) > 0\).
  To compare the computational performance of CFE and STL in this setting, one needs to compare \( L^{-2} \sqrt{\mathrm{D}_{\mathrm{F}^4}\left(q_{\vlambda^*}, \pi\right)}\) versus \({\lVert \bar{\vlambda} - \vlambda^* \rVert}_2^2\).
\end{remark}

\vspace{1ex}
\begin{remark}
  \cref{thm:stl_upperbound_mf} also implies that the mean-field parameterization improves the dimension dependence to a complexity of \(\mathcal{O}\left( \sqrt{d} \kappa^2 \log \left( 1 / \epsilon \right) \right)\).
\end{remark}






\subsection{Should we stick the landing?}
When the variational family is misspecified, it is hard to tell \textit{when} STL would be superior to CFE; the Fisher-Hyv\"arinen divergence and the posterior variance are fundamentally unrelated quantities.
Furthermore, the Fisher-Hyv\"arinen divergence is hard to interpret apart from some relationships with other divergences~\citep{huggins_practical_2018}.
Thus, we conclude by providing a characterization of the Fisher-Hyv\"arinen divergence.

Our final analysis will focus on Gaussian posteriors and the mean-field Gaussian family.
In practice, the STL estimator becomes infeasible to use with full-rank variational families as each evaluation of the log-density \(\log q_{\vlambda}\) involves a back-substitution with a \(\mathcal{O}\left(d^3\right)\) cost and numerical stability becomes a concern.
Therefore, studying the effect of misspecification of mean-field is particularly relevant.

\vspace{1ex}

\begin{theoremEnd}[all end, category=stlgaussianfisher]{lemma}\label{thm:gaussian_fisher_divergence}
  For \(\pi = \mathcal{N}\left(\vmu, \mSigma\right)\) and \(q = \mathcal{N}\left(\vm, \mC \mC^{\top} \right)\), the Fisher-Hyv\"arinen divergence is
  \[
    \DHF{q}{\pi}
    =
    {\lVert \mSigma^{-1} \mC - \mC^{-\top} \rVert}_{\mathrm{F}}^2
    +
    {\lVert \mSigma^{-1}\left(\vm  - \vmu\right) \rVert}_{2}^2.
  \]
\end{theoremEnd}
\begin{proofEnd}
The result is straightforward using the reparameterization representation of the Gaussian.
That is,
\[
  \nabla \log \pi\left(\rvvz\right)
  =
  \nabla \log \pi\left(\mathcal{T}_{\vlambda}\left(\rvvu\right) \right)
  =
  \mSigma^{-1}\left(\mathcal{T}_{\vlambda}\left(\rvvu\right) - \vmu\right).
\]
Using this, we have
  \begin{align*}
    \DHF{q}{\pi}
    &=
    \mathbb{E}_{\rvvz \sim q} \norm{ \nabla \log \pi\left(\rvvz\right) - \nabla \log q\left(\rvvz\right) }_{2}^2
    \\
    &=
    \mathbb{E} \norm{ \mSigma^{-1} \left(\mC \rvvu + \vm  - \vmu\right) - {\left( \mC\mC^{\top} \right)}^{-1} \left(\mC \rvvu + \vm - \vm\right) }_{2}^2
    \\
    &=
    \mathbb{E} \norm{ \mSigma^{-1} \left(\mC \rvvu + \vm  - \vmu\right) - {\left( \mC\mC^{\top} \right)}^{-1} \mC \rvvu }_{2}^2
    \\
    &=
    \mathbb{E} \norm{ \mSigma^{-1} \left(\mC \rvvu + \vm  - \vmu\right) - \mC^{-\top} \rvvu }_{2}^2,
\shortintertext{grouping the terms involving \(\mC\),} 
    &=
    \mathbb{E} \norm{ \left( \mSigma^{-1} \mC - \mC^{-\top} \right) \rvvu  + \mSigma^{-1}\left(\vm  - \vmu\right) }_{2}^2,
\shortintertext{expanding the quadratic,} 
    &=
    \mathbb{E} \norm{ \left( \mSigma^{-1} \mC - \mC^{-\top} \right) \rvvu }_2^2
    +
    2 \inner{ \left( \mSigma^{-1} \mC - \mC^{-\top} \right) \mathbb{E} \rvvu }{ \mSigma^{-1}\left(\vm  - \vmu\right) }
    +
    {\lVert \mSigma^{-1}\left(\vm  - \vmu\right) \rVert}_{2}^2,
\shortintertext{applying \cref{assumption:symmetric_standard},} 
    &=
    \mathbb{E} \norm{ \left( \mSigma^{-1} \mC - \mC^{-\top} \right) \rvvu }_2^2
    +
    {\lVert \mSigma^{-1}\left(\vm  - \vmu\right) \rVert}_{2}^2.
  \end{align*}

The expectation term can be simplified as
  \begin{align*}
    \mathbb{E} \norm{ \left( \mSigma^{-1} \mC - \mC^{-\top} \right) \rvvu }_2^2
    &=
    \mathbb{E} \mathrm{tr} \left( \rvvu^{\top} {\left( \mSigma^{-1} \mC - \mC^{-\top} \right)}^{\top} \left( \mSigma^{-1} \mC - \mC^{-\top} \right) \rvvu \right),
\shortintertext{rotating the elements of the trace,} 
    &=
    \mathrm{tr} \left( {\left( \mSigma^{-1} \mC - \mC^{-\top} \right)}^{\top} \left( \mSigma^{-1} \mC - \mC^{-\top} \right) \mathbb{E} \rvvu \rvvu^{\top} \right),
\shortintertext{applying \cref{assumption:symmetric_standard},} 
    &=
    \mathrm{tr} \left( {\left( \mSigma^{-1} \mC - \mC^{-\top} \right)}^{\top} \left( \mSigma^{-1} \mC - \mC^{-\top} \right) \right)
    \\
    &=
    \norm{ \mSigma^{-1} \mC - \mC^{-\top} }_{\mathrm{F}}^2.
  \end{align*}
\end{proofEnd}

\begin{theoremEnd}[all end, category=gaussianklmeanfield]{lemma}\label{thm:gaussian_kl_meanfield}
  Let \(\pi = \mathcal{N}\left(\vmu, \mSigma\right)\) and \(\mathcal{Q}\) be the mean-field Gaussian variational family.
  Then, the solution of the KL divergence minimization problem
  \[
     q_* = \argmin_{q \in \mathcal{Q}}  \;  \DKL{q}{\pi}, 
  \]
  where \(q_* = \mathcal{N}\left(\vm_*, \mC_* \mC_*^{\top}\right)\) is given as
  \[
    \vm_* = \vmu, \qquad \mC_* = {\mathrm{diag}\left(\mSigma\right)}^{\nicefrac{1}{2}}.
  \]
\end{theoremEnd}
\begin{proofEnd}
  Consider that the KL divergence between Gaussian distributions is given as 
  \begin{align*}
    \mathcal{L}\left(\vm, \mC\right)
    = \DKL{q}{\pi} 
    = \frac{1}{2} \left( 
      {\left(\vm - \vmu\right)} \mSigma^{-1} {\left(\vm - \vmu\right)}
      +
      \log\frac{\abs{\mSigma}}{\abs{\mC\mC^{\top}}}
      +
      \operatorname{tr}\left( \mSigma^{-1} \mC\mC^{\top} \right)
      -
      d
    \right).
  \end{align*}
  Firstly, it is clear that \(\vm = \vm_* = \vmu\) minimizes \(\DKL{q}{\pi}\) with respect to \(\vm\) regardless of \(\mC\).
  Then, we have
  \begin{align*}
    \mathcal{L}\left(\vm_*, \mC\right)
    &=
    \frac{1}{2} \left( 
      \log\frac{\abs{\mSigma}}{\abs{\mC\mC^{\top}}}
      +
      \operatorname{tr}\left( \mSigma^{-1} \mC\mC^{\top} \right)
      -
      d
    \right)
    \propto
    -
    \log \abs{\mC\mC^{\top}}
    +
    \operatorname{tr}\left( \mSigma^{-1} \mC\mC^{\top} \right).
  \end{align*}
  When \(\mC\) is a diagonal matrix, taking the partial derivative with respect to \(\mC\) yields
  \begin{align*}
    \frac{\partial\mathcal{L}}{\partial \mC}  \Big\lvert_{\vm = \vm_*}
    =
    -2 \, \mC^{-1}
    +
    2 \operatorname{diag}\left(\mSigma^{-1}\right) \mC.
  \end{align*}
  The first-order optimality condition with respect to \(\mC\) is then
  \[
    {\left( \mC\mC \right)}^{-1} = \operatorname{diag}\left(\mSigma^{-1}\right).
  \]
  Since \(\mSigma\) is always positive definite, its diagonal elements are always strictly positive.
  Therefore, the unique solution \(\mC^*\) is
  \[
    \mC_* = {\operatorname{diag}\left(\mSigma\right)}^{\nicefrac{1}{2}}.
  \]
\end{proofEnd}
\begin{theoremEnd}[category=stlgaussian]{proposition}\label{thm:fisher_bound}
  Let \(\pi = \mathcal{N}\left(\vmu, \mSigma\right)\) and \(\mathcal{Q}\) be the mean-field Gaussian variational family.
  Then, the Fisher-Hyv\"arinen divergence of the KL minimizer
{%
\setlength{\abovedisplayskip}{.5ex} \setlength{\abovedisplayshortskip}{.5ex}
\setlength{\belowdisplayskip}{-.5ex} \setlength{\belowdisplayshortskip}{-.5ex}
  \[
    q_* = \argmin_{q \in \mathcal{Q}} \DKL{q}{\pi}
  \] 
}%
  is bounded as
{%
\setlength{\abovedisplayskip}{.5ex} \setlength{\abovedisplayshortskip}{.5ex}
\setlength{\belowdisplayskip}{1.ex} \setlength{\belowdisplayshortskip}{1.ex}
  \begin{align*}
    &{\lambda_{\mathrm{max}}\left(\mD\right)}^{-1} 
    {\lVert \mR^{-1}  - \boldupright{I} \rVert}_{\mathrm{F}}^2
    \\
    &\qquad\leq
    \DHF{q_*}{\pi}
    \leq
    {\lambda_{\mathrm{min}}\left(\mD\right)}^{-1} 
    {\lVert \mR^{-1}  - \boldupright{I} \rVert}_{\mathrm{F}}^2,
  \end{align*}
}%
  where \(\mD = \mathrm{diag}\left(\mSigma\right)\) and \(\mR\) is the correlation matrix of \(\pi\) such that \(\mSigma = \mD \mR \mD\).
\end{theoremEnd}
\vspace{-1ex}
\begin{proofEnd}
  First, the Fisher-Hyv\"arinen divergence between Gaussians is given in \cref{thm:gaussian_fisher_divergence} as
  \begin{align*}
    \DHF{q}{\pi}
    =
    {\lVert \mSigma^{-1} \mC - \mC^{-\top} \rVert}_{\mathrm{F}}^2
    +
    {\lVert \mSigma^{-1}\left(\vm  - \vmu\right) \rVert}_{2}^2.
  \end{align*}
  Plugging the KL minimizer \(q_{*}\) given in \cref{thm:gaussian_kl_meanfield}, 
  \begin{align}
    \DHF{q_{*}}{\pi}
    &=
    {\lVert \mSigma^{-1} \mC_* - \mC_*^{-\top} \rVert}_{\mathrm{F}}^2
    +
    {\lVert \mSigma^{-1}\left(\vm_* - \vmu\right) \rVert}_{2}^2
    \nonumber
    \\
    &=
    {\lVert \mSigma^{-1} \mC_* - \mC_*^{-1} \rVert}_{\mathrm{F}}^2.
    \label{eq:stl_gaussian_sandwich_1}
  \end{align}

  From here, we can pull out a \(\mC_*^{-1}\) factor as
  \begin{align}
    {\lVert \mSigma^{-1} \mC_* - \mC_*^{-1} \rVert}_{\mathrm{F}}^2
    =
    {\lVert \mC_*^{-1} \left( \mC_* \mSigma^{-1} \mC - \boldupright{I} \right) \rVert}_{\mathrm{F}}^2.
    \label{eq:stl_gaussian_sandwich_2}
  \end{align}
  And from the property of the Frobenius norm,
  \begin{alignat*}{6}
    & &\quad
    {\lambda_{\mathrm{min}}\left(\mC_*^{-1}\right)}^2 {\lVert \mC_* \mSigma^{-1} \mC_* - \boldupright{I} \rVert}_{\mathrm{F}}^2
    \quad&\leq&\quad
    \, {\lVert \mC_*^{-1} \left( \mC_* \mSigma^{-1} \mC_* - \boldupright{I} \right) \rVert}_{\mathrm{F}}^2 \;
    \quad&\leq&\quad
    {\lambda_{\mathrm{max}}\left(\mC_*^{-1}\right)}^2 {\lVert \mC_* \mSigma^{-1} \mC_* - \boldupright{I} \rVert}_{\mathrm{F}}^2,
\shortintertext{inverting the singular values,}
    &\Leftrightarrow&\quad
    {\lambda_{\mathrm{max}}\left(\mC_*\right)}^{-2} {\lVert \mC_* \mSigma^{-1} \mC_* - \boldupright{I} \rVert}_{\mathrm{F}}^2
    \quad&\leq&\quad
    \,{\lVert \mC_*^{-1} \left( \mC_* \mSigma^{-1} \mC_* - \boldupright{I} \right) \rVert}_{\mathrm{F}}^2 \,
    \quad&\leq&\quad
    {\lambda_{\mathrm{min}}\left(\mC_*\right)}^{-2} {\lVert \mC_* \mSigma^{-1} \mC_* - \boldupright{I} \rVert}_{\mathrm{F}}^2,
\shortintertext{by \cref{eq:stl_gaussian_sandwich_1,eq:stl_gaussian_sandwich_2},}
    &\Leftrightarrow&\quad
    {\lambda_{\mathrm{max}}\left(\mC_*\right)}^{-2} {\lVert \mC_* \mSigma^{-1} \mC_* - \boldupright{I} \rVert}_{\mathrm{F}}^2
    \quad&\leq&\quad
    \,\DHF{q_*}{\pi}\,
    \quad&\leq&\quad
    {\lambda_{\mathrm{min}}\left(\mC_*\right)}^{-2} {\lVert \mC_* \mSigma^{-1} \mC_* - \boldupright{I} \rVert}_{\mathrm{F}}^2.
  \end{alignat*}
    
Denoting \(\mD = \operatorname{diag}\left(\mSigma\right)\), we know that \(\mC_* = \mD^{\nicefrac{1}{2}}\). Then,
  \begin{alignat*}{2}
    {\lambda_{\mathrm{max}}\left(\mD\right)}^{-1} {\lVert \mC_* \mSigma^{-1} \mC_* - \boldupright{I} \rVert}_{\mathrm{F}}^2
    \quad\leq\quad
    \DHF{q_*}{\pi}
    \quad\leq\quad
    {\lambda_{\mathrm{min}}\left(\mD\right)}^{-1} {\lVert \mC_* \mSigma^{-1} \mC_* - \boldupright{I} \rVert}_{\mathrm{F}}^2.
  \end{alignat*}
  
  Clearly, the behavior of the Fisher divergence is fully determined by the term
  \begin{align*}
      {\lVert \mC_{*} \mSigma^{-1} \mC_* - \boldupright{I} \rVert}_{\mathrm{F}}^2.
  \end{align*}
  To further analyze this quantity, notice that the correlation matrix \(\mR\) is related with the covariance \(\mSigma\) as
  \[
     \mSigma 
     = {\operatorname{diag}\left(\mSigma\right)}^{\nicefrac{1}{2}} \, \mR \, {\operatorname{diag}\left(\mSigma\right)}^{\nicefrac{1}{2}}
     = \mC_* \mR \mC_*.
  \]
  Then, it immediately follows that 
  \begin{align*}
      {\lVert \mC_{*} \mSigma^{-1} \mC_* - \boldupright{I} \rVert}_{\mathrm{F}}^2
      =
      {\lVert \mC_{*} {\left( \mC_* \mR \mC_* \right)}^{-1} \mC_* - \boldupright{I} \rVert}_{\mathrm{F}}^2
      =
      {\lVert \mR^{-1}  - \boldupright{I} \rVert}_{\mathrm{F}}^2.
  \end{align*}
\end{proofEnd}

\vspace{1ex}

\begin{remark}
For Gaussians, the 4th-order Fisher-Hyv\"arinen divergence term in \cref{thm:stl_upperbound} can be replaced by its 2nd-order counterpart.
Thus, combined with \cref{thm:stl_lowerbound}, the 2nd-order Fisher-Hyv\"arinen divergence fully characterizes the variance of STL.
\end{remark}

\vspace{1ex}
\begin{remark}
  \cref{thm:fisher_bound} implies that, when approximating a full-rank Gaussian with a mean-field Gaussian, the value of the Fisher-Hyv\"arinen divergence is tightly characterized by the degree of correlation in the posterior; it will increase indefinitely as the posterior correlation matrix becomes singular.
\end{remark}

\vspace{1ex}
\begin{remark}
  We have provided a sufficient condition for the STL estimator to perform poorly compared to the CFE estimator. 
  It is foreseeable that alternative types of model misspecification abundant in practice should yield additional sufficient conditions, \textit{i.e.}, tail mismatch, but we leave this to future works.
\end{remark}

\section{DISCUSSIONS}

\vspace{-.5ex}
\paragraph{Empirically Comparing Estimators}
From our analysis and that of \citet{domke_provable_2023}, it is apparent that for a QV gradient estimator, \(\alpha\) and \(\beta\) sufficiently characterize its behavior on log-concave posteriors: \(\alpha\) characterizes the convergence speed, while \(\beta\) determines the complexity with respect to \(\epsilon\).
It is conceivable that \textit{estimating} these quantities in practical settings would provide a principled way to compare and evaluate different estimators.
Previously, the signal-to-noise (SNR) ratio have been popularized by~\citet{rainforth_tighter_2018}, and since been used by, for example, by \cite{geffner_difficulty_2021,fujisawa_multilevel_2021,rudner_signaltonoise_2021}.
In contrast to the QV coefficients, a \textit{constant} SNR relates with convergence only through the expected strong growth condition~\citep{solodov_incremental_1998,vaswani_fast_2019,schmidt_fast_2013}, which requires strong assumptions to hold (perfect variational family specification, strong log-concavity).
The QV coefficients, \(\alpha\) and \(\beta\), on the other hand, apply to a broader range of settings.

\vspace{-1ex}
\paragraph{Conclusions}
We have analyzed the sticking-the-landing (STL) estimator by~\citet{roeder_sticking_2017}.
When the variational family is perfectly specified, our complexity guarantees automatically guarantees a logarithmic complexity.
Also, from the results of \citet{domke_provable_2019} and \cref{thm:cfe_upperbound} it is known that the gradient variance of CFE at the optimum depends on the mode mismatch \(\norm{\vm^* - \bar{\vz}}_2^2\) plus the variational posterior variance \(\norm{\mC^*}_{\mathrm{F}}^2\).
We show that the STL estimator instead depends on the Fisher-Hyv\"arinen divergence of the variational posterior.
Furthermore, our work demonstrates that it is possible to rigorously show that control variates can accelerate the convergence of BBVI.
It will be interesting to analyze and compare the existing control variates by~\citet{miller_reducing_2017, geffner_approximation_2020, wang_joint_2024}.


\clearpage
\subsubsection*{Acknowledgements}
The authors sincerely thank Jisu Oh (NCSU) for thoroughly proofreading the paper, Justin Domke (UMass Amherst) for discussions on the concurrent results, Kaiwen Wu (UPenn) for helpful discussions, Xi Wang (UMass Amherst) for pointing out a typo, and the anonymous reviewers for comments that improved the readability of the work.

K. Kim was supported by a gift from AWS AI to Penn Engineering's ASSET Center for Trustworthy AI; Y.-A. Ma was funded by the NSF Grants [NSF-SCALE MoDL-2134209] and [NSF-CCF-2112665 (TILOS)], the U.S. Department Of Energy, Office of Science, as well as CDC-RFA-FT-23-0069 from the CDC’s Center for Forecasting and Outbreak Analytics; J. R. Gardner was supported by NSF award [IIS-2145644].

\bibliographystyle{plainnat}
\bibliography{references}

\clearpage

\section*{Checklist}

 \begin{enumerate}
 \item For all models and algorithms presented, check if you include:
 \begin{enumerate}
 \item A clear description of the mathematical setting, assumptions, algorithm, and/or model. \\
     \textbf{\textcolor{ForestGreen}{Yes}}. See~\cref{section:preliminaries}.
   \item An analysis of the properties and complexity (time, space, sample size) of any algorithm. \\
     \textbf{\textcolor{ForestGreen}{Yes}}. See~\cref{section:bbvicomplexity}.
   \item (Optional) Anonymized source code, with specification of all dependencies, including external libraries. \\
     \textbf{\textcolor{black!70}{Not Applicable}}.
 \end{enumerate}

 \item For any theoretical claim, check if you include:
 \begin{enumerate}
   \item Statements of the full set of assumptions of all theoretical results. \\
     \textbf{\textcolor{ForestGreen}{Yes}}. See the theorem statements and \cref{section:preliminaries}
   \item Complete proofs of all theoretical results. \\
     \textbf{\textcolor{ForestGreen}{Yes}}. See \cref{section:proofs}.
   \item Clear explanations of any assumptions. \\
     \textbf{\textcolor{ForestGreen}{Yes}}. See \cref{section:definitions,section:preliminaries} and the main text.
 \end{enumerate}

 \item For all figures and tables that present empirical results, check if you include:
 \begin{enumerate}
   \item The code, data, and instructions needed to reproduce the main experimental results (either in the supplemental material or as a URL). \\
     \textbf{\textcolor{black!70}{Not Applicable}}.
   \item All the training details (e.g., data splits, hyperparameters, how they were chosen). \\
     \textbf{\textcolor{black!70}{Not Applicable}}.
   \item A clear definition of the specific measure or statistics and error bars (e.g., with respect to the random seed after running experiments multiple times). \\
     \textbf{\textcolor{black!70}{Not Applicable}}.
   \item A description of the computing infrastructure used. (e.g., type of GPUs, internal cluster, or cloud provider). \\
     \textbf{\textcolor{black!70}{Not Applicable}}.
 \end{enumerate}

 \item If you are using existing assets (e.g., code, data, models) or curating/releasing new assets, check if you include:
 \begin{enumerate}
   \item Citations of the creator If your work uses existing assets. \\
     \textbf{\textcolor{black!70}{Not Applicable}}.
   \item The license information of the assets, if applicable.\\
     \textbf{\textcolor{black!70}{Not Applicable}}.
   \item New assets either in the supplemental material or as a URL, if applicable. \\
     \textbf{\textcolor{black!70}{Not Applicable}}.
   \item Information about consent from data providers/curators. \\
     \textbf{\textcolor{black!70}{Not Applicable}}.
   \item Discussion of sensible content if applicable, e.g., personally identifiable information or offensive content. \\
     \textbf{\textcolor{black!70}{Not Applicable}}.
 \end{enumerate}

 \item If you used crowdsourcing or conducted research with human subjects, check if you include:
 \begin{enumerate}
   \item The full text of instructions given to participants and screenshots.  \\
     \textbf{\textcolor{black!70}{Not Applicable}}.
   \item Descriptions of potential participant risks, with links to Institutional Review Board (IRB) approvals if applicable.  \\
     \textbf{\textcolor{black!70}{Not Applicable}}.
   \item The estimated hourly wage paid to participants and the total amount spent on participant compensation. \\
     \textbf{\textcolor{black!70}{Not Applicable}}.
 \end{enumerate}

 \end{enumerate}

\newpage
\appendix


\onecolumn

\newpage
{\hypersetup{linkcolor=black}
\tableofcontents
}

\newpage
\section{OVERVIEW OF THEOREMS}\label{section:overviewtheorems}

{\hypersetup{linkcolor=black}
\begin{table*}[h!]
\vspace{-2ex}
  \centering
\caption{Overview of Results}\label{table:theorems}
\vspace{2ex}
  {\normalsize
\makegapedcells
\renewcommand{\arraystretch}{1.3}
\begin{tabularx}{\linewidth}{lXcl}
    \toprule
    \textbf{}
    & \multicolumn{1}{c}{\textbf{DESCRIPTION}}
    & \multicolumn{1}{c}{\textbf{RESULT}}
    & \multicolumn{1}{c}{\textbf{SECTION}}
    \\ \midrule
    %
    \multicolumn{4}{l}{\textit{Gradient Variance Bounds}}
    \\
    & Smoothness of the entropy with triangular scale parameterization
    & \cref{thm:entropy_smoothness}
    & \labelcref{proof:entropy_smoothness}
    \\ \midrule
    \multicolumn{4}{l}{\textit{Gradient Variance Bounds}}
    \\
    & \textbf{Upper bound} for the gradient variance of the \textbf{STL} estimator with the \textbf{full-rank} parameterization
    & \cref{thm:stl_upperbound}
    & \labelcref{proof:stl_upperbound}
    \\
    & \textbf{Upper bound} for the gradient variance of the \textbf{STL} estimator with the \textbf{mean-field} parameterization
    & \cref{thm:stl_upperbound_mf}
    & \labelcref{proof:stl_upperbound_mf}
    \\
    & \textbf{Lower bound} for the gradient variance of the \textbf{STL} estimator with the \textbf{full-rank} parameterization
    & \cref{thm:stl_lowerbound}
    & \labelcref{proof:stl_lowerbound}
    \\
    & \textbf{Worst case lower bound} (unimprovability) for the gradient variance of the \textbf{STL} estimator with the \textbf{full-rank} parameterization
    & \cref{thm:stl_lowerbound_unimprovability}
    & \labelcref{proof:stl_lowerbound_unimprovability}
    \\
    & \textbf{Upper bound} for the gradient variance of the \textbf{CFE} estimator with the \textbf{full-rank} parameterization
    & \cref{thm:cfe_upperbound}
    & \labelcref{proof:cfe_upperbound}
    \\
    & \textbf{Upper bound} for the gradient variance of the \textbf{CFE} estimator with the \textbf{mean-field} parameterization
    & \cref{thm:cfe_upperbound_mf}
    & \labelcref{proof:cfe_upperbound_mf}
    \\
    \midrule
    \multicolumn{4}{l}{\textit{Complexity of Projected SGD}}
    \\
    & Iteration complexity of projected SGD with a \textbf{fixed stepsize} and a gradient estimator satisfying the \textbf{QV} condition on a \textbf{strongly convex} objective function
    & 
    \cref{thm:projsgd_stronglyconvex_fixedstepsize}
    & 
    \labelcref{proof:projsgd_stronglyconvex_fixedstepsize}
    \\
    & Iteration complexity of projected SGD with a \textbf{decreasing stepsize} schedule and a gradient estimator satisfying the \textbf{QV} condition on a \textbf{strongly convex} objective function 
    & 
    \cref{thm:projsgd_stronglyconvex_decstepsize}
    & 
    \labelcref{proof:projsgd_stronglyconvex_decstepsize}
    \\
    & Iteration complexity of projected SGD with a \textbf{fixed stepsize} and a gradient estimator satisfying the \textbf{adaptive QV} condition on a \textbf{strongly convex} objective function
    & 
    \cref{thm:projsgd_stronglyconvex_adaptive_complexity}
    & 
    \labelcref{proof:projsgd_stronglyconvex_adaptive_complexity}
    \\
    & Iteration complexity of projected SGD with a \textbf{decreasing stepsize} schedule and a gradient estimator satisfying the \textbf{adaptive QV} condition on a \textbf{strongly convex} objective function
    & 
    \cref{thm:projsgd_stronglyconvex_decstepsize_adaptive_complexity}
    & 
    \labelcref{proof:projsgd_stronglyconvex_decstepsize_adaptive_complexity}
    \\
    \midrule
    \multicolumn{4}{l}{\textit{Complexity of BBVI}} \\
    & Iteration complexity of BBVI with projected SGD using a \textbf{fixed stepsize} and the \textbf{CFE} gradient estimator on a \textbf{strongly log-concave} posterior 
    & 
    \cref{thm:projsgd_bbvicfe_complexity}
    & 
    \labelcref{proof:projsgd_bbvicfe_complexity}
    \\
    & Iteration complexity of BBVI with projected SGD using a \textbf{decreasing stepsize} schedule and the \textbf{CFE} gradient estimator on a \textbf{strongly log-concave} posterior 
    & 
    \cref{thm:projsgd_bbvicfe_decstepsize_complexity}
    & 
    \labelcref{proof:projsgd_bbvicfe_decstepsize_complexity}
    \\
    & Iteration complexity of BBVI with projected SGD using a \textbf{fixed stepsize} and the \textbf{STL} gradient estimator on a \textbf{strongly log-concave} posterior 
    & \cref{thm:projsgd_bbvistl_complexity}
    & \labelcref{proof:projsgd_bbvistl_complexity}
    \\
    & Iteration complexity of BBVI with projected SGD using a \textbf{decreasing stepsize} schedule and the \textbf{STL} gradient estimator on a \textbf{strongly log-concave} posterior 
    & 
    \cref{thm:projsgd_bbvistl_decstepsize_complexity}
    & 
    \labelcref{proof:projsgd_bbvistl_decstepsize_complexity}
    \\ \bottomrule
\end{tabularx}
  }%
\end{table*}
}

\newpage

\begin{table*}[t]
  \centering
  \vspace{-3ex}
\caption{Overview of Complexity Analyses of BBVI}\label{table:relatedworks}
  {\normalsize
\setlength{\tabcolsep}{4pt}
\renewcommand{\arraystretch}{1.2}
\begin{threeparttable}
\begin{tabular}{cccccccclc}
    \toprule
    \multicolumn{5}{c}{\textbf{Regularity of \(\pi\)}}
    & \multicolumn{1}{c}{\multirow{2}{*}{\textbf{\(q_{\vlambda^*} = \pi\)}}}
    & \multicolumn{1}{c}{\multirow{2}{*}{\textbf{\makecell{Optimized\\Parameters}}}}
    & \multicolumn{1}{c}{\multirow{2}{*}{\textbf{\makecell{Gradient\\Estimator\tnote{1}}}}}
    & \multicolumn{1}{c}{\multirow{2}{*}{\textbf{\makecell{Iteration\\Complexity}}}}
    & \multicolumn{1}{c}{\multirow{2}{*}{\textbf{Reference}}}
    \\\cmidrule{1-5}
    \multicolumn{1}{c}{\textbf{\(\mu\)-PL}}
    & \multicolumn{1}{c}{\textbf{LC}}
    & \multicolumn{1}{c}{\textbf{\(\mu\)-SLC}}
    & \multicolumn{1}{c}{\textbf{\(L\)-LS}}
    & \multicolumn{1}{c}{\textbf{LQ}}
    &
    & 
    &
    &
    &
    \\ \midrule
    \textcolor{black!20}{\ding{52}} & \textcolor{black!20}{\ding{52}} & \textcolor{black!20}{\ding{52}} & \textcolor{black!20}{\ding{52}} & \ding{52} &  \textcolor{black!20}{\ding{52}} & scale only & exact & \(\mathcal{O}\left(\log\left( L \epsilon^{-1} \right) \right)\) &  \citealp{hoffman_blackbox_2020} \\
    \textcolor{black!20}{\ding{52}} & \textcolor{black!20}{\ding{52}} & \textcolor{black!20}{\ding{52}} & \textcolor{black!20}{\ding{52}} & \ding{52} &  \textcolor{black!20}{\ding{52}} & scale only & CFE & \(\mathcal{O}\left(\kappa^2 \epsilon^{-1} \right)\) & \citealp{hoffman_blackbox_2020} \\
    \textcolor{black!20}{\ding{52}} & \textcolor{black!20}{\ding{52}} & \textcolor{black!20}{\ding{52}} & \textcolor{black!20}{\ding{52}} & \ding{52} &           & scale only & n/a\tnote{2} & \(\mathcal{O}\left(L \epsilon^{-1} \right)\)\tnote{3} & \citealp{bhatia_statistical_2022} \\
   \textcolor{black!20}{\ding{52}}  & \textcolor{black!20}{\ding{52}} & \ding{52} & \ding{52} &           &           & scale only & n/a\tnote{2} & \(\mathcal{O}\left(L \epsilon^{-1} \right)\)\tnote{3} & \citealp{bhatia_statistical_2022} \\
    \ding{52} &           &           & \ding{52} &           &           & loc. \& scale & CFE   & \(\mathcal{O}\left(L^2 \kappa \epsilon^{-4} \right)\) & \citealp{kim_convergence_2023} \\
    \textcolor{black!20}{\ding{52}} & \textcolor{black!20}{\ding{52}} & \ding{52} & \ding{52} &           &           & loc. \& scale & CFE   & \(\mathcal{O}\left(\kappa^2 \epsilon^{-1} \right)\) &  \makecell{\citealp{kim_convergence_2023}\\\citealp{domke_provable_2023}} \\
    & \ding{52} &  &  \ding{52} &   &           & loc. \& scale & CFE, STL & \(\mathcal{O}\left(L^2 \epsilon^{-2} \right)\) &  \citealp{domke_provable_2023} \\
    \textcolor{black!20}{\ding{52}} & \textcolor{black!20}{\ding{52}} & \ding{52} & \ding{52} &           &           & loc. \& scale & STL & \(\mathcal{O}\left(\kappa^2 \epsilon^{-1} \right)\) & \citealp{domke_provable_2023a} \\
    \textcolor{black!20}{\ding{52}} & \textcolor{black!20}{\ding{52}} & \ding{52} & \ding{52} &           & \ding{52} & loc. \& scale & STL & \(\mathcal{O}\left(\kappa^2 \log \epsilon^{-1} \right)\) & \citealp{domke_provable_2023a} \\
    \rowcolor{blue!10}
    \textcolor{black!20}{\ding{52}} & \textcolor{black!20}{\ding{52}} & \ding{52} & \ding{52} &           &           & loc. \& scale & STL   & \(\mathcal{O}\left(\kappa^2 \epsilon^{-1} \right)\) & \cref{thm:projsgd_bbvistl_decstepsize_complexity}\\
    \rowcolor{blue!10}
    \textcolor{black!20}{\ding{52}} & \textcolor{black!20}{\ding{52}} & \ding{52} & \ding{52} &           & \ding{52} & loc. \& scale & STL   & \(\mathcal{O}\left(\kappa^2 \log \epsilon^{-1} \right)\) &  \cref{thm:projsgd_bbvistl_complexity}
    \\ \bottomrule
\end{tabular}
\begin{tablenotes}
\item[*] PL: Polyak-\L{}ojasiewicz, LC: log-concave, SLC: strongly-log-concave, LQ: log-quadratic (\(\pi\) is Gaussian), \(\kappa = L/\mu\), and \(q_{\vlambda_*} = \pi\) implies that ``the variational family is perfectly specified'' such that \(\pi \in \mathcal{Q}\).
\item[*] Conditions implied by other stronger conditions (marked with \ding{52}) are marked with \textcolor{black!20}{\ding{52}}.
\item[*] Analyses that a-priori assumed regularity of the ELBO were omitted.
\item[*] The explicit dimension dependences are omitted, but in general, \(\mathcal{O}\left(d\right)\) for full-rank, which is tight~\citep{domke_provable_2019}, and the best known for mean-field is {\scriptsize\(\mathcal{O}\left(\sqrt{d}\right)\)}~\citep{kim_practical_2023}.
The algorithm of \citet{bhatia_statistical_2022} is able to trade the dimension dependence for statistical accuracy.
\item[1] The precise definitions of the gradient estimators are in \cref{section:gradient_estimators}.
\item[2] This algorithm uses stochastic power method-like iterations.
\item[3] The per-iteration sample complexity also depends on \(L, d, \epsilon\).
\end{tablenotes}
\end{threeparttable}
  }%
  \vspace{-1ex}
\end{table*}


\vspace{-1ex}
\section{RELATED WORKS}\label{section:related}
\vspace{-1ex}

\paragraph{Analyzing the Computational Properties of BBVI}
Since its inception by \citet{ranganath_black_2014,titsias_doubly_2014}, theoretical results on BBVI have been developing on two different axes: 
\begin{enumerate*}[label=\textbf{(\alph*)}]
    \item Analyzing the regularity of the ELBO such as convexity and smoothness \citep{challis_gaussian_2013,titsias_doubly_2014}, 
    \item and analyzing the variance of the Monte Carlo gradient estimators \citep{fan_fast_2015,xu_variance_2019,mohamed_monte_2020,buchholz_quasimonte_2018}.
\end{enumerate*}
While some convergence analyses of BBVI have been provided \citep{regier_fast_2017,khan_faster_2016,khan_kullbackleibler_2015,buchholz_quasimonte_2018,liu_quasimonte_2021,alquier_concentration_2020,fujisawa_multilevel_2021,alquier_nonexponentially_2021}, these works \textit{a priori} assumed the regularity of the ELBO and the gradient estimators.
Due to the difficulty of rigorously establishing these conditions, later works by \citet{hoffman_blackbox_2020,bhatia_statistical_2022} have worked with simplified or alternative implementations of BBVI.
Meanwhile, \citet{xu_computational_2022} showed these regularities can be realized asymptotically in high probability.
In expectation, however, it was only recently that regularity conditions on the ELBO~\citep{domke_provable_2020,kim_convergence_2023} and the reparameterization gradient estimator~\citep{kim_practical_2023,domke_provable_2019} were shown to be realizable under mild conditions without modifying the algorithms used in practice.

\paragraph{Concurrent Results by \citet{domke_provable_2023a}}
While our work builds on top of the QV-based framework of \citet{domke_provable_2023}, a similar convergence result on the STL estimator appeared in its later version~\citep{domke_provable_2023a} concurrently with our work.
However, our results differ in several aspects:
\begin{enumerate}
    \setlength\itemsep{-.5ex}
    \item We bound the family-misspecification term \(T_{\text{\ding{184}}}\) with the Fisher-Hyv\"arinen divergence \(D_{\mathrm{F}^4}\left(q_{\vlambda^*}, \pi\right)\), while \citet{domke_provable_2023a} bound it with a quadratic involving the smoothness constant of the residual function \(r\left(\vz\right) \triangleq \log q_{\vlambda^*}\left(\vz\right) - \log \pi\left(\vz\right)\).
    
    \item For the Gaussian posterior case, the constants of \cref{thm:stl_upperbound} are tighter that of \citet{domke_provable_2023a}.
    
    \item We also provide an upper bound for the mean-field variational family in \cref{thm:stl_upperbound_mf}.
    
    \item We establish a lower bound on the gradient variance of STL, quantifying the tightness of the bounds.
\end{enumerate}

\onecolumn

\newpage
\section{PROOFS}\label{section:proofs}

\vspace{-1ex}
\subsection{Definitions}\label{section:definitions}
\vspace{1ex}

\begin{definition*}[\textbf{\(L\)-Smoothness}]
  A function \(f : \mathcal{X} \to \mathbb{R}\) is \(L\)-smooth if it satisfies
  \[
    \norm{ \nabla f\left(\vx\right) - \nabla f\left(\vy\right) }_2 \leq L \norm{ \vx - \vy }_2
  \]
  for all \(\vx, \vy \in \mathcal{X}\) and some \(0 < L < \infty\).
\end{definition*}

\begin{definition*}[\textbf{\(\mu\)-Strong Convexity}]
  A function \(f : \mathcal{X} \to \mathbb{R}\) is \(\mu\)-strongly convex if it satisfies
  \[
     \frac{\mu}{2} \norm{\vx - \vy}_2^2  \leq f\left(\vy\right) - f\left(\vx\right) - \inner{ \nabla f\left(\vx\right) }{ \vy - \vx }
  \]
  for all \(\vx, \vy \in \mathcal{X}\) and some \(0 < \mu < \infty\).
\end{definition*}

\vspace{1ex}
\begin{remark}
    We say a function \(f\) is only convex if it satisfies the strong convexity inequality with \(\mu = 0\).
\end{remark}





\newpage
\subsection{Auxiliary Lemmas}
\vspace{1ex}


\begin{theoremEnd}[all end, category=external]{lemma}[\citealt{domke_provable_2019}, Lemma 9]
\label{thm:u_identities}
  Let \(\rvvu = \left(\rvu_1, \rvu_2, \ldots, \rvu_d\right)\) be a \(d\)-dimensional vector-valued random variable with zero-mean independently and identically distributed components.
  Then,
  \begin{alignat*}{4}
    &\mathbb{E}\rvvu \rvvu^{\top} &= \left( \mathbb{E} \rvu_i^2 \right) \boldupright{I},\qquad
    &\mathbb{E}\norm{\rvvu}_2^2 &&= d \, \mathbb{E} \rvu_i^2, \\
    &\mathbb{E} \rvvu \left( 1 + \norm{\rvvu}_2^2 \right) &= \left( \mathbb{E} \rvu_i^3 \right) \mathbf{1}, \qquad
    &\mathbb{E} \rvvu \rvvu^{\top} \rvvu \rvvu^{\top} &&= \left( \left(d - 1\right) \, {\left( \mathbb{E} \rvu_i^2 \right)}^2 + \mathbb{E}\rvu_i^4 \right) \boldupright{I}.
  \end{alignat*}
\end{theoremEnd}


\begin{theoremEnd}[all end, category=external]{lemma}\label{thm:jacobian_reparam_inner}
  Let \(\mathcal{T}_{\vlambda}: \mathbb{R}^p \times \mathbb{R}^d \rightarrow \mathbb{R}^d\) be the location-scale reparameterization function (\cref{def:reparam}).
  Then, for any differentiable function \(f\), we have
  \[
    \norm{\nabla_{\vlambda} f\left( \mathcal{T}_{\vlambda}\left(\vu\right) \right) }_2^2 
    = 
    J_{\mathcal{T}}\left(\rvvu\right) \norm{\nabla f\left( \mathcal{T}_{\vlambda}\left(\vu\right) \right) }_2^2.
  \]
  for any \(\vlambda \in \mathbb{R}^p\) and \(\vu \in \mathbb{R}^d\), where \(J_{\mathcal{T}}\left(\vu\right) : \mathbb{R}^d \to \mathbb{R}\) is a function defined as
  \begin{alignat*}{3}
    &J_{\mathcal{T}}\left(\vu\right) &= 1 + {\textstyle\sum^{d}_{i=1} u_i^2}                         &\qquad \text{for the full-rank and} \\
    &J_{\mathcal{T}}\left(\vu\right) &= 1 + {\textstyle\sqrt{\sum^{d}_{i=1} u_i^4}} &\qquad \text{for the mean-field parameterizations.}
  \end{alignat*}
\end{theoremEnd}
\begin{proofEnd}
  The result is a collection of the results of \citet[Lemma 1]{domke_provable_2019} for the full-rank parameterization and \citet[Lemma 2]{kim_practical_2023} for the mean-field parameterization.
\end{proofEnd}

\begin{theoremEnd}[all end, category=external]{lemma}[Corollary 2; \citealp{kim_convergence_2023}]\label{thm:u_normsquared_marginalization}
  Assume \cref{assumption:symmetric_standard} and let \(\mathcal{T}_{\vlambda}: \mathbb{R}^d \rightarrow \mathbb{R}^d\) (\cref{def:reparam}) be the location-scale reparameterization function.
  Then, for any \(\vlambda, \vlambda^{\prime} \in \mathbb{R}^p\),
  \[
    \mathbb{E} J_{\mathcal{T}}\left(\rvvu\right) \norm{\mathcal{T}_{\vlambda^{\prime}}\left(\rvvu\right) - \mathcal{T}_{\vlambda}\left(\rvvu\right) }_2^2
    \leq
    C\left(d, \varphi\right) \norm{\vlambda - \vlambda'}_2^2,
  \]
  where \(\; C\left(d, \varphi\right) = d + k_{\varphi} \;\) for the full-rank and \(\;C\left(d, \varphi\right) = 2 k_{\varphi} \sqrt{d} + 1\;\) for the mean-field parameterizations. 
\end{theoremEnd}

\begin{theoremEnd}[all end, category=external]{lemma}[Lemma 2; \citealp{domke_provable_2019}]\label{thm:normdist_1pnormu}
  Assume \cref{assumption:symmetric_standard} and let \(\mathcal{T}_{\vlambda}: \mathbb{R}^d \rightarrow \mathbb{R}^d\) (\cref{def:reparam}) be the location-scale reparameterization function.
  Then, for the full-rank parameterization,
  {\small
  \[
    \mathbb{E} J_{\mathcal{T}}\left(\rvvu\right) \norm{\mathcal{T}_{\vlambda}\left(\rvvu\right) - \bar{\vz}}_2^2
    =
    C_1\left(d, \varphi\right) \norm{ \vm - \bar{\vz} }_2^2 + C_2\left(d, \varphi\right) \norm{\mC}_{\mathrm{F}}^2.
  \]
  }%
  where
  \begin{alignat*}{3}
    C_1\left(d, \varphi\right)  &= d + 1,  \qquad  &C_2\left(d, \varphi\right) &= d + k_{\varphi}, &&\quad \text{for the full-rank and} \\
    C_1\left(d, \varphi\right)  &= \sqrt{d k_{\varphi}} + k \sqrt{d} + 1, \qquad
    &C_2\left(d, \varphi\right) &= 2 \kappa \sqrt{d} + 1, &&\quad \text{for the mean-field parameterizations.}
  \end{alignat*}
\end{theoremEnd}
\begin{proofEnd}
  The result is a collection of the results of \citet[Lemma 2]{domke_provable_2019} for the full-rank parameterization and \citet[Lemma 2]{kim_practical_2023} for the mean-field parameterization.
\end{proofEnd}

\printProofs[external]
\printProofs[gradvarlemmas]

\clearpage
\subsection{Smoothness Under Triangular Scale Parameterization}\label{proof:entropy_smoothness}
\vspace{1ex}
\printProofs[entropysmooth]

\clearpage
\subsection{Upper Bound on Gradient Variance of STL}

\subsubsection{General Decomposition}
\vspace{1ex}
\printProofs[stlupperboundlemma]

\newpage
\subsubsection{Full-Rank Parameterization}
\vspace{1ex}
\printProofs[stlupperboundfr]

\newpage
\subsubsection{Mean-Field Parameterization}\label{section:stl_meanfield}
\vspace{1ex}
\printProofs[stlupperboundmf]

\clearpage
\subsection{Lower Bound on Gradient Variance of STL}
\subsubsection{General Lower Bound}
\vspace{1ex}
\printProofs[stllowerbound]

\newpage
\subsubsection{Unimprovability}
\vspace{1ex}
\printProofs[stllowerboundunimprovability]

\clearpage
\subsection{Upper Bound on Gradient Variance of CFE}
\subsubsection{Full-Rank Parameterization}
\vspace{1ex}
\printProofs[cfeupperbound]

\newpage
\subsubsection{Mean-Field Parameterization}
\vspace{1ex}
\printProofs[cfeupperboundmf]

\clearpage
\subsection{Non-Asymptotic Complexity of Projected SGD}\label{section:projsgdcomplexity}

To precisely compare the computational complexity resulting from different estimators, we refine the convergence analyses of~\citet{domke_provable_2023}.
Specifically, we obtain precise complexity guarantees from their ``anytime convergence'' statements.
This type of convergence analysis, which has been popular in the ERM sample selection strategy literature~\citep[\S 1.1]{csiba_importance_2018}, is convenient for comparing the lower-order and constant factor improvements of different gradient estimators.

\subsubsection{QVC Gradient Estimator}
\vspace{1ex}
\printProofs[complexityprojsgdqvcfixed]
\newpage
\printProofs[complexityprojsgdqvcdec]

\clearpage
\subsubsection{Adaptive QVC Gradient Estimator}\label{section:complexity_adaptiveqvc}
\vspace{1ex}

As mentioned at the beginning of \cref{section:gradient_variance}, we established \textit{adaptive} QV bounds.
For the complexity guarantees for strongly convex objectives (\cref{thm:projsgd_stronglyconvex_fixedstepsize,thm:projsgd_stronglyconvex_decstepsize}), it is possible to optimize the free parameter \(\delta\) in the bounds, such that they automatically adapt to other problem-specific constants.
In this section, we do this for both SGD with fixed stepsize (\cref{thm:projsgd_stronglyconvex_adaptive_complexity}) and a decreasing (\cref{thm:projsgd_stronglyconvex_decstepsize_adaptive_complexity}) stepsize schedule.
\vspace{2ex}
\printProofs[complexityprojsgdadaptiveqvcfixed]

\newpage
\printProofs[complexityprojsgdadaptiveqvcdec]

\clearpage
\subsection{Non-Asymptotic Complexity of BBVI}
\subsubsection{CFE Gradient Estimator}
\vspace{1ex}
\printProofs[complexitybbvicfefixed]
\newpage
\printProofs[complexitybbvicfedec]

\newpage
\subsubsection{STL Gradient Estimator}
\vspace{1ex}
\printProofs[complexitybbvistl]
\newpage
\printProofs[complexitybbvistldec]

\newpage
\subsection{Fisher-Hyv\"arinen Divergence Between Gaussians}
\vspace{1ex}
\printProofs[stlgaussianfisher]
\newpage
\printProofs[gaussianklmeanfield]
\newpage
\printProofs[stlgaussian]



\end{document}